\documentclass[10pt]{article} %
\usepackage[preprint]{tmlr}

\usepackage{amsmath,amsfonts,bm}

\def\eqref#1{equation~\ref{#1}}

\def\1{\bm{1}}

\DeclareMathAlphabet{\mathsfit}{\encodingdefault}{\sfdefault}{m}{sl}
\SetMathAlphabet{\mathsfit}{bold}{\encodingdefault}{\sfdefault}{bx}{n}

\usepackage{hyperref}
\hypersetup{hidelinks}
\usepackage{url}

\usepackage{amsmath}
\usepackage{amssymb}
\usepackage{mathtools}
\usepackage{amsthm}
\usepackage{enumitem}
\usepackage{adjustbox}
\usepackage{tikz}
\usetikzlibrary{arrows.meta,positioning,shapes.geometric,backgrounds}
\usepackage{subcaption}
\pgfdeclarelayer{bg}
\pgfsetlayers{bg,main}
\usepackage{dblfloatfix}
\usepackage{algorithm}
\usepackage{algorithmic}
\usepackage{multirow}
\usepackage{booktabs}
\usepackage{array}
\usepackage{graphicx}
\graphicspath{{./}}

\newcommand{\meanstd}[2]{\shortstack{#1\\[-0.2ex]$\pm$#2}}

\theoremstyle{plain}

\theoremstyle{definition}

\theoremstyle{remark}

\title{EVA-Flow: Learning Shared Chemistry for Unified \\
Environment-Aware Molecular Conformation Generation}

\author{\name Bing Yan \email bing.yan@nyu.edu \\
      \addr FAIR at Meta; New York University
      \AND
      \name Anuroop Sriram \email anuroops@meta.com \\
      \addr FAIR at Meta
      \AND
      \name Zachary Ward Ulissi \email zulissi@meta.com \\
      \addr FAIR at Meta
      \AND
      \name Kyunghyun Cho \email kyunghyun.cho@nyu.edu \\
      \addr New York University
      \AND
      \name Ricky T.Q. Chen \email rtqichen@meta.com \\
      \addr FAIR at Meta
      \AND
      \name Benjamin Kurt Miller \email bkmi@meta.com \\
      \addr FAIR at Meta; New York University}

\def\month{10}  %
\def\year{2026}
\def\openreview{\url{https://openreview.net/forum?id=pD4Smou3Xp}}

\begin{document}

\maketitle

\begin{abstract}
Molecular 3D structures underpin simulations of thermodynamics, kinetics, crystallization, and nucleation, making their accurate prediction central to drug discovery and materials design.
However, molecular geometry can change dramatically across environments. Existing models either ignore environmental effects or train separate models for each environment. We hypothesize that transfer learning across contexts is possible for conformer generation and that a general model will perform better when training data are limited, as in ligand binding and crystallization.
We introduce \emph{EVA-Flow}, a unified framework for environment-aware molecular conformation generation that learns shared chemical structure and adapts it through environment-conditioned generative modeling.
Across vacuum, protein-ligand docking, solvation, and crystal packing, pretraining and cross-environment finetuning provide their largest gains in the three data-scarce settings.
Evaluation against multiple reference conformations shows that EVA-Flow covers a broad range of solvation conformations. Analyses of molecules shared across environments support its ability to generate different conformations for the same molecule in different environments.
\end{abstract}

\noindent\textbf{Code release:} \url{https://huggingface.co/facebook/eva-flow}.

\section{Introduction}
\label{sec:intro}

Molecules do not exist in a single fixed shape. A drug molecule may be synthesized in solution, crystallized for storage, redissolved in the body, and finally bind a target protein. At each stage the surrounding environment, whether solvent~\citep{sobornova2024influence}, crystal lattice~\citep{doi:10.1021/jacs.5c09997,C4SC01132E,cruz2014conformational}, or protein pocket~\citep{GAO2010901,10.1371/journal.pone.0000820,weikl2014conformational}, reshapes its 3D geometry.

Despite this, most generative models treat conformer generation as environment-agnostic: molecules are modeled in vacuum~\citep{xu2022geodiff, hassan2024etflow} or within a single context such as protein--ligand docking~\citep{corso2022diffdock, corso2024deep}, with a separate model for each setting.
This fragmentation implicitly assumes that conformational distributions across environments are unrelated tasks. Yet while conformations change across environments, the underlying chemistry does not---the environment imposes physical forces that deform a shared molecular structure into different geometries.

This motivates formulating conformer generation as learning \emph{shared invariant chemistry} under environment-dependent physical constraints, enabling a single model to transfer chemical knowledge across environments via environment-specific conditioning.

\begin{figure}[t]
    \centering
    \includegraphics[width=0.98\linewidth]{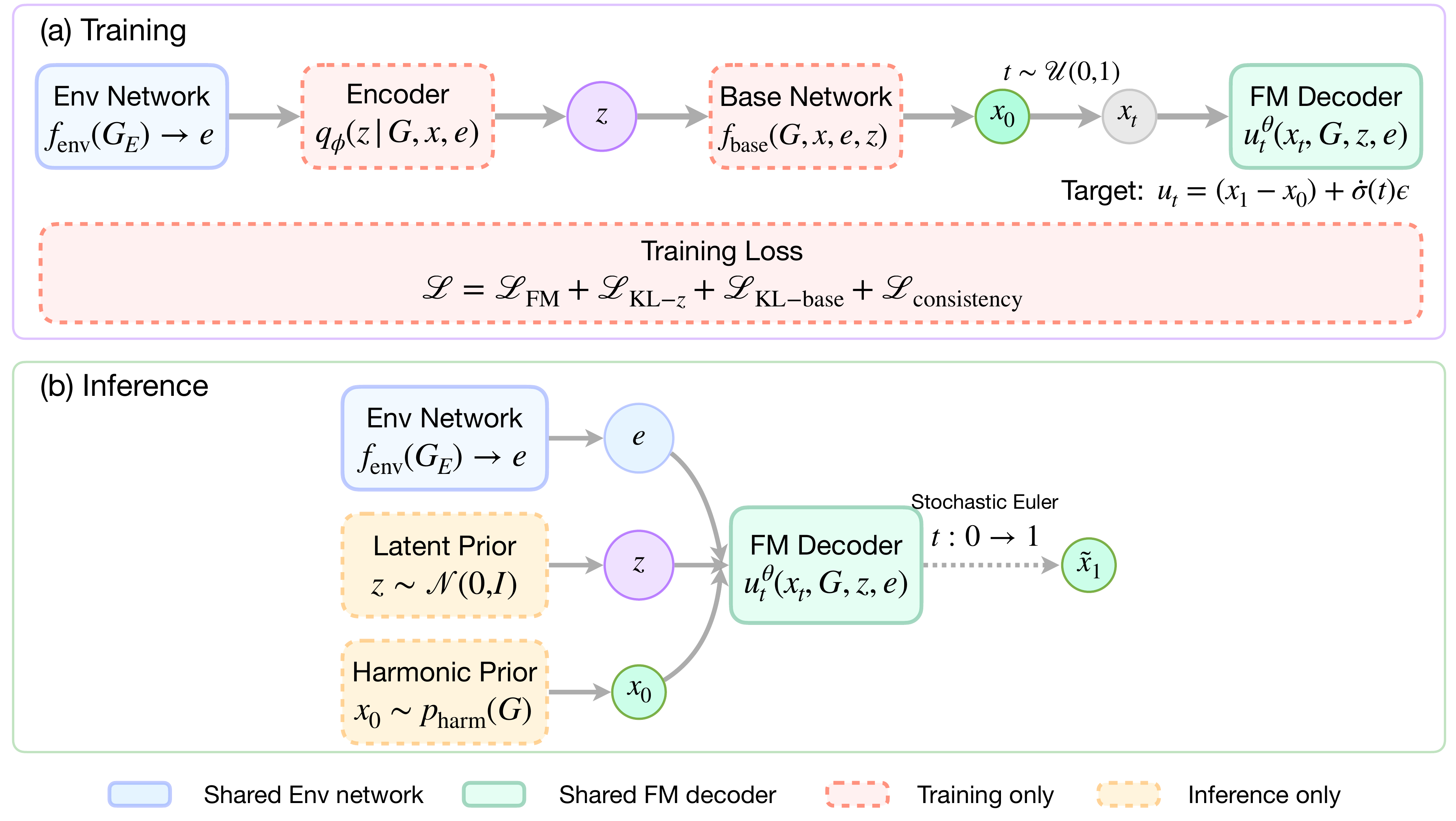}
    \caption{Architecture of EVA-Flow.
    \textbf{(a) Training.} The environment graph $G_E$ is embedded by $f_{\mathrm{env}}$ into a vector $e$.
    The encoder $q_\phi(z \mid G, x, e)$ maps the molecular graph $G$, ground-truth positions $x$, and environment $e$ to a latent $z$.
    The base network $f_{\mathrm{base}}(G, x, e, z)$ produces a distribution $p_0(x_0 \mid G, x, e, z)$ from which $x_0$ is sampled.
    The FM decoder learns $u_t^\theta(x_t, G, z, e)$ with stochastic interpolation $x_t = (1{-}t)\,x_0 + t\,x_1 + \sigma(t)\epsilon$, targeting $u_t = (x_1 - x_0) + \dot\sigma(t)\epsilon$.
    Total loss: $\mathcal{L} = \mathcal{L}_{\mathrm{FM}} + \mathcal{L}_{\mathrm{KL\text{-}z}} + \mathcal{L}_{\mathrm{KL\text{-}base}} + \mathcal{L}_{\mathrm{consistency}}$.
    \textbf{(b) Inference.} The encoder and base network are bypassed: $e = f_{\mathrm{env}}(G_E)$, $z \sim \mathcal{N}(\mathbf{0}, I)$, and $x_0 \sim p_{\mathrm{harm}}(G)$ are sampled independently. All three condition the FM decoder, which generates conformations from $t{=}0$ to $t{=}1$ using the stochastic Euler sampler described in Section~\ref{sec:architecture}. The Env Network and FM Decoder share weights between training and inference.}
    \label{fig:model}
\end{figure}

We propose \emph{EVA-Flow}, a unified generative framework for environment-aware molecular conformation generation (Figure~\ref{fig:model}). EVA-Flow couples a variational autoencoder (VAE)~\citep{kingma2019introduction} encoder with a flow-matching (FM) decoder~\citep{lipman2022flow} that generates conformations conditioned on the molecular graph $G$ and environment embedding $e$.

Our contributions are:
\begin{itemize}[leftmargin=0.5cm]
    \item We formulate conformer generation as learning shared chemistry under environment-specific constraints, rather than training isolated models per environment.
    \item We introduce a unified environment-conditioned flow-matching architecture that handles diverse environments within a single model, combining an environment encoder, a learned harmonic base, and a variational latent that together support low-data cross-environment transfer.
    \item We demonstrate that pretraining and cross-environment finetuning improve accuracy most clearly in data-scarce environments, and provide same-molecule and multi-reference evidence that the model responds to environmental context. UMA relaxation separately supports physical validity for generated Vacuum conformers (Appendix~\ref{appendix:uma}).
\end{itemize}

\section{Related Work}
\label{sec:related}

\textbf{Flow matching.}
Flow matching (FM) trains a continuous normalizing flow by regressing the velocity field that transports a base distribution to the data, yielding a simulation-free objective~\citep{lipman2022flow}. Rectified flow~\citep{liu2022flow} and conditional FM~\citep{tong2023improving} extend this to straight-line paths and conditional generation. Equivariant~\citep{hassan2024etflow} and geometric~\citep{chen2023flow} variants adapt FM to 3D molecular settings.

\textbf{VAE + FM.}
Hybrid VAE--flow models take two forms: (1) FM in the latent space replaces the VAE prior~\citep{dao2023flow}, or (2) FM serves as the decoder, mapping a base distribution to data conditioned on a latent code~\citep{fischer2023boosting,sargent2025flow}. We adopt the latter, using an environment-aware FM decoder for conformer generation.

\textbf{Molecular conformation generation.}
Classical force-field and MD approaches are accurate but expensive. Neural methods generate conformers via equivariant diffusion~\citep{xu2022geodiff}, flows~\citep{hassan2024etflow}, or torsion models~\citep{jing2022torsional}. Most target a single environment---typically vacuum or protein--ligand docking~\citep{corso2024deep}---requiring separate models per setting. EVA-Flow addresses this limitation with a single model that conditions on structured environments.

\textbf{Diffusion in latent space for molecules.}
\citet{joshi2025all} run diffusion in a learned latent space and decode to coordinates. EVA-Flow also uses a VAE encoder but performs FM in data space and explicitly conditions on the environment.

\section{Environment-Aware Flow Matching}
\label{sec:method}

\subsection{Problem Formulation}
\begin{samepage}
Let $G=(V,\mathcal{E})$ denote a molecular graph with nodes representing atoms and edges representing bonds. Each atom is associated with a 3D position $x \in \mathbb{R}^{3N}$, where $N$ is the number of atoms. A molecule exists in an environment $E$, such as a protein pocket, solvent box, or crystalline lattice. Our general objective is the conditional distribution $p(x|G,E)$, capturing how environments modulate molecular conformations. We seek one model that learns shared molecular structure and transfers across the environment types observed during training.

\end{samepage}

EVA-Flow uses two graph objects with distinct roles. The \emph{target molecular graph} $G$ always describes the molecule whose centered coordinates $x$ are generated; its nodes are molecular atoms and its edges are covalent bonds. The \emph{environment graph} $G_\mathrm{E}$ is conditioning context only: it is embedded into a fixed-dimensional vector $e=f_\mathrm{env}(G_\mathrm{E})$ and is never generated. Table~\ref{tab:graph-roles} makes the environment-specific constructions explicit. During training the encoder and learned base additionally access the target coordinates $x$; at inference only $G$, $G_\mathrm{E}$, and their listed attributes are required.

\begin{table*}[t]
\centering
\caption{Roles and construction of the target graph $G$ and environment graph $G_\mathrm{E}$. The target graph has the same semantics in every setting; only the context representation changes. ``Coordinates'' indicates coordinates supplied as input, not the target coordinates to be generated.}
\label{tab:graph-roles}
\begingroup
\footnotesize
\setlength{\tabcolsep}{3pt}
\renewcommand{\arraystretch}{1.15}
\begin{tabular}{@{}>{\raggedright\arraybackslash}p{0.1\linewidth}>{\raggedright\arraybackslash}p{0.11\linewidth}>{\raggedright\arraybackslash}p{0.17\linewidth}>{\raggedright\arraybackslash}p{0.15\linewidth}>{\raggedright\arraybackslash}p{0.23\linewidth}>{\raggedright\arraybackslash}p{0.17\linewidth}@{}}
\toprule
Stage/\newline environment & Target $G$ & $G_\mathrm{E}$ nodes & $G_\mathrm{E}$ edges & Context attributes / coordinates & Available at inference \\
\midrule
Pretraining & Molecule atoms/bonds & Random 30--70\% atom-induced subgraph & Retained covalent bonds & Atom features and masked-subgraph coordinates & N/A (pretraining only) \\
Vacuum & Molecule atoms/bonds & Bemis--Murcko scaffold atoms & Scaffold covalent bonds & Atom features; coordinates set to zero & Yes, constructed from input molecule \\
Docking & Ligand atoms/bonds & Protein-pocket heavy atoms & Atoms within the same residue & Atom features and pocket coordinates & Yes, extracted pocket \\
Solvation & Solute atoms/bonds & Atoms of a water proxy & Water covalent bonds & sCAV, vCAV, eSOLV, eNELEC; fixed proxy coordinates & Yes, solvent descriptors \\
Packing & Molecule atoms/bonds & Copy of molecular atoms & Molecular covalent bonds & RDKit geometry, space group, lattice $(a,b,c,\alpha,\beta,\gamma)$ & Yes, crystal metadata \\
\bottomrule
\end{tabular}
\endgroup
\end{table*}

The molecular encoder/decoder backbone and conditioning mechanism are shared across environments. Environment-specific graph constructions incorporate domain knowledge (Table~\ref{tab:graph-roles}).

\subsection{Model Architecture}
\label{sec:architecture}
EVA-Flow is a VAE with an FM decoder (Figure \ref{fig:model}).
\begin{enumerate}[leftmargin=*]
    \item \textbf{Environment Embedding Network} (Figure \ref{fig:model}).
    The environment is represented as a graph $G_\mathrm{E} = (V_\mathrm{E}, \mathcal{E}_\mathrm{E})$, where nodes correspond to atoms with associated features and edges correspond to bonds or interactions. The construction of $G_\mathrm{E}$ is described in Section \ref{sec:dataset}. The environment network, $f_\mathrm{env}$, consists of three graph convolutional network (GCN) layers~\citep{zhang2019graph}, followed by a two-layer MLP projection to the environment embedding vector $e \in \mathbb{R}^d$:
    \begin{equation}
        e = f_\mathrm{env}(G_\mathrm{E})
    \end{equation}
    The embedding $e$ conditions both the molecular encoder and decoder. The GCN concatenates coordinates with node features and is not rotation invariant. Appendix~\ref{appendix:env-equivariance} compares it with an SE(3)-equivariant environment encoder.
    
    \item \textbf{Encoder} (Figure \ref{fig:model}).
    The encoder is a GNN-based module that takes as input the molecular graph $G$, atomic positions $x$, and the environment embedding $e$. 
    Input node features are formed by concatenating the atomic number, node attributes, and atomic positions. 
    These features are processed by an SE(3)-equivariant transformer (TorchMD-ET)~\citep{thölke2022torchmdnetequivarianttransformersneural,hassan2024etflow}, producing node embeddings of size $[N,\,\text{hidden}]$.
    Each node embedding is then concatenated with the environment embedding $e$, and projected through two linear layers to produce the mean and log-variance for a node-level latent variational posterior distribution:
    \begin{equation}
    \begin{aligned}
    &\quad q_\phi(z \mid G, x, e) \\
    &= \prod_{i=1}^{N} \mathcal{N}\!\Big(
    z_i \,\Big|\,
    \mu_{\phi,i}(G, x, e), 
    \mathrm{diag}\!\big(\sigma_{\phi,i}^2(G, x, e)\big)
    \Big) \\
    &= \mathcal{N}\!\big(\mu_\phi(G, x, e), \, \mathrm{diag}(\sigma_\phi^2(G, x, e))\big).
    \end{aligned}
    \end{equation}
    where $\mu_{\phi,i}, \sigma_{\phi,i}^2 \in \mathbb{R}^d$ denote the mean and variance predicted for node $i$. Stacking these predictions over all nodes gives $\mu_\phi, \sigma_\phi^2 \in \mathbb{R}^{N \times d}$.
    The latent pathway captures residual molecule--environment variability during training. The reported checkpoint has a collapsed latent posterior, while the coupled base and latent aid low-data cross-environment transfer (Appendices~\ref{appendix:latent-diagnostics} and~\ref{appendix:latent-free-grid}).

    \item \textbf{Base Distribution Network} (Figure \ref{fig:model}).
    We use an SE(3)-equivariant transformer (TorchMD-ET) to parameterize a conditional Harmonic base distribution over atomic coordinates. The base distribution is commonly parameterized with zero mean and the target molecule's graph Laplacian as precision matrix. We condition this distribution by predicting the mean and diagonal scaling:
    \begin{equation}
    \mu_\mathrm{base}, \, \lambda_\mathrm{base} = f_\mathrm{base}(G,x,e,z),
    \end{equation}
    where $\mu_\mathrm{base} \in \mathbb{R}^{N\times 3}$ is vector-valued and $\lambda_\mathrm{base} \in \mathbb{R}^{N}$ is scalar-valued. Our conditional distribution is then: 
    \begin{align}
        L_\mathrm{base} &= L + \mathrm{diag}(\lambda_\mathrm{base}), \\
        \Sigma_\mathrm{base} &= L_\mathrm{base}^{\dagger}, \\
        p_0(x_0 \mid G,x,e,z)
        &= \mathcal{N}\!\Big(
        x_0 \,\Big|\,
        \mu_\mathrm{base}, \Sigma_{\text{base}}\Big).
    \end{align}
    where $(\cdot)^{\dagger}$ is the Moore--Penrose pseudoinverse to handle global translational symmetries.
    Let $L_\mathrm{base} = P D P^{\top}$ be the eigendecomposition. Sampling is performed via:
    \begin{equation}
    x_0 = \mu_\mathrm{base} + PD^{-1/2}\epsilon, \quad \epsilon \sim \mathcal{N}(\mathbf{0}, I)
    \end{equation}
    The network is trained by minimizing the KL divergence between the learned distribution and a Harmonic prior defined by the molecular graph topology $\mathcal{N}(\mathbf{0}, L^{\dagger})$. During training the network accesses the ground truth $x$ for conditioning. The base network is not used at inference.

    \item \textbf{FM Decoder} (Figure \ref{fig:model}).
    The decoder generates conformations by transporting samples from the base distribution $p_0$ to the data distribution using stochastic flow matching. 
    We adopt the TorchMD equivariant network to parameterize a time-dependent vector field $u_t^\theta$ that takes as input the current molecular structure $x_t$, the molecular graph $G$, the latent variables $z$, and the environment embedding $e$:
    \begin{align}
    x_0 &\sim p_0'(x_0) = \begin{cases} p_0(x_0 \mid G,x,z,e) & \text{Training} \\
    \mathcal{N}\left(x_0 \mid \mathbf{0}, L^{\dagger}\right) & \text{Inference}
    \end{cases} \\
    \dot x_t &= u_t^\theta(x_t, G, z, e)
    \label{eqn:ode}
    \end{align}
    We use a stochastic interpolation between base and target samples. Given a base sample $x_0 \sim p_0'$ and a data sample $x_1 \sim p_\mathrm{data}$, we define:
    \begin{equation}
    \begin{aligned}
    \mu_t &= (1-t)x_0 + t x_1, \\
    x_t &= \mu_t + \sigma(t)\,\epsilon, \quad \epsilon \sim \mathcal{N}(\mathbf{0}, I)
    \end{aligned}
    \end{equation}

    We use the noise schedule:
    \begin{equation}
        \sigma(t) = \sigma \sqrt{t(1-t)}
    \end{equation}

    \item \textbf{Inference} (Figure \ref{fig:model}).
    At inference time, given the environment graph $G_E$, we compute the embedding $e=f_{\mathrm{env}}(G_E)$ and sample a latent $z\sim p(z)=\mathcal N(\mathbf{0},I)$.
    The initial positions are sampled from the simple Harmonic prior $x_0 \sim \mathcal{N}(\mathbf{0}, L^{\dagger})$.
    Starting at $x_{t_0}=x_0$, we integrate the learned flow field from $t=0$ to $t=1$ using stochastic Euler steps with zero-center-of-mass Gaussian perturbations. For $0=t_0<t_1<\cdots<t_n=1$,
    \begin{equation}
    \begin{aligned}
        x_{t_{i+1}} &= x_{t_i} + u_\theta(x_{t_i}, t_i; G,z,e)\,\Delta t_i \; \\
        &\quad + \sqrt{\sigma_{\mathrm{samp}}^2(t_{i+1})-\sigma_{\mathrm{samp}}^2(t_i)}\,\xi_i, \\
        \xi_i &\sim \mathcal N(\mathbf{0},I), \qquad \text{with COM}(\xi_i)=0,
    \end{aligned}
    \end{equation}
    where the sampler uses the variance schedule $\sigma_{\mathrm{samp}}^2(t)=t^2$ and $\Delta t_i=t_{i+1}-t_i$. We refer to this procedure as the ``SDE sampler'' in the experimental tables (Appendix~\ref{appendix:impl}).
    The final conformer $\hat{x}$ is obtained after $n$ steps.

\end{enumerate}

\begin{figure}[t]
\centering
\includegraphics[width=0.48\linewidth]{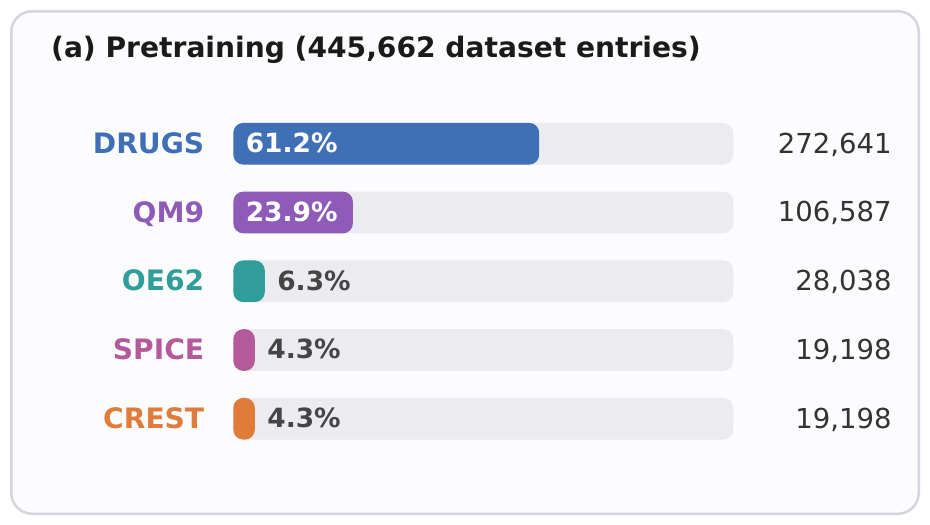}\hfill
\includegraphics[width=0.48\linewidth]{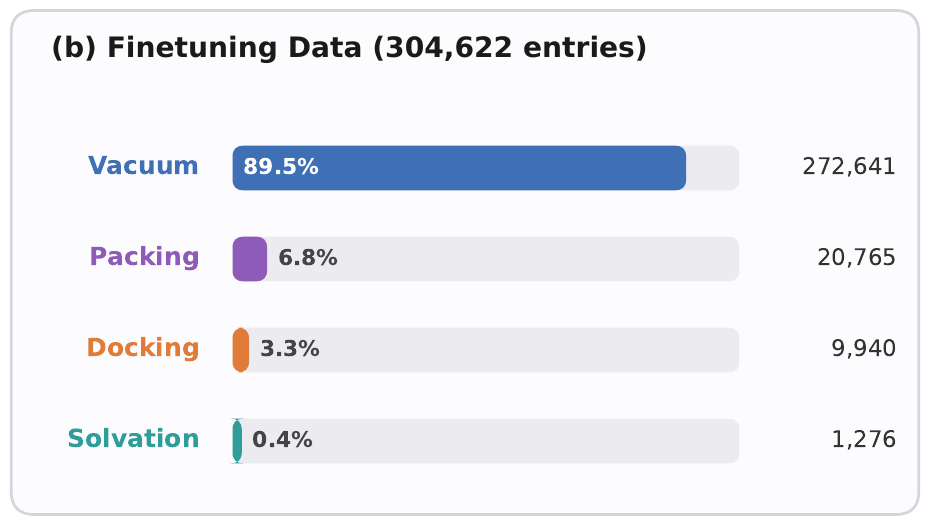}

\vspace{0.4em}

\includegraphics[width=0.48\linewidth]{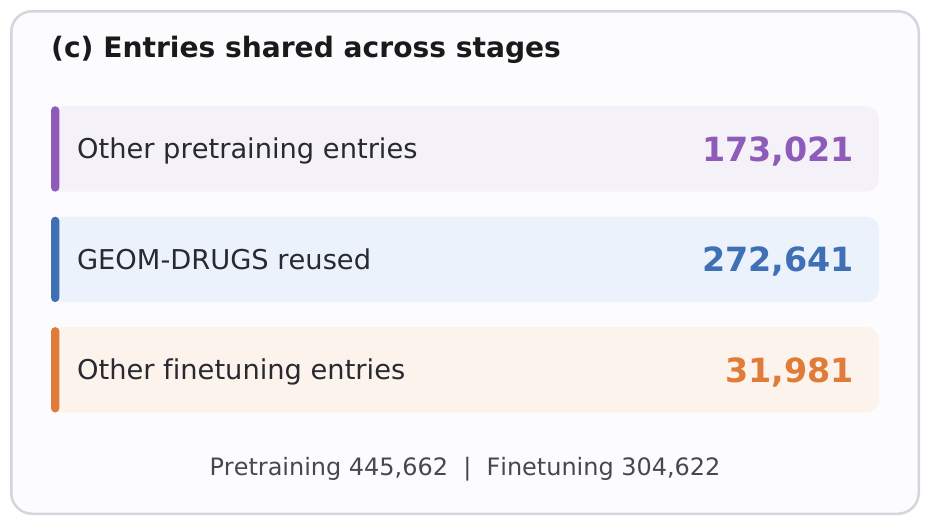}\hfill
\includegraphics[width=0.48\linewidth]{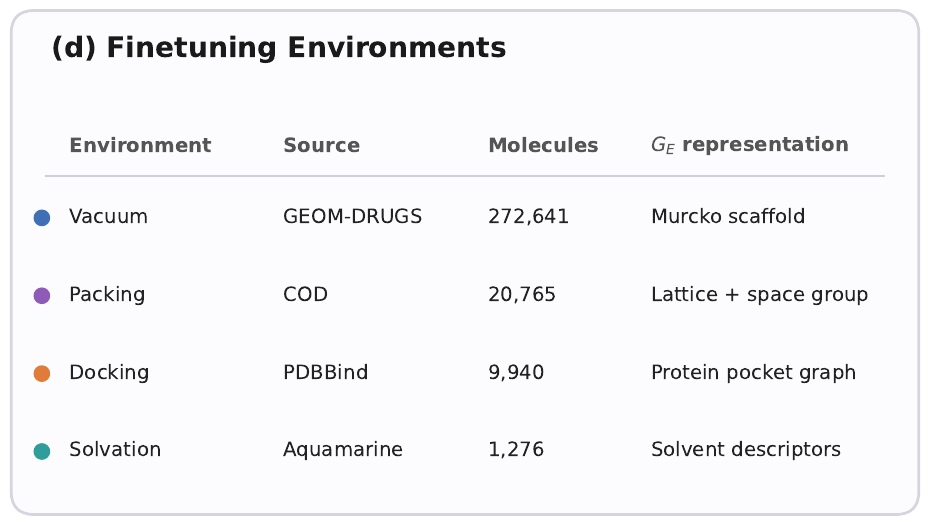}

\caption{Dataset composition. \textbf{(a)} Pretraining corpus by source ($445{,}662$ molecule entries), counting SPICE and CREST-relaxed SPICE separately. \textbf{(b)} Finetuning corpus by environment ($304{,}622$ molecule entries). \textbf{(c)} Reuse of all $272{,}641$ GEOM-DRUGS entries across pretraining and Vacuum finetuning; the remaining counts denote entries in the other datasets. \textbf{(d)} Data source and environment graph $G_E$ representation for each finetuning environment.}
\label{fig:dataset}
\end{figure}

\subsection{Training Objectives}
\label{sec:loss}
We jointly optimize the encoder, environment embedding network, base distribution network, and FM decoder with four loss terms (Figure \ref{fig:model}):
\begin{enumerate}[leftmargin=*]
    \item \textbf{Latent Prior Regularization}. This VAE loss regularizes the latent posterior toward a standard normal distribution:
    \begin{equation}
        \mathcal{L}_\mathrm{KL-z} = D_\mathrm{KL}(q_\phi(z|x,G,e)\,||\,\mathcal{N}(\mathbf{0}, I)).
    \end{equation}

    \item \textbf{Base Distribution Regularization}.
    The learned base distribution is regularized by KL divergence toward a simple Harmonic prior induced by graph topology $L$:
    \begin{equation}
    \mathcal{L}_\mathrm{base} = D_\mathrm{KL}\big( \mathcal{N}(\mu_\mathrm{base},\Sigma_\mathrm{base}) \;\|\; \mathcal{N}(\mathbf{0}, L^{\dagger}) \big).
    \end{equation}

    \item \textbf{Latent Consistency Loss}. This loss penalizes disagreement between the encoder's posterior mean given target conformation $x$ and initial conformation $x_0 \sim p'_0(x_0)$:
    \begin{equation}
        \mathcal{L}_\mathrm{consistency} = ||\mu_\phi(x_0,G,e)-\mu_\phi(x,G,e)||_2^2.
    \end{equation}

    \item \textbf{FM Loss}. This loss trains the decoder to transport samples from the base distribution to the data distribution through continuous-time dynamics.
    \begin{equation}
        \mathcal{L}_\mathrm{FM} = \mathbb{E}_{t,x_t}[||u_t^\theta(x_t,G,z,e)-u_t(x_0,x)||^2],
    \end{equation}
    where $u_t$ is the analytically defined target velocity field:
    \begin{equation}
    \begin{aligned}
        u_t &= (x_1 - x_0) + \dot\sigma(t)\,\epsilon, \quad \epsilon \sim \mathcal{N}(\mathbf{0}, I), \\
        \dot\sigma(t) &= \sigma \frac{1-2t}{2\sqrt{t(1-t)}}
    \end{aligned}
    \end{equation}
\end{enumerate}
All networks are trained jointly at each iteration.
We weight all loss terms equally:
\begin{equation}
    \mathcal{L} = \mathcal{L}_\mathrm{KL-z} + \mathcal{L}_\mathrm{KL-base} + \mathcal{L}_\mathrm{consistency} + \mathcal{L}_\mathrm{FM}.
\end{equation}

These terms regulate two train--inference transitions while preserving coordinate accuracy. $\mathcal{L}_\mathrm{FM}$ is the primary geometric objective and learns the coordinate transport. $\mathcal{L}_\mathrm{KL-z}$ makes sampling $z\sim\mathcal{N}(\mathbf{0}, I)$ valid at inference; $\mathcal{L}_\mathrm{KL-base}$ similarly aligns the learned training-time base with the graph-only harmonic prior used at inference. The consistency term prevents the latent from encoding incidental coordinate-level differences by requiring the same molecule--environment state to have a stable representation at $x$ and $x_0$. Thus, the KL and consistency terms constrain information pathways that could otherwise become training-time shortcuts, while the FM term retains the information required for accurate generation.

\section{Experiments}
\label{sec:experiments}

\subsection{Datasets}
\label{sec:dataset}

\textbf{Pretraining}.
For large-scale pretraining, we combine datasets containing conformations of small organic and drug-like molecules. 
This includes relaxed conformations from GEOM-QM9 and GEOM-DRUGS~\citep{axelrod2022geom,D5DD00206K}, as well as higher-energy conformations from the SPICE PubChem subset~\citep{eastman2023spice}. 
We also relax SPICE conformations using \texttt{CREST}~\citep{pracht2024crest} to obtain low-energy conformations.
We include the OE62 dataset~\citep{stuke2020atomic}, which consists of crystal-forming molecules relaxed in vacuum. 
The pretraining corpus contains $445{,}662$ molecule entries across five datasets and 9.8M conformations (Figure \ref{fig:dataset}a). SPICE and CREST-relaxed SPICE each contribute $19{,}198$ entries for the same molecules.

During pretraining, we construct $G_\mathrm{E}$ by uniformly sampling a mask ratio between $0.3$ and $0.7$ and retaining the corresponding atom-induced subgraph of the target molecule, including its covalent edges and coordinates. This denoising-style context exposes the shared backbone to diverse partial molecular environments. During Vacuum finetuning and evaluation, we instead apply the RDKit~\citep{RDKit} \texttt{MurckoScaffold} package~\citep{RDKitMurckoScaffoldDocs} to extract the molecule's Bemis--Murcko scaffold~\citep{bemis1996properties}; scaffold coordinates are set to zero so that the context encodes topology rather than the target conformation.

\textbf{Finetuning}.
We finetune our model in four environments:
\begin{enumerate}[leftmargin=*]
\item \textbf{Vacuum}. Molecules in vacuum are isolated. We use the GEOM-DRUGS dataset and construct $G_\mathrm{E}$ from the molecule's Bemis--Murcko scaffold as described above (Figure \ref{fig:dataset}(d)).

\item \textbf{Protein--Ligand Docking}. In this task, molecules (ligands) adapt their internal conformation to a known protein binding pocket. We use the PDBBind-v2020 dataset~\citep{Liu2015PDBbind}. The environment graph is constructed using heavy atoms in the extracted protein pocket; edges join atoms in the same residue, and pocket coordinates are included (Figure \ref{fig:dataset}(d)). We center the ligand and predict its pocket-conditioned internal conformation; unlike end-to-end docking systems, EVA-Flow does not predict the ligand's global translation or orientation in the protein.

\item \textbf{Solvation}. Dissolved molecules interact with the solvent. We use the Aquamarine dataset~\citep{medrano2024dataset}, which includes conformations relaxed in implicit water. 
To construct the environment graph, we use a water molecule and add solvation descriptors to the node features such as cavity surface area (sCAV), cavity volume (vCAV), free energy in electrolyte (eSOLV), and non-electrostatic free energy (eNELEC) (Figure \ref{fig:dataset}(d)). These descriptors represent the interaction between molecules and solvent.

\item \textbf{Crystal Packing}. Molecules interact with neighboring molecules in crystals. We use the Crystallography Open Database (COD)~\citep{gravzulis2009crystallography, gravzulis2012crystallography} and retain only small organic molecular crystals. 
We convert the fractional coordinates from the CIF files to Cartesian positions, and recover the bonds with \texttt{OpenBabel}~\citep{o2008pybel,o2011open}.
For the environment graph, we use a copy of the molecular graph with an RDKit proxy geometry and append the space-group number (sg) and lattice parameters $(a, b, c, \alpha, \beta, \gamma)$ to its features (Figure \ref{fig:dataset}(d)).
\end{enumerate}

Sections \ref{appendix:data} and \ref{appendix:distribution} describe data processing and the distributions of atom and rotatable-bond counts.

\begin{figure}[t]
    \centering
    \begin{subfigure}[t]{0.48\linewidth}
        \centering
        \caption{Vacuum}
        \includegraphics[width=\linewidth]{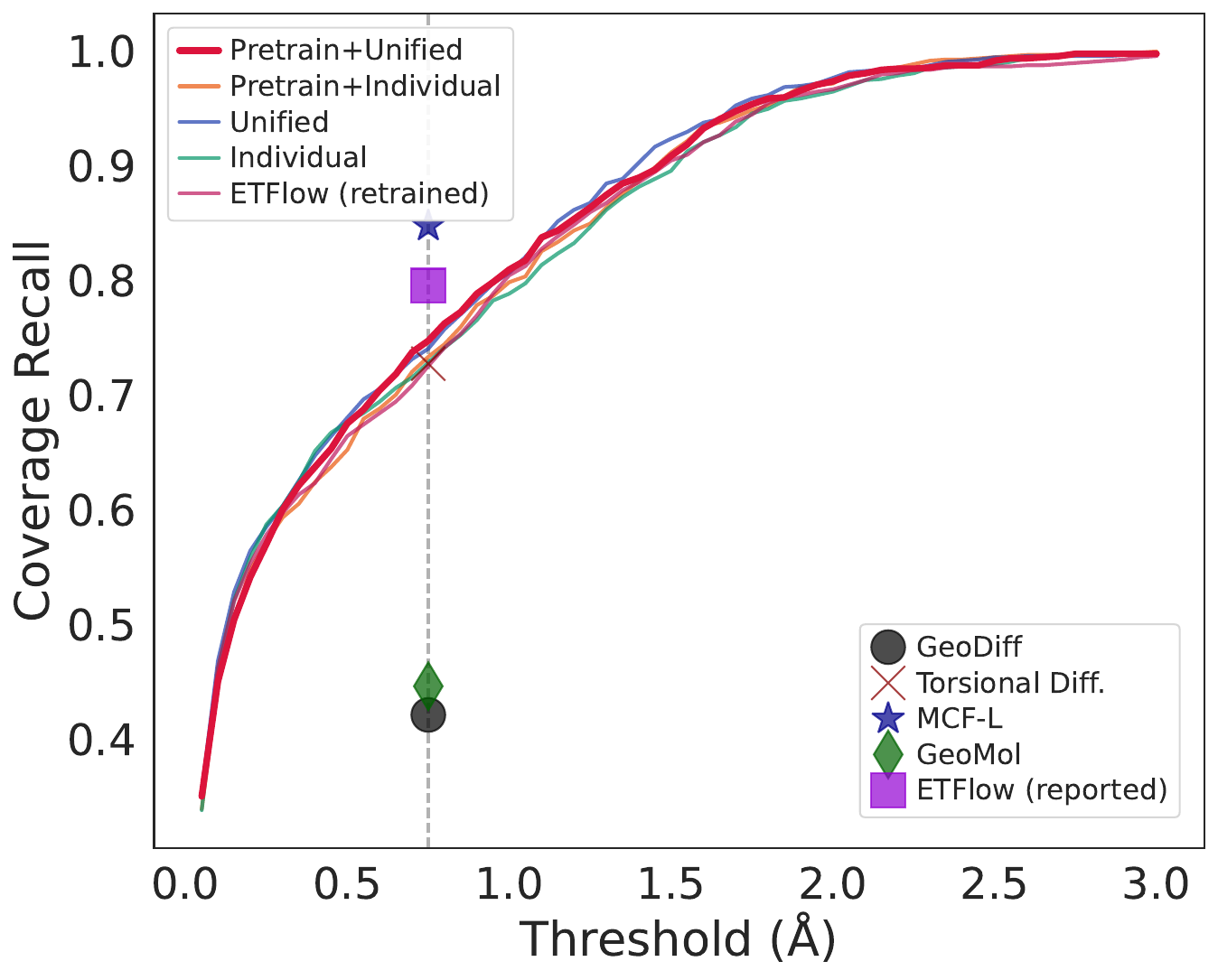}
        \label{fig:vac}
    \end{subfigure}
    \hfill
    \begin{subfigure}[t]{0.48\linewidth}
        \centering
        \caption{Docking}
        \includegraphics[width=\linewidth]{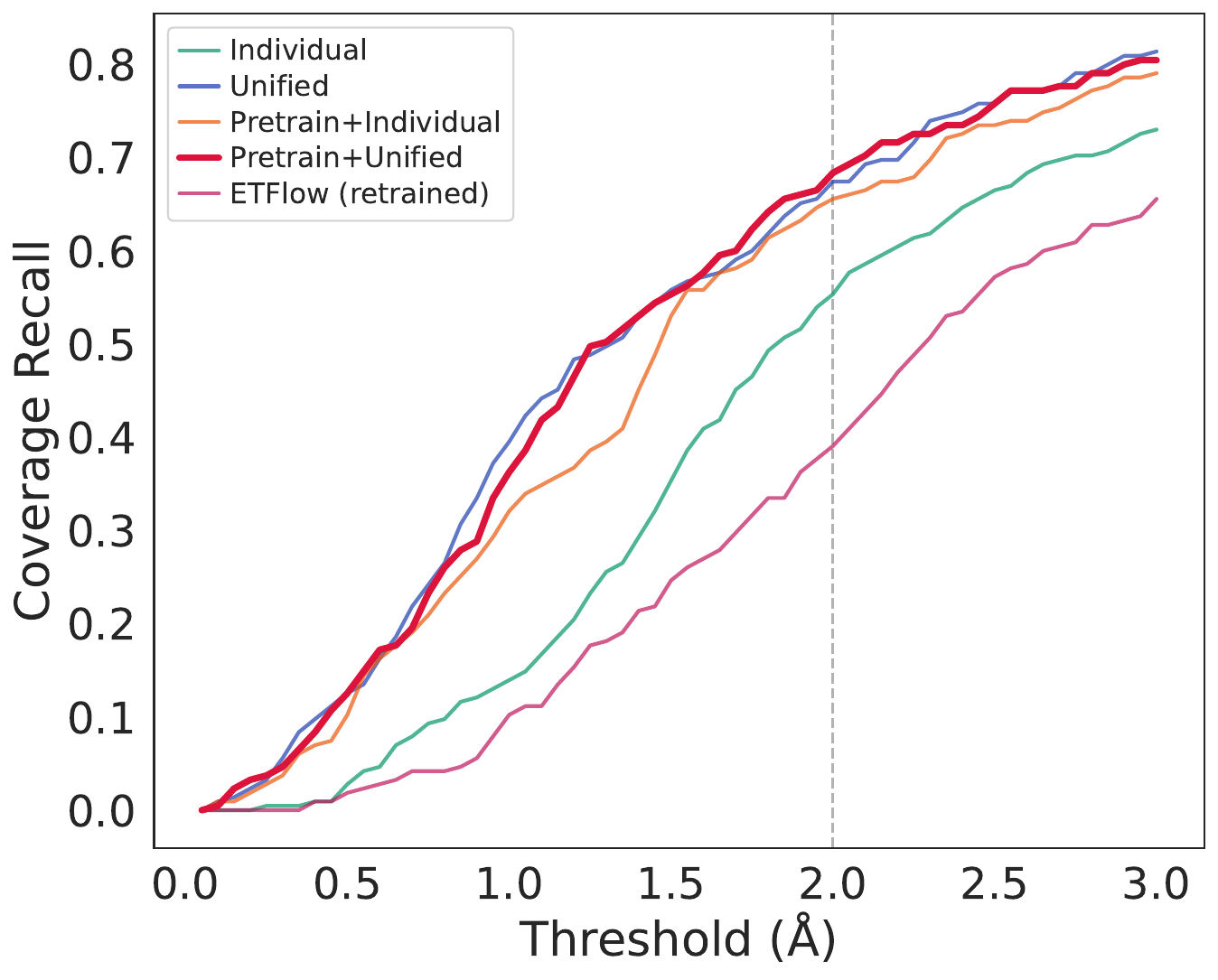}
        \label{fig:dock}
    \end{subfigure}

    \vspace{0.6em}

    \begin{subfigure}[t]{0.48\linewidth}
        \centering
        \caption{Solvation}
        \includegraphics[width=\linewidth]{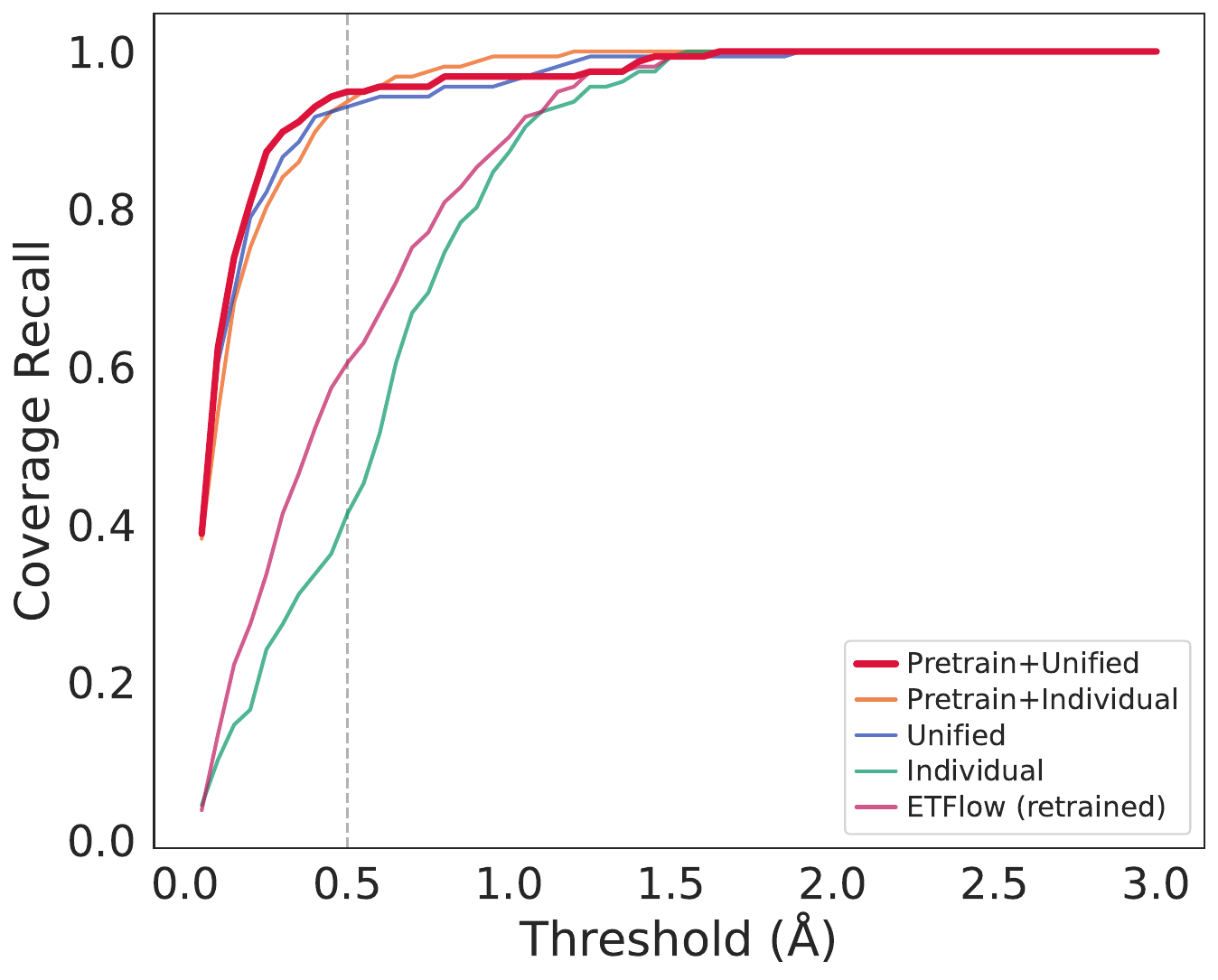}
        \label{fig:solv}
    \end{subfigure}
    \hfill
    \begin{subfigure}[t]{0.48\linewidth}
        \centering
        \caption{Packing}
        \includegraphics[width=\linewidth]{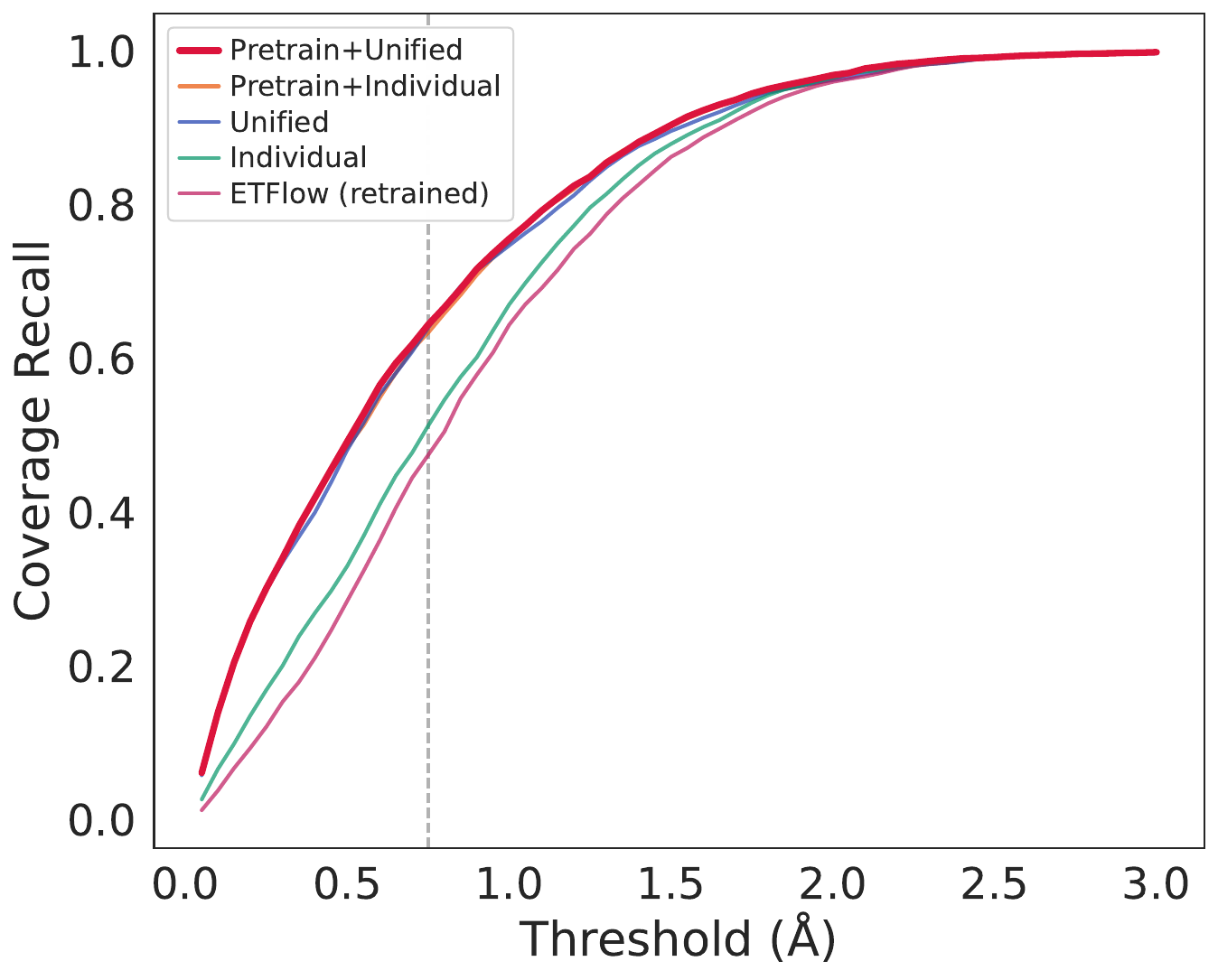}
        \label{fig:pack}
    \end{subfigure}

    \caption{Recall COV as a function of the RMSD threshold across the four environments. In every panel the four EVA-Flow training setups (Individual, Unified, Pretrain+Individual, Pretrain+Unified) are compared to retrained ETFlow. \textbf{(a)} Vacuum (cleaned GEOM-DRUGS~\citep{D5DD00206K}); reference points denote published Recall COV from \citet{xu2022geodiff, ganea2021geomol, jing2022torsional, wang2024swallowing, hassan2024etflow}. \textbf{(b)} Docking (PDBBind), \textbf{(c)} Solvation (Aquamarine), and \textbf{(d)} Packing (COD) compare models trained and evaluated under the same centered-conformation protocol. Dashed vertical lines mark the environment-specific RMSD thresholds ($0.75/2.0/0.5/0.75\,\text{\AA}$).}
\label{fig:results}
\end{figure}

\subsection{Experimental Setup}
We compare four training strategies.
\begin{enumerate}[leftmargin=*]
    \item \textbf{Pretraining + Individual Finetuning}. We pretrain the model on the pretraining datasets to learn shared chemistry from many molecules and conformations, then finetune it on a single environment.
    \item \textbf{Pretraining + Unified Finetuning}. We pretrain the model on the datasets in Section \ref{sec:dataset}, then finetune it on all finetuning environments. This setup tests adaptation across environments.
    \item \textbf{Unified Finetuning}. We train the model across all environments without pretraining.
    \item \textbf{Individual Finetuning}. We train separate models for each environment without pretraining. This provides a baseline for assessing the effects of unification and pretraining.
\end{enumerate}

\textbf{Evaluation protocols}.
The available supervision differs by benchmark: Vacuum and Solvation provide multi-conformer references for ensemble evaluation, whereas Docking and Packing provide one physical reference and therefore test environment-conditioned structure prediction.
The \textbf{4-conformer protocol} is our default across all four environments: each model generates $4$ conformers per test molecule, evaluated against one reference (Table~\ref{tab:results}, Figure~\ref{fig:results}). We report mean COV and mean/median AMR.
The \textbf{2K protocol} generates $2K$ conformers for a molecule with $K$ references. We use it for Vacuum comparisons with published baselines~\citep{ganea2021geomol,jing2022torsional,wang2024swallowing,hassan2024etflow} (Table~\ref{tab:results-2k}) and multi-reference Solvation evaluation against all available AQM conformers (Appendix~\ref{appendix:solv-multiref}).

\textbf{Evaluation metrics}.
We evaluate generated conformers against ground-truth conformers using distance-based RMSD. We report the Recall and Precision Coverage (COV, $\uparrow$) and average minimum RMSD (AMR, $\downarrow$) (Section \ref{appendix:eval}).
For crystal packing, we also align the generated conformation with the ground-truth one and compute a symmetry-aware RMSD (sRMSD). This value reflects the minimal atomic displacement after accounting for lattice symmetry and periodicity.

\subsection{Results}

\begin{table*}[t]
\centering
\caption{Conformation generation under the \textbf{4-conformer protocol} (default) across four environments. Each model generates $4$ conformers per molecule, compared with one reference. We report Recall and Precision COV ($\uparrow$, mean) and AMR ($\downarrow$, mean and median) at environment-specific RMSD thresholds: $0.5~\text{\AA}$ for Solvation, $0.75~\text{\AA}$ for Vacuum and Packing, and $2.0~\text{\AA}$ for Docking. For Packing, we additionally report symmetry-aware RMSD (sRMSD, $\downarrow$, mean). Except for sRMSD, metric cells are mean $\pm$ standard deviation over five independent SDE sampling seeds from one fixed trained checkpoint (Appendix~\ref{appendix:impl}). Best means within each environment are \textbf{bolded}.}
\label{tab:results}
\begingroup
\small
\setlength{\tabcolsep}{4pt}
\begin{tabular}{@{}llccccccc@{}}
\toprule
\multirow{3}{*}{Environment} & \multirow{3}{*}{Setup}
& \multicolumn{3}{c}{Recall} & \multicolumn{3}{c}{Precision} & \multirow{3}{*}{sRMSD$\downarrow$} \\
\cmidrule(lr){3-5} \cmidrule(lr){6-8}
& & COV$\uparrow$ & \multicolumn{2}{c}{AMR$\downarrow$}
  & COV$\uparrow$ & \multicolumn{2}{c}{AMR$\downarrow$} & \\
\cmidrule(lr){3-5} \cmidrule(lr){6-8}
& & mean & mean & median & mean & mean & median & \\
\midrule
\multirow{5}{*}{Vacuum}
& Pretrain+Unified & \textbf{\meanstd{0.747}{0.013}} & \meanstd{0.468}{0.012} & \meanstd{0.144}{0.024} & \meanstd{0.463}{0.004} & \meanstd{0.969}{0.006} & \meanstd{0.900}{0.023} & -- \\[0.15em]
& Pretrain+Individual & \meanstd{0.733}{0.010} & \meanstd{0.481}{0.012} & \meanstd{0.146}{0.019} & \meanstd{0.454}{0.004} & \meanstd{0.980}{0.008} & \meanstd{0.925}{0.016} & -- \\[0.15em]
& Unified & \meanstd{0.740}{0.007} & \textbf{\meanstd{0.456}{0.008}} & \meanstd{0.131}{0.011} & \meanstd{0.463}{0.003} & \textbf{\meanstd{0.954}{0.005}} & \textbf{\meanstd{0.879}{0.011}} & -- \\[0.15em]
& Individual & \meanstd{0.728}{0.005} & \meanstd{0.487}{0.004} & \meanstd{0.134}{0.006} & \meanstd{0.460}{0.003} & \meanstd{0.978}{0.004} & \meanstd{0.911}{0.017} & -- \\[0.15em]
& ETFlow (retrained) & \meanstd{0.725}{0.012} & \meanstd{0.488}{0.011} & \textbf{\meanstd{0.130}{0.013}} & \textbf{\meanstd{0.467}{0.007}} & \meanstd{0.972}{0.013} & \meanstd{0.919}{0.022} & -- \\[0.15em]
\midrule
\multirow{5}{*}{Docking}
& Pretrain+Unified & \textbf{\meanstd{0.684}{0.013}} & \meanstd{1.899}{0.027} & \textbf{\meanstd{1.257}{0.037}} & \meanstd{0.512}{0.008} & \meanstd{2.471}{0.023} & \meanstd{1.864}{0.039} & -- \\[0.15em]
& Pretrain+Individual & \meanstd{0.656}{0.010} & \meanstd{1.980}{0.013} & \meanstd{1.463}{0.021} & \meanstd{0.458}{0.006} & \meanstd{2.581}{0.019} & \meanstd{2.076}{0.034} & -- \\[0.15em]
& Unified & \meanstd{0.674}{0.021} & \textbf{\meanstd{1.889}{0.016}} & \meanstd{1.303}{0.060} & \textbf{\meanstd{0.536}{0.012}} & \textbf{\meanstd{2.391}{0.021}} & \textbf{\meanstd{1.732}{0.023}} & -- \\[0.15em]
& Individual & \meanstd{0.553}{0.013} & \meanstd{2.458}{0.034} & \meanstd{1.833}{0.039} & \meanstd{0.378}{0.009} & \meanstd{3.135}{0.032} & \meanstd{2.387}{0.075} & -- \\[0.15em]
& ETFlow (retrained) & \meanstd{0.391}{0.016} & \meanstd{2.768}{0.029} & \meanstd{2.288}{0.084} & \meanstd{0.242}{0.012} & \meanstd{3.580}{0.019} & \meanstd{2.958}{0.052} & -- \\[0.15em]
\midrule
\multirow{5}{*}{Solvation}
& Pretrain+Unified & \textbf{\meanstd{0.949}{0.019}} & \textbf{\meanstd{0.153}{0.013}} & \textbf{\meanstd{0.072}{0.016}} & \textbf{\meanstd{0.780}{0.009}} & \textbf{\meanstd{0.367}{0.007}} & \textbf{\meanstd{0.232}{0.013}} & -- \\[0.15em]
& Pretrain+Individual & \meanstd{0.936}{0.018} & \meanstd{0.157}{0.006} & \meanstd{0.087}{0.020} & \meanstd{0.693}{0.006} & \meanstd{0.431}{0.007} & \meanstd{0.334}{0.017} & -- \\[0.15em]
& Unified & \meanstd{0.930}{0.034} & \meanstd{0.166}{0.017} & \meanstd{0.083}{0.018} & \meanstd{0.769}{0.014} & \meanstd{0.370}{0.009} & \meanstd{0.249}{0.023} & -- \\[0.15em]
& Individual & \meanstd{0.414}{0.021} & \meanstd{0.580}{0.020} & \meanstd{0.568}{0.026} & \meanstd{0.177}{0.009} & \meanstd{0.961}{0.045} & \meanstd{0.851}{0.044} & -- \\[0.15em]
& ETFlow (retrained) & \meanstd{0.605}{0.034} & \meanstd{0.475}{0.025} & \meanstd{0.366}{0.035} & \meanstd{0.264}{0.013} & \meanstd{0.824}{0.015} & \meanstd{0.786}{0.028} & -- \\[0.15em]
\midrule
\multirow{5}{*}{Packing}
& Pretrain+Unified & \textbf{\meanstd{0.644}{0.004}} & \textbf{\meanstd{0.664}{0.002}} & \textbf{\meanstd{0.509}{0.005}} & \meanstd{0.410}{0.001} & \meanstd{1.073}{0.001} & \meanstd{1.008}{0.005} & 0.055 \\[0.15em]
& Pretrain+Individual & \meanstd{0.633}{0.005} & \meanstd{0.672}{0.003} & \meanstd{0.529}{0.006} & \meanstd{0.400}{0.002} & \meanstd{1.087}{0.001} & \meanstd{1.022}{0.005} & 0.055 \\[0.15em]
& Unified & \meanstd{0.640}{0.003} & \meanstd{0.682}{0.005} & \meanstd{0.527}{0.006} & \textbf{\meanstd{0.414}{0.002}} & \textbf{\meanstd{1.067}{0.002}} & \textbf{\meanstd{0.997}{0.008}} & \textbf{0.054} \\[0.15em]
& Individual & \meanstd{0.512}{0.006} & \meanstd{0.819}{0.004} & \meanstd{0.734}{0.008} & \meanstd{0.303}{0.002} & \meanstd{1.222}{0.004} & \meanstd{1.161}{0.004} & 0.064 \\[0.15em]
& ETFlow (retrained) & \meanstd{0.474}{0.003} & \meanstd{0.878}{0.005} & \meanstd{0.794}{0.006} & \meanstd{0.274}{0.002} & \meanstd{1.274}{0.006} & \meanstd{1.203}{0.009} & 0.069 \\[0.15em]
\bottomrule
\end{tabular}
\endgroup
\end{table*}

\begin{table*}[t]
\centering
\caption{Conformation generation for Vacuum under the \textbf{2K protocol}: each model generates $2K$ conformers per molecule, where $K$ is the number of reference conformers. We report mean and median Recall and Precision COV ($\uparrow$) and AMR ($\downarrow$) at RMSD threshold $0.75~\text{\AA}$. Best results are \textbf{bolded}. The models are trained on the original GEOM-DRUGS dataset and evaluated on its full test set. Literature numbers for GeoDiff, GeoMol, Torsional Diffusion, MCF-L, and ETFlow (reported) are taken from Table 1 of \citet{hassan2024etflow}; ETFlow (reported) refers to the ET-Flow base ($8.3$M) variant. ETFlow (retrained) and EVA-Flow are trained on all available conformers per molecule on the original GEOM-DRUGS and evaluated on the same test set.}
\label{tab:results-2k}
\resizebox{\linewidth}{!}{%
\begin{tabular}{@{}lcccccccc@{}}
\toprule
\multirow{3}{*}{Setup}
& \multicolumn{4}{c}{Recall} & \multicolumn{4}{c}{Precision} \\
\cmidrule(lr){2-5} \cmidrule(lr){6-9}
& \multicolumn{2}{c}{COV$\uparrow$} & \multicolumn{2}{c}{AMR$\downarrow$}
& \multicolumn{2}{c}{COV$\uparrow$} & \multicolumn{2}{c}{AMR$\downarrow$} \\
\cmidrule(lr){2-3} \cmidrule(lr){4-5} \cmidrule(lr){6-7} \cmidrule(lr){8-9}
& mean & median & mean & median & mean & median & mean & median \\
\midrule
GeoDiff & 0.421 & 0.378 & 0.835 & 0.809 & 0.249 & 0.145 & 1.136 & 1.090 \\
GeoMol & 0.446 & 0.414 & 0.875 & 0.834 & 0.430 & 0.364 & 0.928 & 0.841 \\
Torsional Diff. & 0.727 & 0.800 & 0.582 & 0.565 & 0.552 & 0.569 & 0.778 & 0.729 \\
MCF-L & \textbf{0.847} & \textbf{0.922} & \textbf{0.390} & \textbf{0.247} & 0.668 & 0.713 & 0.618 & 0.530 \\
ETFlow (reported) & 0.795 & 0.846 & 0.452 & 0.419 & \textbf{0.744} & \textbf{0.810} & \textbf{0.541} & \textbf{0.470} \\
\midrule
ETFlow (retrained) & 0.712 & 0.742 & 0.562 & 0.530 & 0.704 & 0.758 & 0.594 & 0.520 \\
EVA-Flow (Pretrain+Unified) & 0.721 & 0.756 & 0.550 & 0.527 & 0.707 & 0.781 & 0.589 & 0.510 \\
\bottomrule
\end{tabular}%
}
\end{table*}

Table~\ref{tab:results} and Figure~\ref{fig:results} summarize results under the default 4-conformer protocol. On Vacuum under the 2K protocol, EVA-Flow is comparable to retrained ETFlow and has lower Recall COV but higher Precision COV than MCF-L (Table~\ref{tab:results-2k}). Its accuracy gains are concentrated in the data-scarce Docking, Solvation, and Packing settings.

\textbf{Pretraining improves generalization.}
Pretraining provides gains over Individual training in the data-scarce environments (Table \ref{tab:results}, Figure \ref{fig:results}). Vacuum differences among EVA-Flow setups are small relative to the observed inference-sampler variability, consistent with its much larger training set.

\textbf{Unified training outperforms individual models.}
Unified setups substantially surpass Individual finetuning in Docking, Solvation, and Packing, whereas the differences within Vacuum are small. A data-volume control shows that the gains persist beyond raw target-environment exposure (Appendix~\ref{appendix:exposure}).

\textbf{Baseline comparisons.}
We compare with retrained ETFlow using matched task definitions and evaluation code. EVA-Flow also outperforms ETKDG+MMFF in all four environments and GFN2-xTB on Solvation, where both physical references collapse (Appendix~\ref{appendix:physical-baselines}). Baseline training details and comparison scope are provided in Appendix~\ref{appendix:etflow-reconciliation}.

\begin{table*}[t]
    \centering
    \caption{Reference and generated conformations (Pretrain+Unified) for molecules shared between Solvation and Packing or Solvation and Docking. The conformers are aligned to the Solvation reference.}
    \label{tab:vis}
    \begin{tabular}{@{}cccc|cccc@{}}
    \toprule
    \multicolumn{4}{c|}{Solvation-Packing} & \multicolumn{4}{c}{Solvation-Docking} \\
    \midrule
    ref\_solv & gen\_solv & ref\_pack & gen\_pack & ref\_solv & gen\_solv & ref\_dock & gen\_dock \\
    \midrule
    \includegraphics[trim={2cm 1cm 2cm 1cm},clip,height=1.2cm]{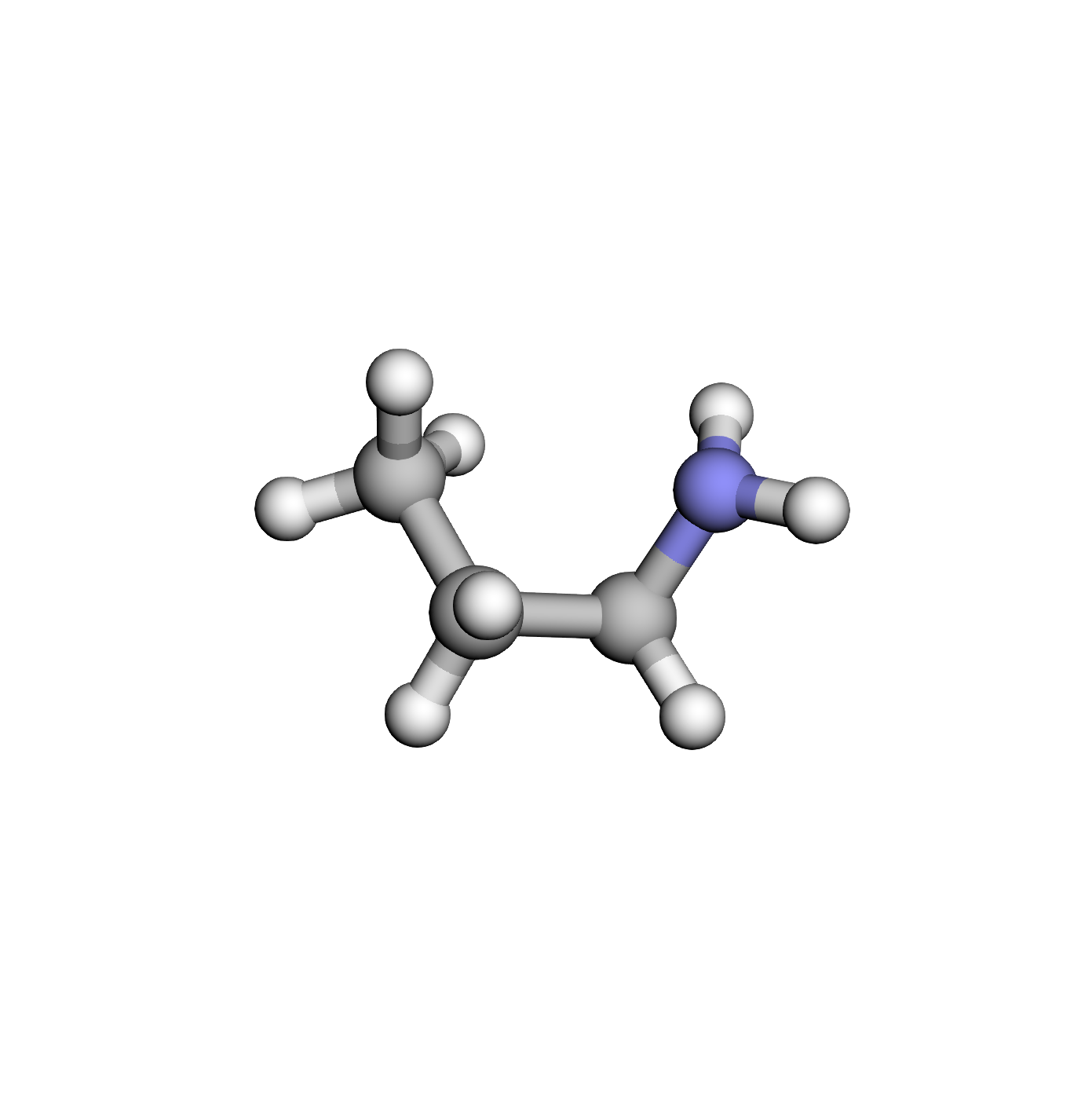} & \includegraphics[trim={2cm 1cm 2cm 1cm},clip,height=1.2cm]{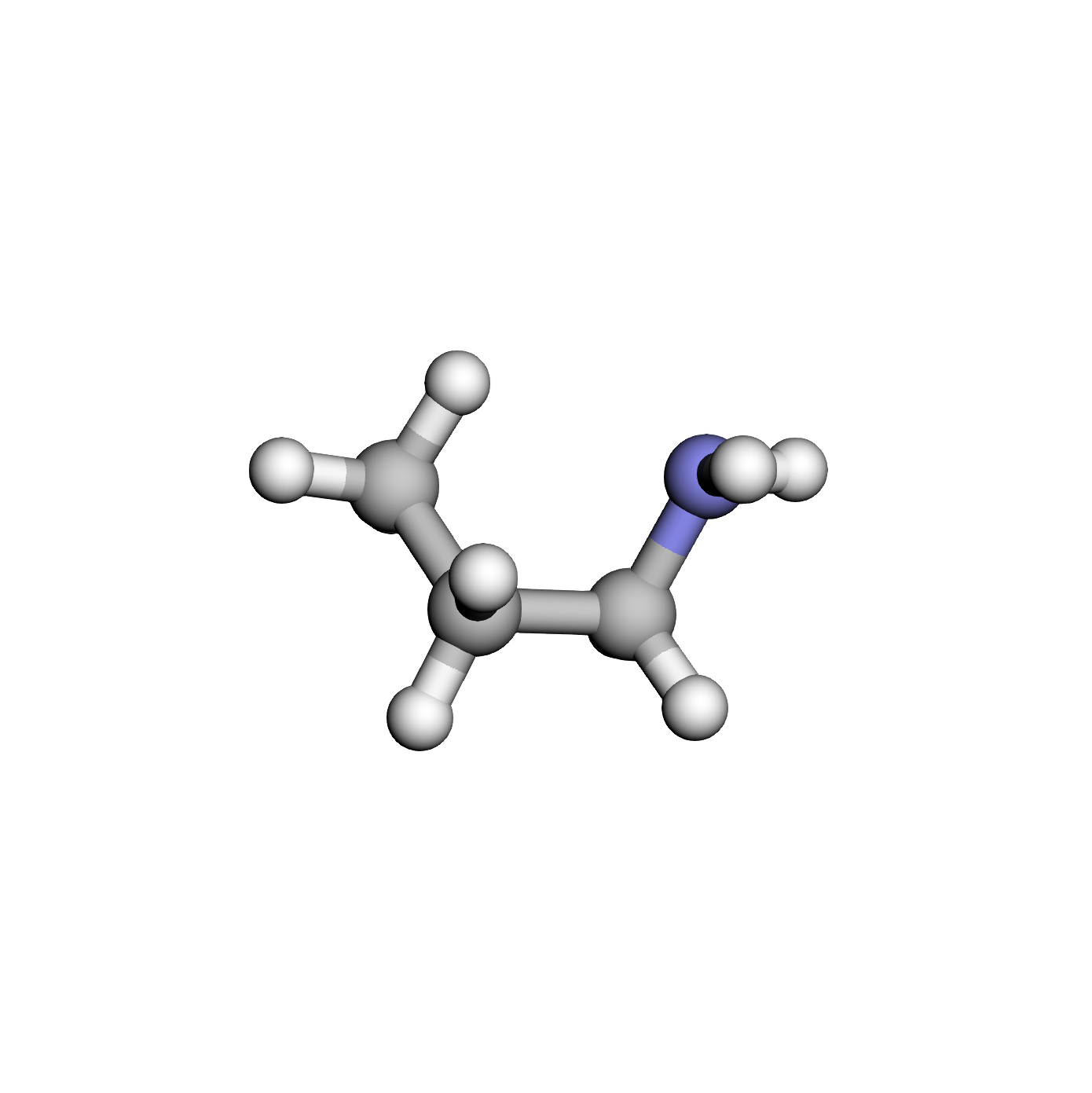} & \includegraphics[trim={2cm 1cm 2cm 1cm},clip,height=1.2cm]{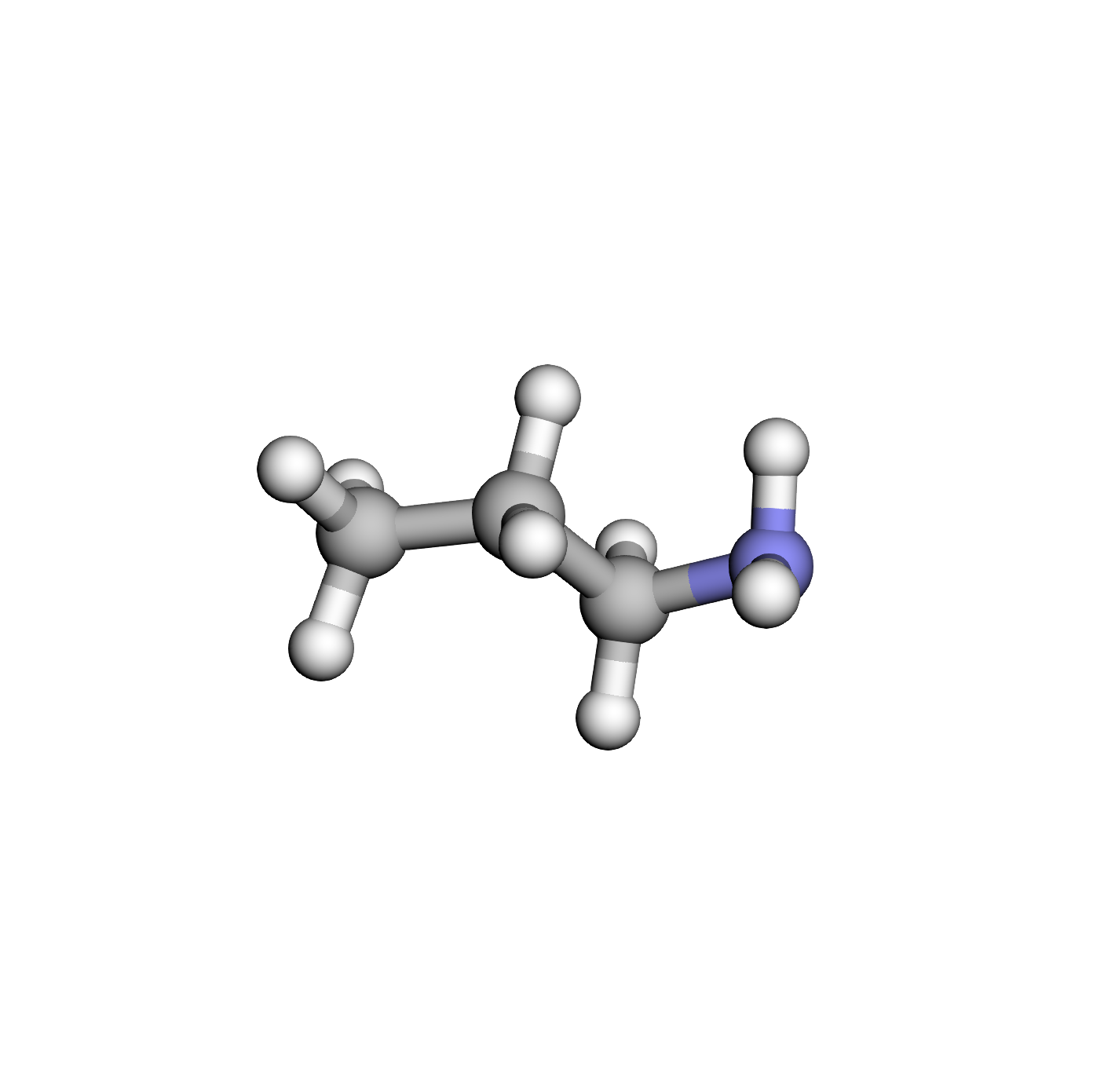} & \includegraphics[trim={2cm 1cm 2cm 1cm},clip,height=1.2cm]{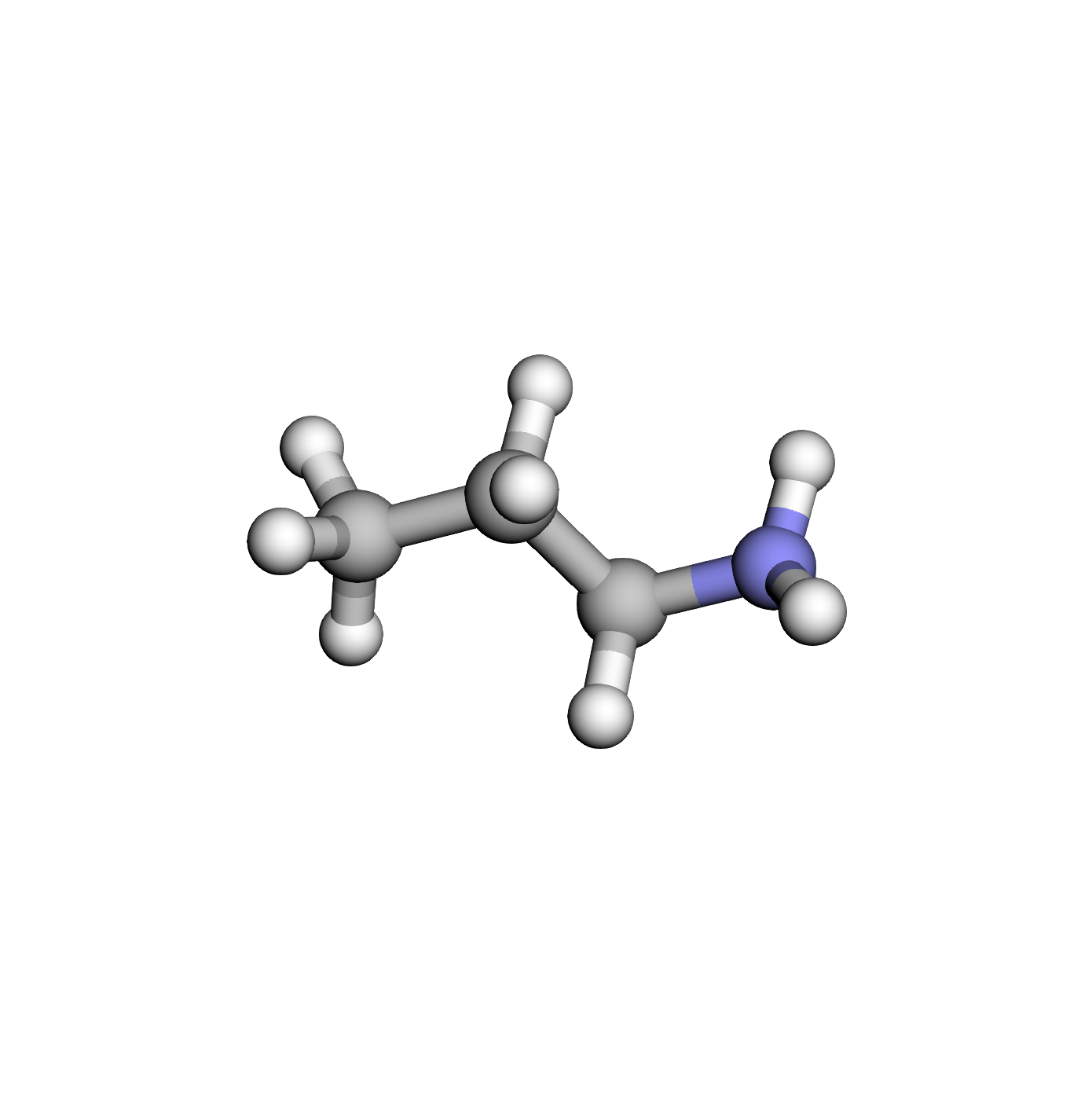} & \includegraphics[trim={2cm 1cm 2cm 1cm},clip,height=1.2cm]{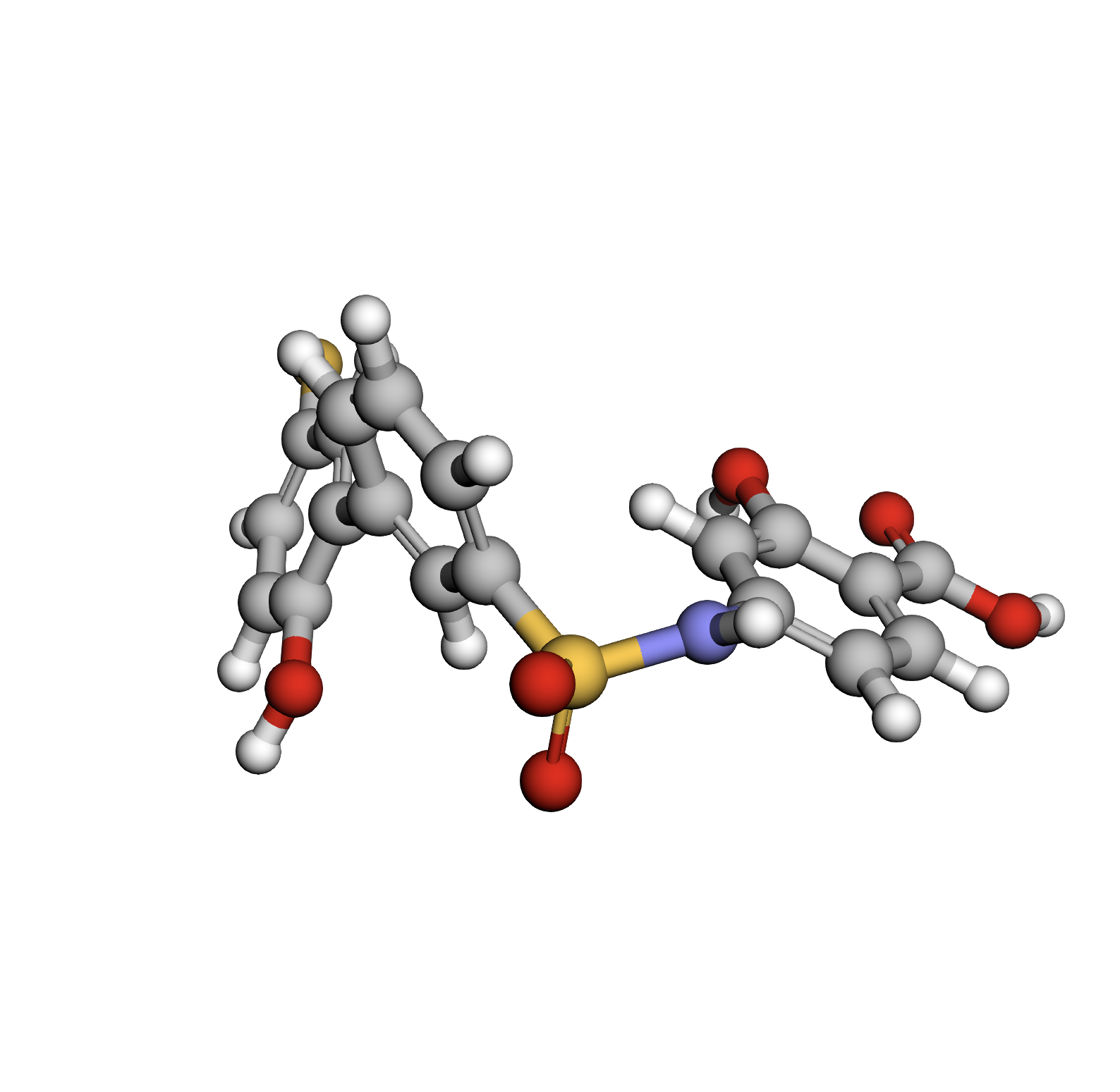} & \includegraphics[trim={2cm 1cm 2cm 1cm},clip,height=1.2cm]{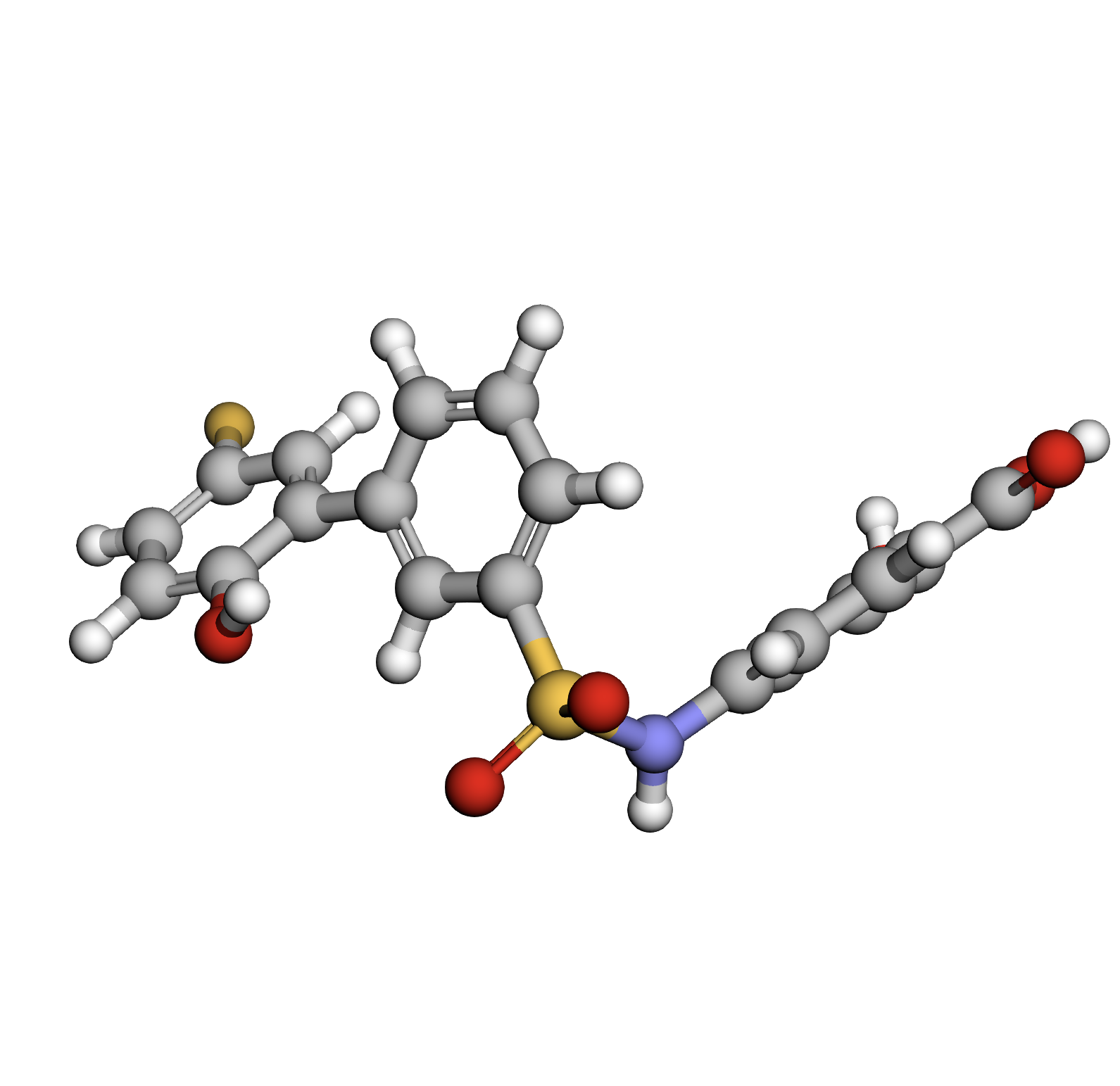} & \includegraphics[trim={2cm 1cm 2cm 1cm},clip,height=1.2cm]{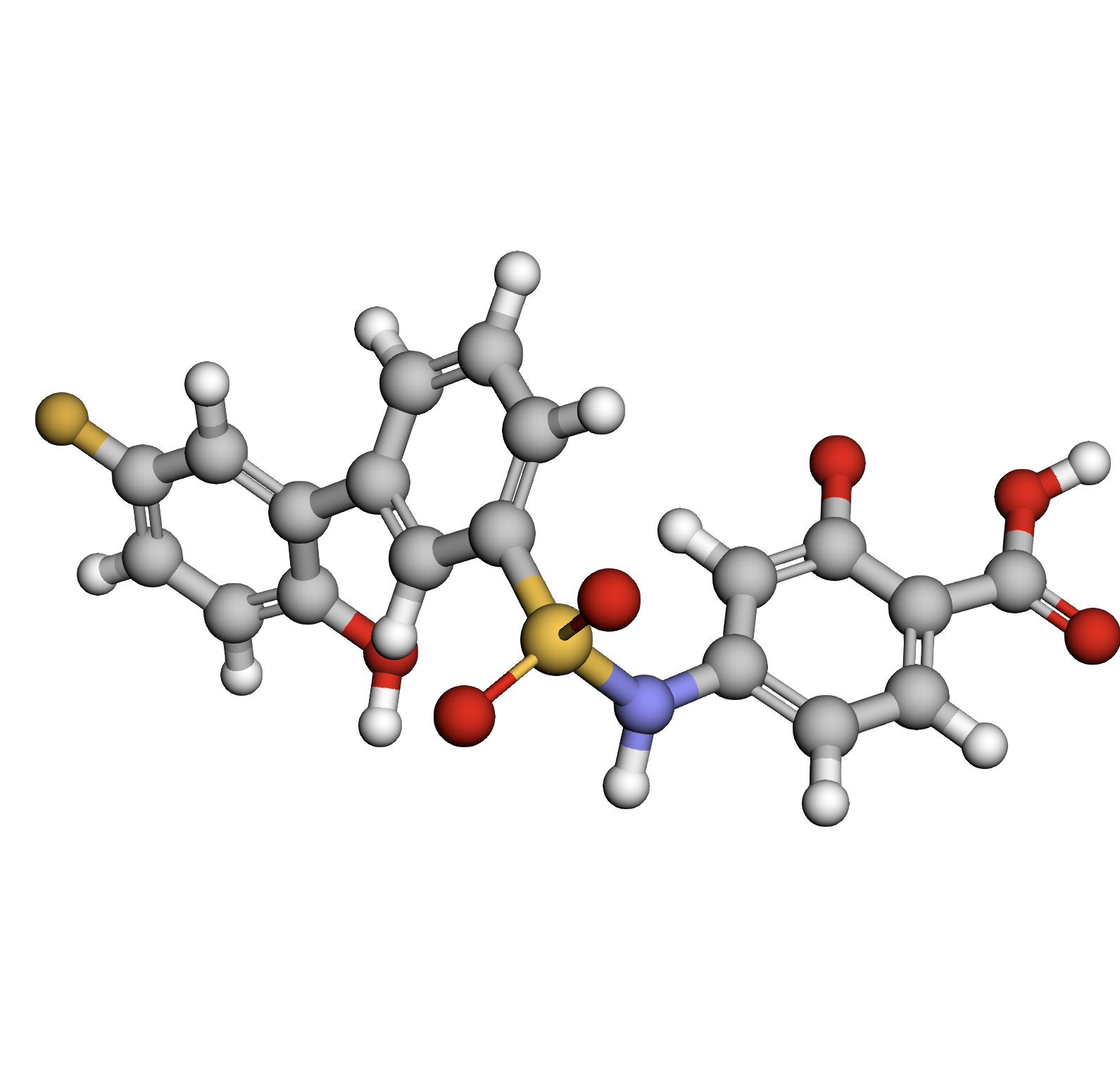} & \includegraphics[trim={2cm 1cm 2cm 1cm},clip,height=1.2cm]{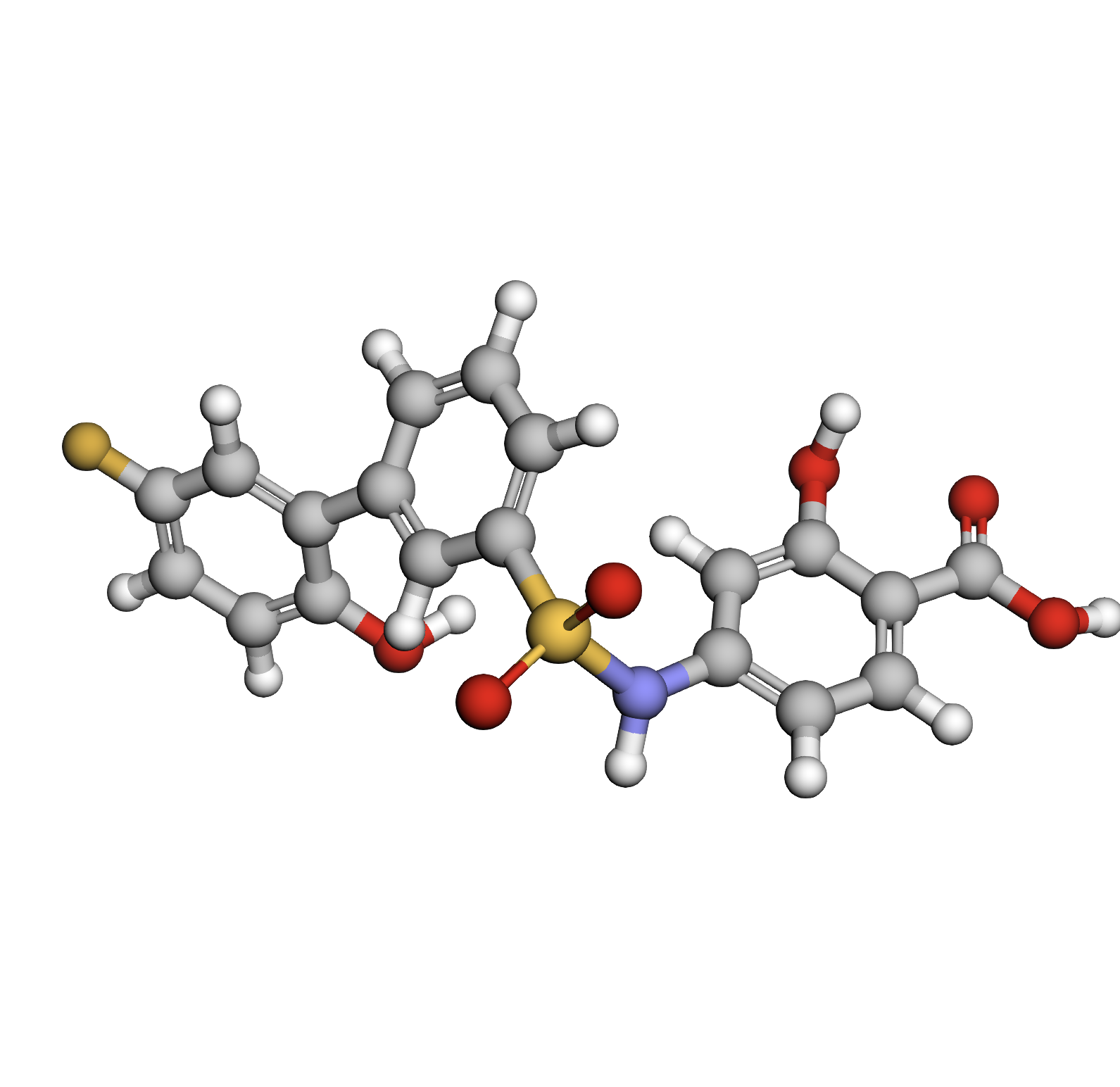} \\
    \includegraphics[trim={0cm 1cm 0cm 1cm},clip,height=1.2cm]{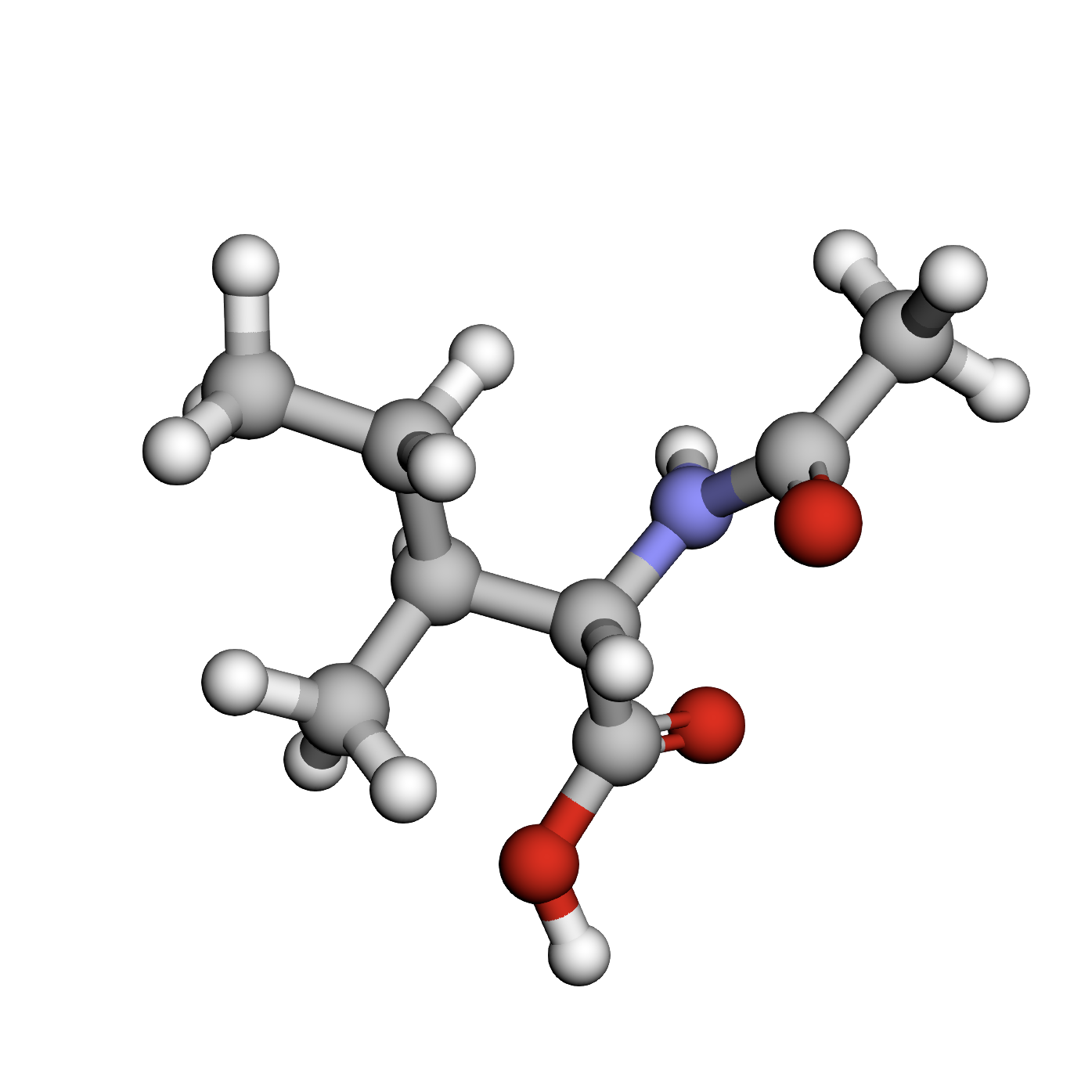} & \includegraphics[trim={0cm 1cm 0cm 1cm},clip,height=1.2cm]{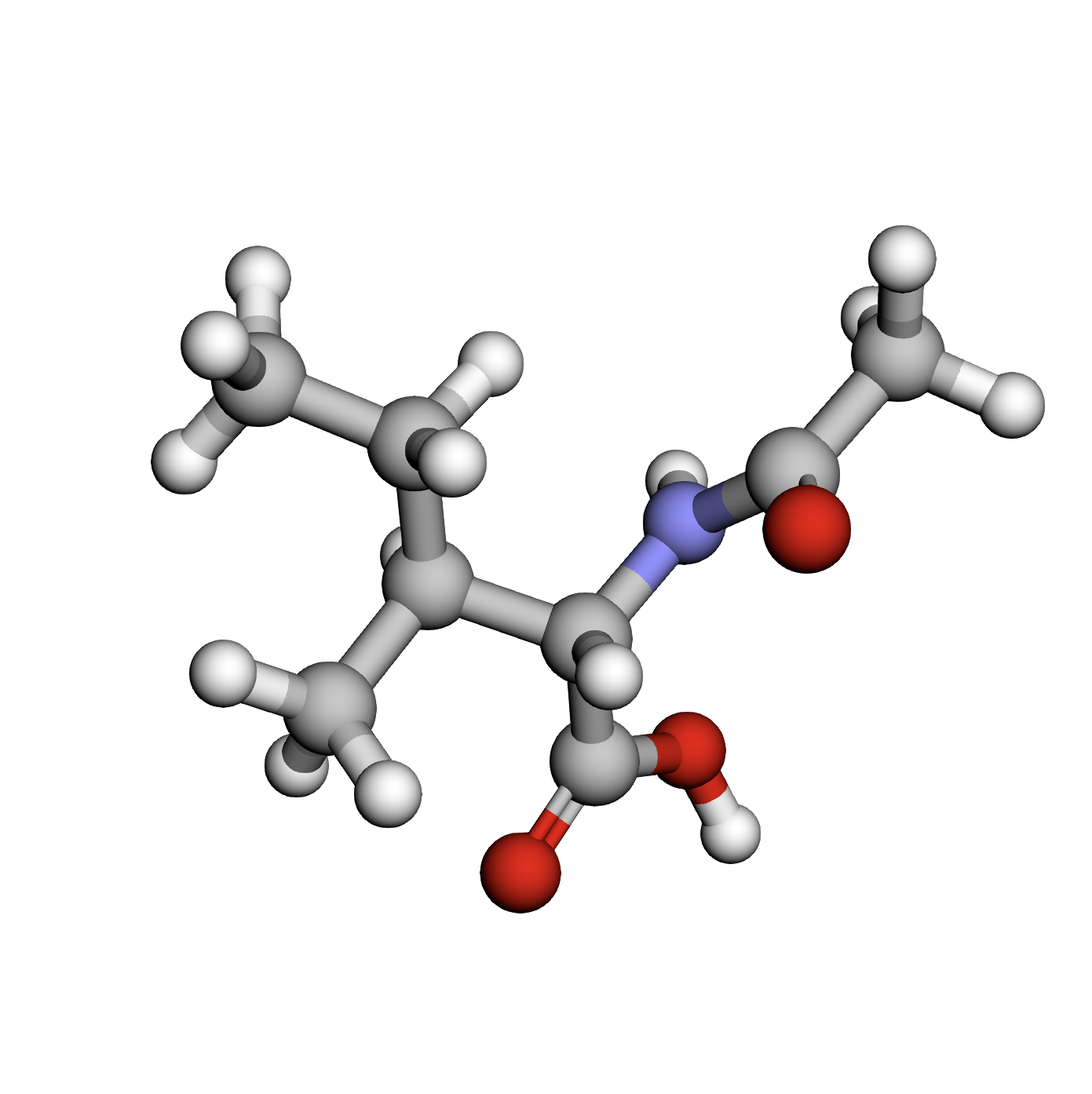} & \includegraphics[trim={0cm 1cm 0cm 1cm},clip,height=1.2cm]{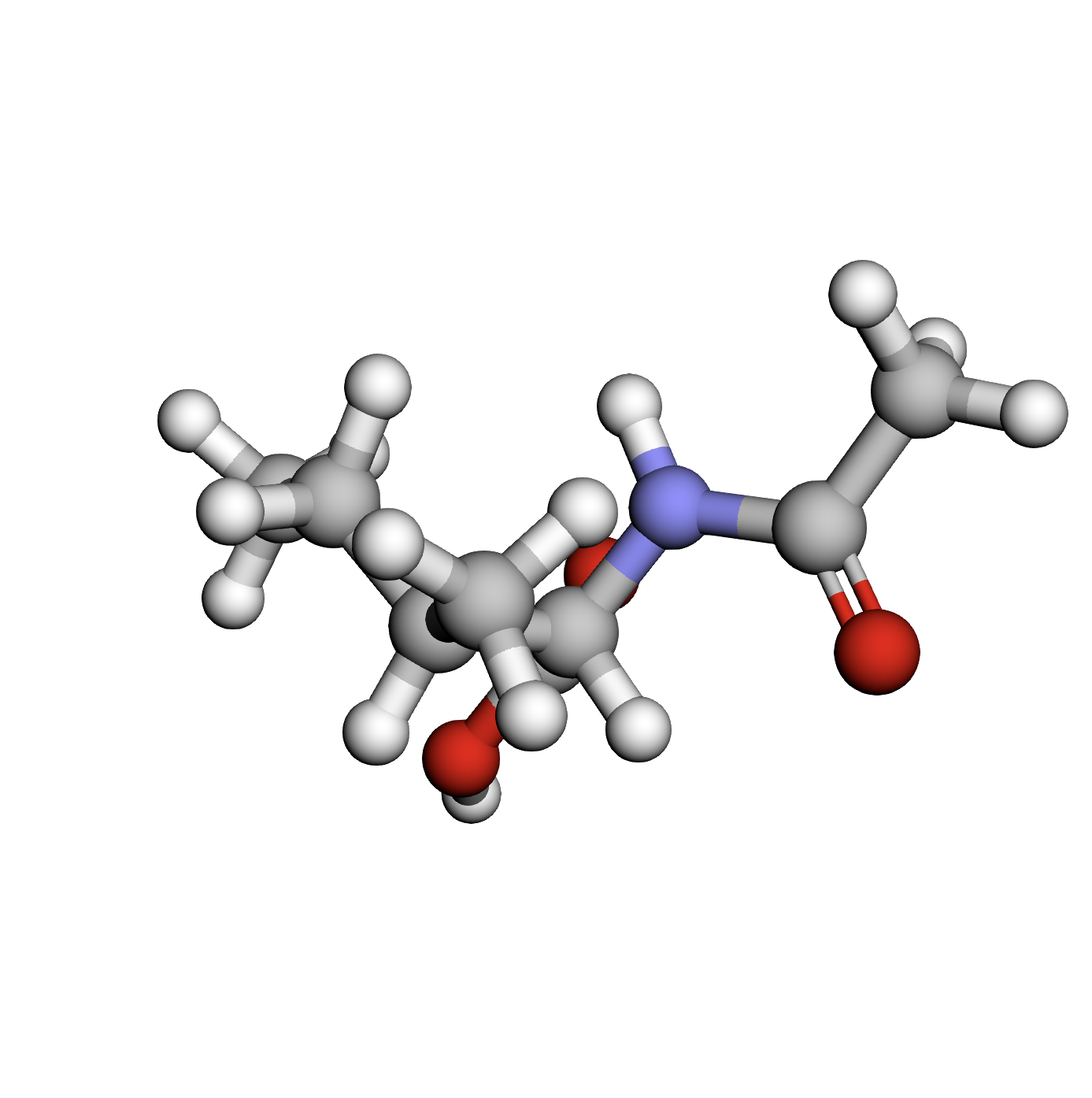} & \includegraphics[trim={0cm 1cm 0cm 1cm},clip,height=1.2cm]{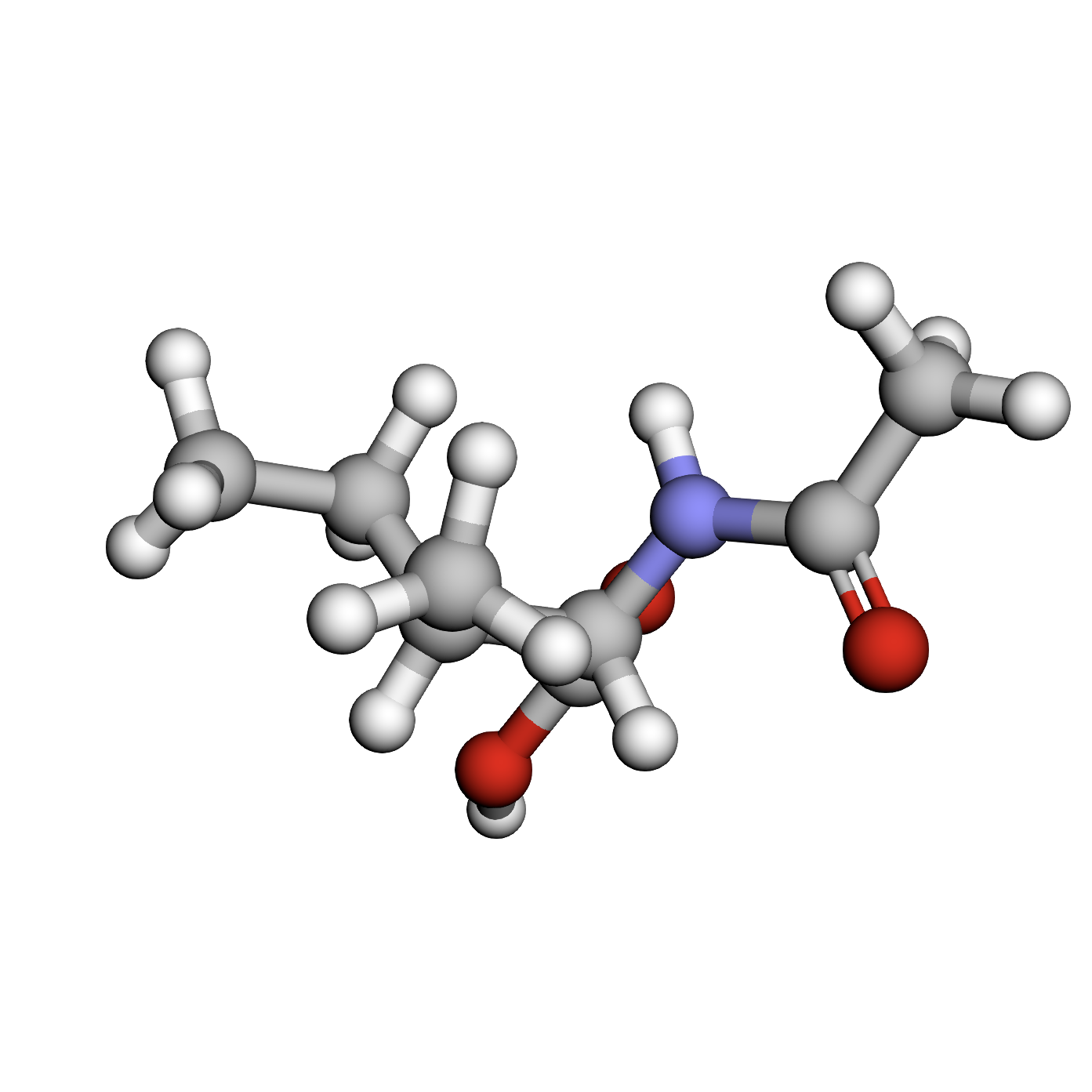} & \includegraphics[trim={0cm 1cm 0cm 1cm},clip,height=1.2cm]{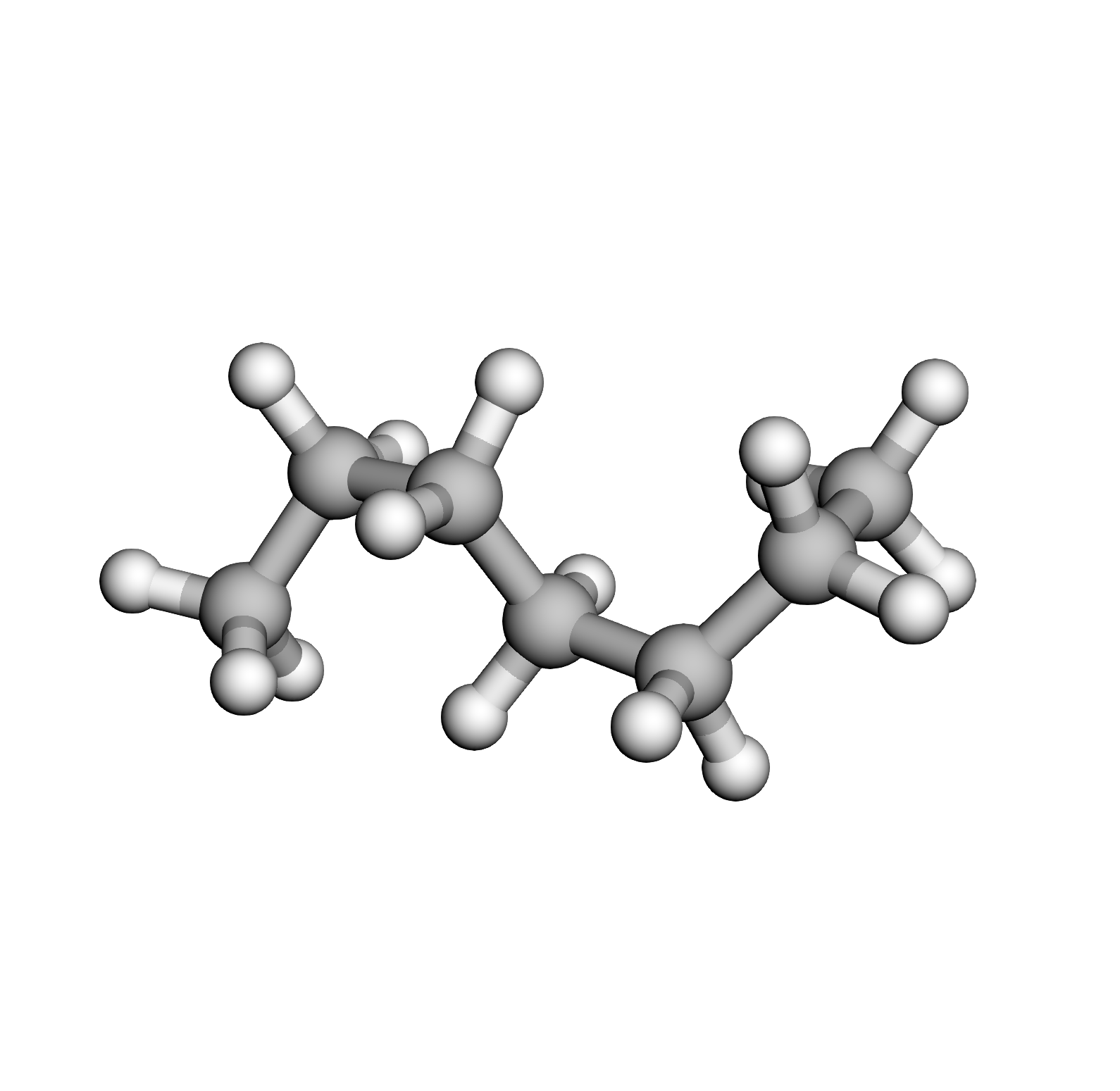} & \includegraphics[trim={0cm 1cm 0cm 1cm},clip,height=1.2cm]{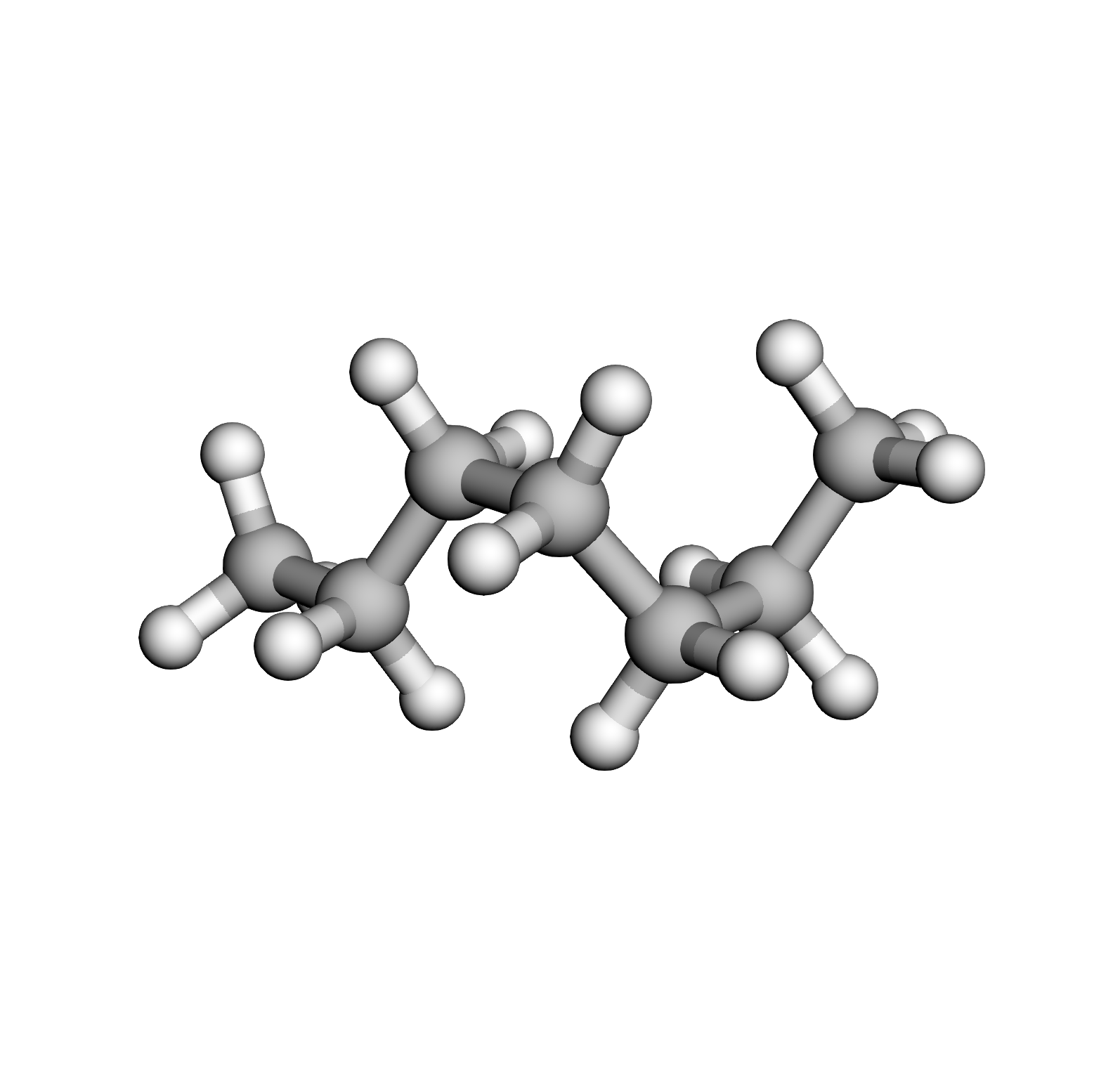} & \includegraphics[trim={0cm 1cm 0cm 1cm},clip,height=1.2cm]{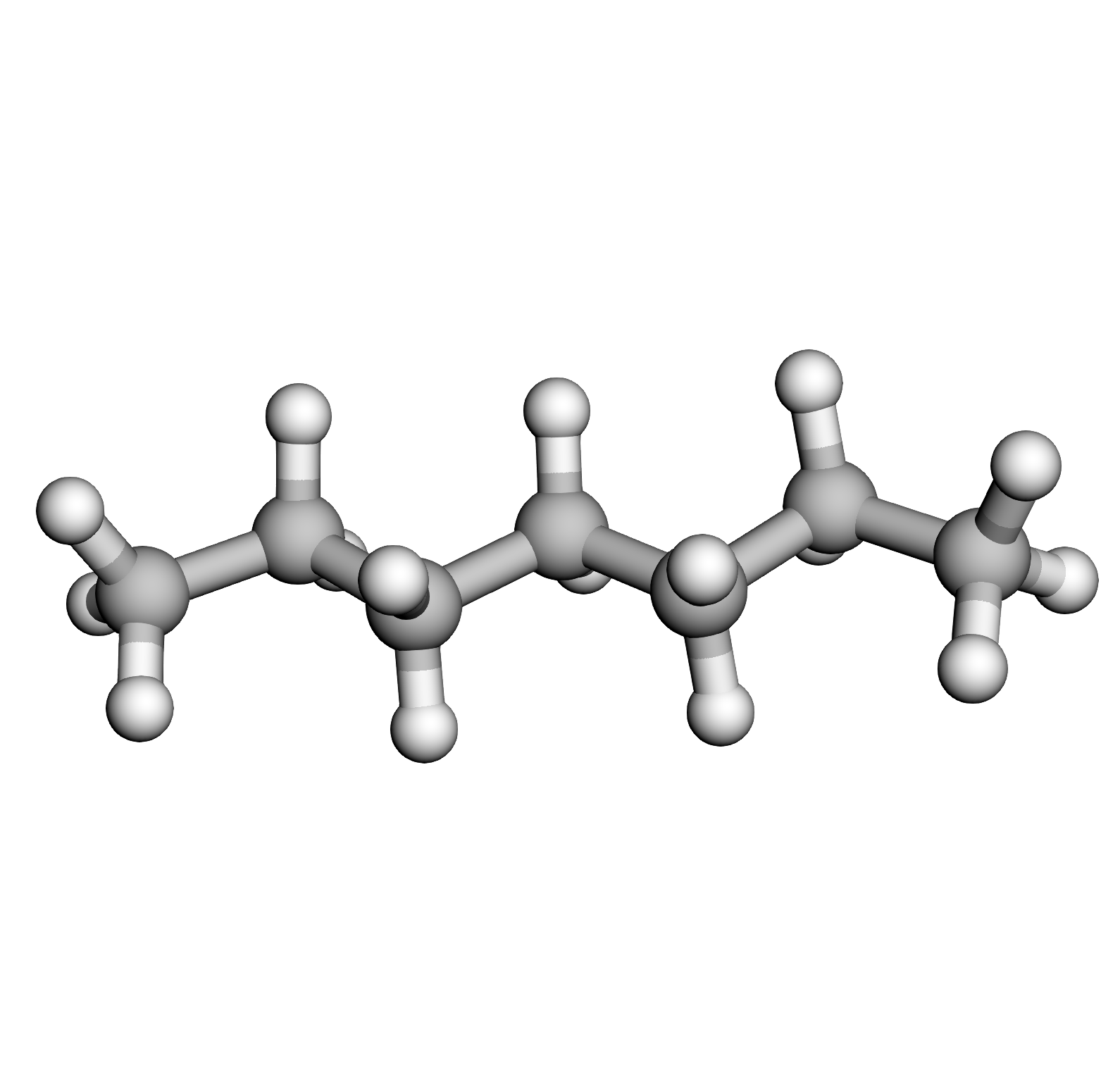} & \includegraphics[trim={0cm 1cm 0cm 1cm},clip,height=1.2cm]{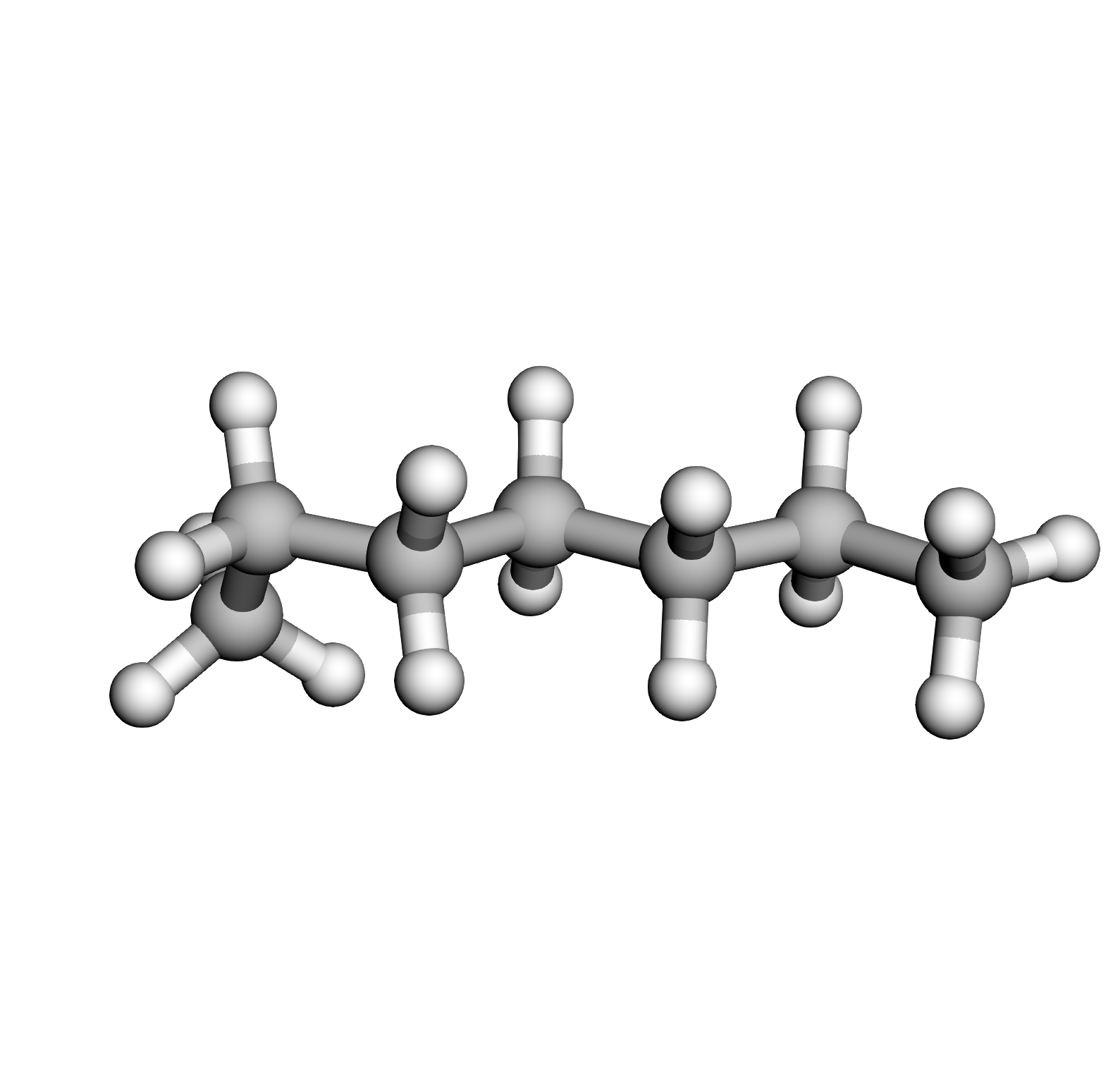} \\
    \bottomrule
    \end{tabular}
\end{table*}

\subsection{Analysis}
\begin{figure*}[t]
    \centering
    \begin{subfigure}[t]{0.32\textwidth}
        \centering
        \caption{Shared molecules}
        \includegraphics[width=\linewidth]{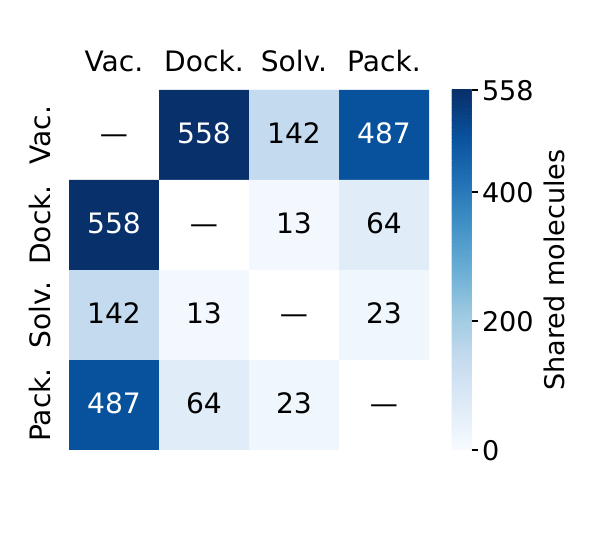}
        \label{fig:heatmap}
    \end{subfigure}
    \hfill
    \begin{subfigure}[t]{0.32\textwidth}
        \centering
        \caption{Solvation-Packing ($n=23$)}
        \includegraphics[width=\linewidth]{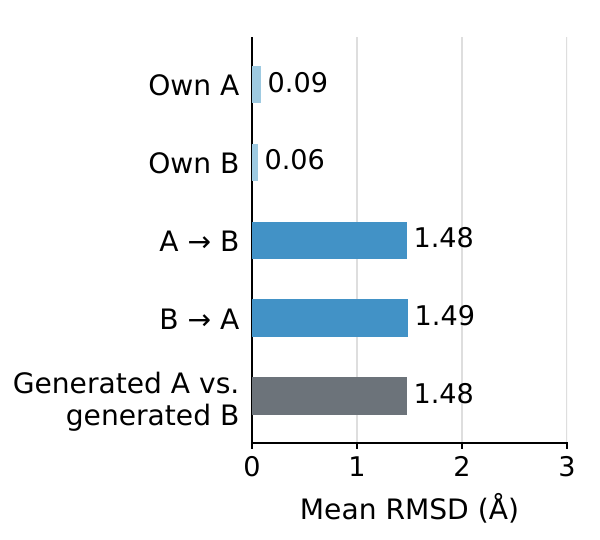}
        \label{fig:solv-pack}
    \end{subfigure}
    \hfill
    \begin{subfigure}[t]{0.32\textwidth}
        \centering
        \caption{Solvation-Docking ($n=13$)}
        \includegraphics[width=\linewidth]{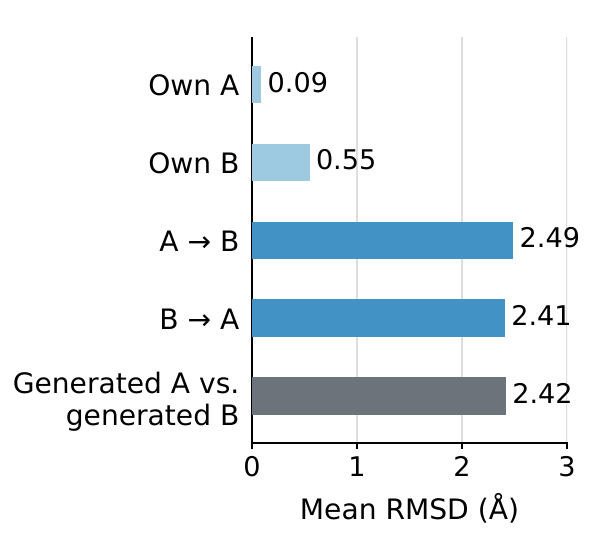}
        \label{fig:solv-dock-main}
    \end{subfigure}
\caption{Shared-molecule analysis (Pretrain+Unified), using values from Table~\ref{tab:overlap-scaled}. (a) Pairwise counts of shared molecules; darker cells indicate more molecules, and dashes mark the diagonal. (b,\,c) Mean heavy-atom RMSD (\AA) after alignment for five comparisons. In both panels, $A$ denotes Solvation; $B$ denotes Packing in (b) and Docking in (c). ``Own $A$'' and ``Own $B$'' compare each conditioned generation with its own environment's reference. $A\!\rightarrow\!B$ compares the $A$-conditioned generation with the reference from $B$, and conversely for $B\!\rightarrow\!A$. ``Generated A vs.\ generated B'' compares the two environment-conditioned generations and corresponds to generation shift in Table~\ref{tab:overlap-scaled}.}
\label{fig:overlap}
\end{figure*}

We evaluate environment conditioning using molecules matched by canonical SMILES across all six environment pairs. For each pair, the Pretrain+Unified model generates conformations of the same molecules conditioned on each environment. Figure~\ref{fig:overlap} and Table~\ref{tab:overlap-scaled} compare these generations with the environment-specific references using mean heavy-atom RMSD after alignment.

\begin{table*}[t]
\centering
\caption{Shared-molecule analysis (Pretrain+Unified), corresponding to Figure~\ref{fig:overlap}. All values are mean heavy-atom RMSD in \AA.}
\label{tab:overlap-scaled}
\begin{tabular}{lrrrrrr}
\toprule
Pair ($A/B$) & $n$ & own $A$ & own $B$ & cross $A$ & cross $B$ & generation shift \\
\midrule
Solvation--Vacuum & 142 & 0.12 & 0.64 & 3.41 & 3.31 & 3.41 \\
Docking--Vacuum & 558 & 0.75 & 0.76 & 1.22 & 1.21 & 1.20 \\
Vacuum--Packing & 487 & 0.33 & 0.35 & 0.64 & 0.65 & 0.64 \\
Docking--Solvation & 13 & 0.55 & 0.09 & 2.41 & 2.49 & 2.42 \\
Packing--Solvation & 23 & 0.06 & 0.09 & 1.49 & 1.48 & 1.48 \\
Packing--Docking & 64 & 0.28 & 0.36 & 0.45 & 0.38 & 0.42 \\
\bottomrule
\end{tabular}
\end{table*}

Across all six environment pairs, generated conformations are closer on average to their own environment's reference than to the other environment's reference (Table~\ref{tab:overlap-scaled}). The gap is largest for Solvation--Vacuum and smallest for Packing--Docking. Together with the visualizations in Table~\ref{tab:vis}, this pattern supports EVA-Flow's ability to generate environment-specific conformations for the same molecule.

\begin{samepage}
Packing--Docking has similar reference geometries and similar generated conformations across environments, consistent with the model responding to context rather than forcing differences from environment labels. Zeroing domain-specific context features degrades performance in Docking, Solvation, and Packing (Appendix~\ref{appendix:envsens}), confirming the use of environment representations. Generated samples also exhibit non-trivial conformational spread (Appendix~\ref{appendix:diversity}).

\end{samepage}

\begin{figure*}[t]
    \centering
    \includegraphics[width=0.98\linewidth]{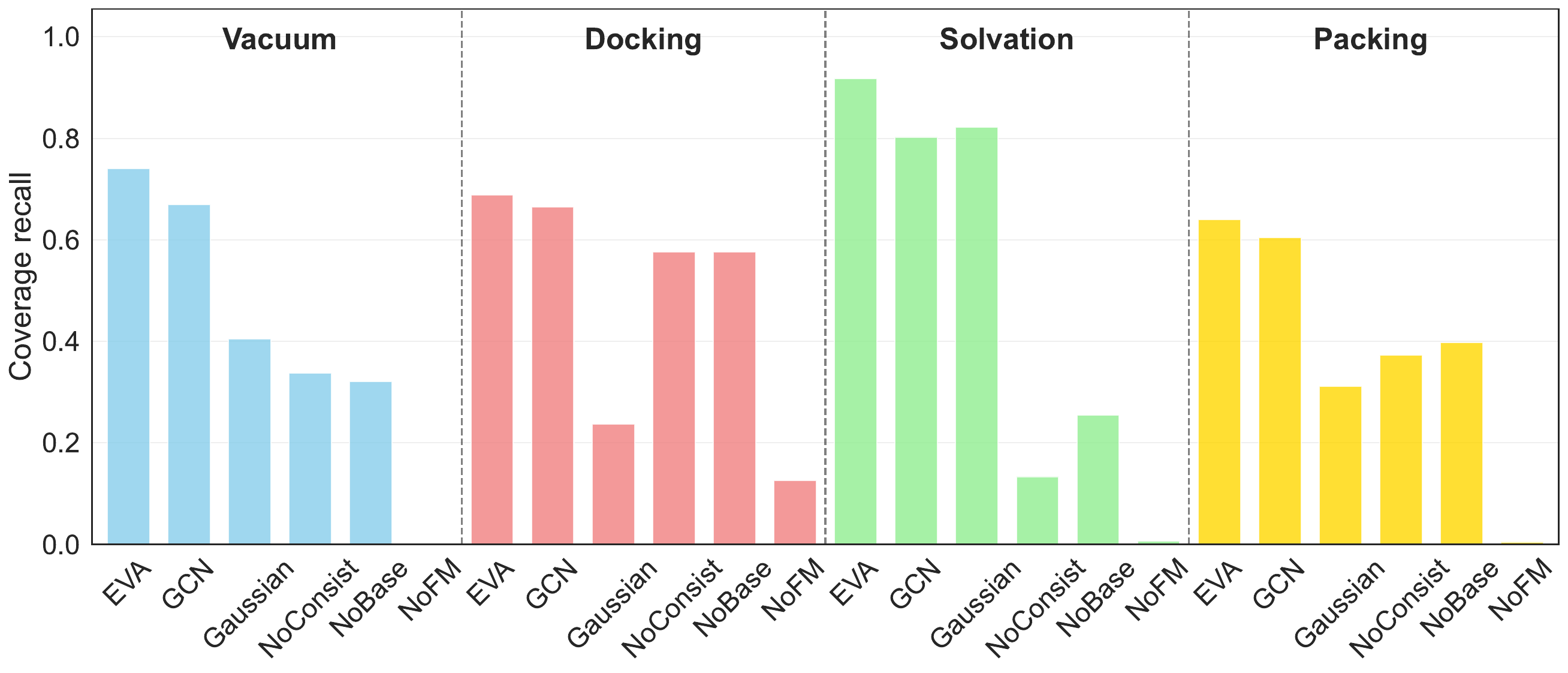}
\caption{Recall COV for the small model in the Unified setup across Vacuum, Docking, Solvation, and Packing.
`EVA' refers to the standard setup; `GCN' uses non-equivariant networks for the encoder and base; `Gaussian' uses a Gaussian base distribution instead of a Harmonic one; `NoConsist' removes latent consistency loss; `NoBase' replaces the learned base distribution network for initial positions $x_0$ with a Harmonic distribution determined only by the graph; `NoFM' removes the FM decoder and uses a regression decoder.}
\label{fig:ablation}
\end{figure*}

\clearpage
\subsection{Ablation Studies}

\begin{samepage}
We ablate components using the Unified setup with the small model (Figure \ref{fig:ablation}). The full model (EVA) is defined in Sections \ref{sec:architecture} and \ref{sec:loss}. Variants: \textbf{GCN} replaces SE(3)-equivariant modules with graph convolutional networks; \textbf{Gaussian} uses a diagonal Gaussian base instead of the Harmonic distribution; \textbf{NoConsist} removes the latent consistency loss; \textbf{NoBase} replaces the learned base network with the graph-topology Harmonic prior $\mathcal{N}(\mathbf{0}, L^{\dagger})$; \textbf{NoFM} replaces the FM decoder with a regression decoder; and \textbf{Latent-free} removes both the learned base and latent, retaining environment conditioning, the fixed harmonic base, and the FM decoder.

\end{samepage}

\begin{samepage}
Three components are essential across all environments: the FM decoder, the Harmonic base, and SE(3)-equivariance (Figure~\ref{fig:ablation}). Replacing the FM decoder with regression collapses Recall COV to $0.000/0.126/0.006/0.005$. The Harmonic base outperforms a diagonal Gaussian, and SE(3)-equivariant modules outperform GCN. The learned base and VAE latent provide consistent gains in the tested cross-environment transfer regimes. Removing both lowers Recall COV in all twelve transfer comparisons, with the largest gaps in low-data Solvation (up to $-0.255$ Recall COV), while effects under Individual training vary by environment (Appendix~\ref{appendix:latent-free-grid}). In the Unified ablations, NoBase and NoConsist also reduce Recall COV. Base and latent diagnostics support an optimization role for these components (Appendices~\ref{appendix:base-consistency} and~\ref{appendix:latent-diagnostics}). Complete metrics are in Appendix~\ref{appendix:ablation-full}.

\end{samepage}

\section{Conclusion}
\label{sec:conclusion}

We introduced EVA-Flow, a unified framework for environment-aware molecular conformation generation. Its shared flow-matching backbone conditions on heterogeneous representations of vacuum, protein pockets, solvation descriptors, and crystal metadata. The largest gains from pretraining and unified finetuning occur in the three data-scarce environments. Multi-reference Solvation evaluation supports broad ensemble coverage, while shared-molecule and context-ablation analyses support environment-responsive generation.

More expressive backbones and scaling could improve accuracy, while multi-reference benchmarks and environment-specific physical validation would extend evaluation beyond Solvation and Vacuum. Extending to unseen environment types such as surface adsorption or solid--liquid interfaces remains future work.

\section*{Broader Impact Statement}
EVA-Flow is intended to support molecular modeling in chemistry, materials science, and drug discovery. More accurate conformation generation could reduce the computational cost of downstream simulations. Like other molecular generative and predictive tools, the method may produce inaccurate geometries and should not replace experimental validation or high-accuracy physical simulation in safety-critical applications. We recommend validating generated structures with appropriate physical calculations and domain expertise before downstream use.

\bibliography{main}

\begin{thebibliography}{45}
\providecommand{\natexlab}[1]{#1}
\providecommand{\url}[1]{\texttt{#1}}
\expandafter\ifx\csname urlstyle\endcsname\relax
  \providecommand{\doi}[1]{doi: #1}\else
  \providecommand{\doi}{doi: \begingroup \urlstyle{rm}\Url}\fi

\bibitem[Axelrod \& Gomez-Bombarelli(2022)Axelrod and
  Gomez-Bombarelli]{axelrod2022geom}
Simon Axelrod and Rafael Gomez-Bombarelli.
\newblock {GEOM}, energy-annotated molecular conformations for property
  prediction and molecular generation.
\newblock \emph{Scientific Data}, 9\penalty0 (1):\penalty0 185, 2022.

\bibitem[Bemis \& Murcko(1996)Bemis and Murcko]{bemis1996properties}
Guy~W Bemis and Mark~A Murcko.
\newblock The properties of known drugs. 1. molecular frameworks.
\newblock \emph{Journal of medicinal chemistry}, 39\penalty0 (15):\penalty0
  2887--2893, 1996.

\bibitem[Chattopadhyay et~al.(2025)Chattopadhyay, Hill, Wright, Beran, and
  Cruz-Cabeza]{doi:10.1021/jacs.5c09997}
Amrita Chattopadhyay, Adam~R. Hill, Sarah~E. Wright, Gregory J.~O. Beran, and
  Aurora~J. Cruz-Cabeza.
\newblock Lattice energy partitions in crystals of flexible molecules and the
  40\% limit.
\newblock \emph{Journal of the American Chemical Society}, 147\penalty0
  (43):\penalty0 39192--39203, 2025.
\newblock \doi{10.1021/jacs.5c09997}.
\newblock URL \url{https://doi.org/10.1021/jacs.5c09997}.
\newblock PMID: 41094351.

\bibitem[Chen \& Lipman(2023)Chen and Lipman]{chen2023flow}
Ricky~TQ Chen and Yaron Lipman.
\newblock Flow matching on general geometries.
\newblock \emph{arXiv preprint arXiv:2302.03660}, 2023.

\bibitem[Corso et~al.(2022)Corso, St{\"a}rk, Jing, Barzilay, and
  Jaakkola]{corso2022diffdock}
Gabriele Corso, Hannes St{\"a}rk, Bowen Jing, Regina Barzilay, and Tommi
  Jaakkola.
\newblock {DiffDock}: Diffusion steps, twists, and turns for molecular docking.
\newblock \emph{arXiv preprint arXiv:2210.01776}, 2022.

\bibitem[Corso et~al.(2024)Corso, Deng, Fry, Polizzi, Barzilay, and
  Jaakkola]{corso2024deep}
Gabriele Corso, Arthur Deng, Benjamin Fry, Nicholas Polizzi, Regina Barzilay,
  and Tommi Jaakkola.
\newblock Deep confident steps to new pockets: Strategies for docking
  generalization.
\newblock \emph{ArXiv}, pp.\  arXiv--2402, 2024.

\bibitem[Cruz-Cabeza \& Bernstein(2014)Cruz-Cabeza and
  Bernstein]{cruz2014conformational}
Aurora~J Cruz-Cabeza and Joel Bernstein.
\newblock Conformational polymorphism.
\newblock \emph{Chemical reviews}, 114\penalty0 (4):\penalty0 2170--2191, 2014.

\bibitem[Dao et~al.(2023)Dao, Phung, Nguyen, and Tran]{dao2023flow}
Quan Dao, Hao Phung, Binh Nguyen, and Anh Tran.
\newblock Flow matching in latent space.
\newblock \emph{arXiv preprint arXiv:2307.08698}, 2023.

\bibitem[Eastman et~al.(2023)Eastman, Behara, Dotson, Galvelis, Herr, Horton,
  Mao, Chodera, Pritchard, Wang, et~al.]{eastman2023spice}
Peter Eastman, Pavan~Kumar Behara, David~L Dotson, Raimondas Galvelis, John~E
  Herr, Josh~T Horton, Yuezhi Mao, John~D Chodera, Benjamin~P Pritchard,
  Yuanqing Wang, et~al.
\newblock {SPICE}, a dataset of drug-like molecules and peptides for training
  machine learning potentials.
\newblock \emph{Scientific Data}, 10\penalty0 (1):\penalty0 11, 2023.

\bibitem[Fischer et~al.(2023)Fischer, Gui, Ma, Stracke, Baumann, and
  Ommer]{fischer2023boosting}
Johannes~S Fischer, Ming Gui, Pingchuan Ma, Nick Stracke, Stefan~A Baumann, and
  Bj{\"o}rn Ommer.
\newblock Boosting latent diffusion with flow matching.
\newblock \emph{arXiv preprint arXiv:2312.07360}, 2023.

\bibitem[Friede et~al.(2024)Friede, H{\"o}lzer, Ehlert, and
  Grimme]{friede2024dxtb}
Marvin Friede, Christian H{\"o}lzer, Sebastian Ehlert, and Stefan Grimme.
\newblock dxtb---an efficient and fully differentiable framework for extended
  tight-binding.
\newblock \emph{The Journal of Chemical Physics}, 161\penalty0 (6), 2024.

\bibitem[Ganea et~al.(2021)Ganea, Pattanaik, Coley, Barzilay, Jensen, Green,
  and Jaakkola]{ganea2021geomol}
Octavian Ganea, Lagnajit Pattanaik, Connor Coley, Regina Barzilay, Klavs
  Jensen, William Green, and Tommi Jaakkola.
\newblock {GeoMol}: Torsional geometric generation of molecular {3D} conformer
  ensembles.
\newblock \emph{Advances in Neural Information Processing Systems},
  34:\penalty0 13757--13769, 2021.

\bibitem[Gao et~al.(2010)Gao, Park, and Stern]{GAO2010901}
Cen Gao, Min-Sun Park, and Harry~A. Stern.
\newblock Accounting for ligand conformational restriction in calculations of
  protein-ligand binding affinities.
\newblock \emph{Biophysical Journal}, 98\penalty0 (5):\penalty0 901--910, 2010.
\newblock ISSN 0006-3495.
\newblock \doi{https://doi.org/10.1016/j.bpj.2009.11.018}.
\newblock URL
  \url{https://www.sciencedirect.com/science/article/pii/S0006349509017469}.

\bibitem[Gra{\v{z}}ulis et~al.(2009)Gra{\v{z}}ulis, Chateigner, Downs, Yokochi,
  Quir{\'o}s, Lutterotti, Manakova, Butkus, Moeck, and
  Le~Bail]{gravzulis2009crystallography}
Saulius Gra{\v{z}}ulis, Daniel Chateigner, Robert~T Downs, Alex~FT Yokochi,
  Miguel Quir{\'o}s, Luca Lutterotti, Elena Manakova, Justas Butkus, Peter
  Moeck, and Armel Le~Bail.
\newblock Crystallography open database--an open-access collection of crystal
  structures.
\newblock \emph{Applied Crystallography}, 42\penalty0 (4):\penalty0 726--729,
  2009.

\bibitem[Gra{\v{z}}ulis et~al.(2012)Gra{\v{z}}ulis, Da{\v{s}}kevi{\v{c}},
  Merkys, Chateigner, Lutterotti, Quiros, Serebryanaya, Moeck, Downs, and
  Le~Bail]{gravzulis2012crystallography}
Saulius Gra{\v{z}}ulis, Adriana Da{\v{s}}kevi{\v{c}}, Andrius Merkys, Daniel
  Chateigner, Luca Lutterotti, Miguel Quiros, Nadezhda~R Serebryanaya, Peter
  Moeck, Robert~T Downs, and Armel Le~Bail.
\newblock Crystallography open database (cod): an open-access collection of
  crystal structures and platform for world-wide collaboration.
\newblock \emph{Nucleic acids research}, 40\penalty0 (D1):\penalty0 D420--D427,
  2012.

\bibitem[Grimme et~al.(2017)Grimme, Bannwarth, and Shushkov]{grimme2017robust}
Stefan Grimme, Christoph Bannwarth, and Philip Shushkov.
\newblock A robust and accurate tight-binding quantum chemical method for
  structures, vibrational frequencies, and noncovalent interactions of large
  molecular systems parametrized for all spd-block elements (z= 1--86).
\newblock \emph{Journal of chemical theory and computation}, 13\penalty0
  (5):\penalty0 1989--2009, 2017.

\bibitem[Groom et~al.(2016)Groom, Bruno, Lightfoot, and
  Ward]{groom2016cambridge}
Colin~R Groom, Ian~J Bruno, Matthew~P Lightfoot, and Suzanna~C Ward.
\newblock The cambridge structural database.
\newblock \emph{Structural Science}, 72\penalty0 (2):\penalty0 171--179, 2016.

\bibitem[Hassan et~al.(2024)Hassan, Shenoy, Lee, Stark, Thaler, and
  Beaini]{hassan2024etflow}
Majdi Hassan, Nikhil Shenoy, Jungyoon Lee, Hannes Stark, Stephan Thaler, and
  Dominique Beaini.
\newblock {ET}-flow: Equivariant flow-matching for molecular conformer
  generation.
\newblock In \emph{The Thirty-eighth Annual Conference on Neural Information
  Processing Systems}, 2024.
\newblock URL \url{https://openreview.net/forum?id=avsZ9OlR60}.

\bibitem[Jing et~al.(2022)Jing, Corso, Chang, Barzilay, and
  Jaakkola]{jing2022torsional}
Bowen Jing, Gabriele Corso, Jeffrey Chang, Regina Barzilay, and Tommi Jaakkola.
\newblock Torsional diffusion for molecular conformer generation.
\newblock \emph{Advances in neural information processing systems},
  35:\penalty0 24240--24253, 2022.

\bibitem[Joshi et~al.(2025)Joshi, Fu, Liao, Gharakhanyan, Miller, Sriram, and
  Ulissi]{joshi2025all}
Chaitanya~K Joshi, Xiang Fu, Yi-Lun Liao, Vahe Gharakhanyan, Benjamin~Kurt
  Miller, Anuroop Sriram, and Zachary~W Ulissi.
\newblock All-atom diffusion transformers: Unified generative modelling of
  molecules and materials.
\newblock \emph{arXiv preprint arXiv:2503.03965}, 2025.

\bibitem[Kingma et~al.(2019)Kingma, Welling, et~al.]{kingma2019introduction}
Diederik~P Kingma, Max Welling, et~al.
\newblock An introduction to variational autoencoders.
\newblock \emph{Foundations and Trends{\textregistered} in Machine Learning},
  12\penalty0 (4):\penalty0 307--392, 2019.

\bibitem[Landrum \& contributors(2025)Landrum and contributors]{RDKit}
Greg Landrum and contributors.
\newblock {RDKit}: Open-source cheminformatics.
\newblock \url{https://www.rdkit.org}, 2025.
\newblock Version 2025.03.6; accessed 2025-09-14.

\bibitem[Lipman et~al.(2022)Lipman, Chen, Ben-Hamu, Nickel, and
  Le]{lipman2022flow}
Yaron Lipman, Ricky~TQ Chen, Heli Ben-Hamu, Maximilian Nickel, and Matt Le.
\newblock Flow matching for generative modeling.
\newblock \emph{arXiv preprint arXiv:2210.02747}, 2022.

\bibitem[Liu et~al.(2022)Liu, Gong, and Liu]{liu2022flow}
Xingchao Liu, Chengyue Gong, and Qiang Liu.
\newblock Flow straight and fast: Learning to generate and transfer data with
  rectified flow.
\newblock \emph{arXiv preprint arXiv:2209.03003}, 2022.

\bibitem[Liu et~al.(2015)Liu, Li, Han, Li, Liu, Zhao, Nie, Liu, and
  Wang]{Liu2015PDBbind}
Zhihai Liu, Yan Li, Li~Han, Jie Li, Jie Liu, Zhixiong Zhao, Wei Nie, Yuchen
  Liu, and Renxiao Wang.
\newblock {PDB}-wide collection of binding data: current status of the
  {PDBbind} database.
\newblock \emph{Bioinformatics}, 31\penalty0 (3):\penalty0 405--412, 2015.
\newblock \doi{10.1093/bioinformatics/btu626}.

\bibitem[Medrano~Sandonas et~al.(2024)Medrano~Sandonas, Van~Rompaey, Fallani,
  Hilfiker, Hahn, Perez-Benito, Verhoeven, Tresadern, Kurt~Wegner, Ceulemans,
  et~al.]{medrano2024dataset}
Leonardo Medrano~Sandonas, Dries Van~Rompaey, Alessio Fallani, Mathias
  Hilfiker, David Hahn, Laura Perez-Benito, Jonas Verhoeven, Gary Tresadern,
  Joerg Kurt~Wegner, Hugo Ceulemans, et~al.
\newblock Dataset for quantum-mechanical exploration of conformers and solvent
  effects in large drug-like molecules.
\newblock \emph{Scientific Data}, 11\penalty0 (1):\penalty0 742, 2024.

\bibitem[Nikitin et~al.(2025)Nikitin, Dunn, Koes, and Isayev]{D5DD00206K}
Filipp Nikitin, Ian Dunn, David~Ryan Koes, and Olexandr Isayev.
\newblock {GEOM}-drugs revisited: toward more chemically accurate benchmarks
  for {3D} molecule generation.
\newblock \emph{Digital Discovery}, 4:\penalty0 3282--3291, 2025.
\newblock \doi{10.1039/D5DD00206K}.
\newblock URL \url{http://dx.doi.org/10.1039/D5DD00206K}.

\bibitem[O'Boyle et~al.(2008)O'Boyle, Morley, and Hutchison]{o2008pybel}
Noel~M O'Boyle, Chris Morley, and Geoffrey~R Hutchison.
\newblock Pybel: a python wrapper for the openbabel cheminformatics toolkit.
\newblock \emph{Chemistry Central Journal}, 2\penalty0 (1):\penalty0 5, 2008.

\bibitem[O'Boyle et~al.(2011)O'Boyle, Banck, James, Morley, Vandermeersch, and
  Hutchison]{o2011open}
Noel~M O'Boyle, Michael Banck, Craig~A James, Chris Morley, Tim Vandermeersch,
  and Geoffrey~R Hutchison.
\newblock Open babel: An open chemical toolbox.
\newblock \emph{Journal of cheminformatics}, 3\penalty0 (1):\penalty0 33, 2011.

\bibitem[Ong et~al.(2013)Ong, Richards, Jain, Hautier, Kocher, Cholia, Gunter,
  Chevrier, Persson, and Ceder]{ONG2013314}
Shyue~Ping Ong, William~Davidson Richards, Anubhav Jain, Geoffroy Hautier,
  Michael Kocher, Shreyas Cholia, Dan Gunter, Vincent~L. Chevrier, Kristin~A.
  Persson, and Gerbrand Ceder.
\newblock Python materials genomics (pymatgen): A robust, open-source python
  library for materials analysis.
\newblock \emph{Computational Materials Science}, 68:\penalty0 314--319, 2013.
\newblock ISSN 0927-0256.
\newblock \doi{https://doi.org/10.1016/j.commatsci.2012.10.028}.
\newblock URL
  \url{https://www.sciencedirect.com/science/article/pii/S0927025612006295}.

\bibitem[Pracht et~al.(2024)Pracht, Grimme, Bannwarth, Bohle, Ehlert, Feldmann,
  Gorges, M{\"u}ller, Neudecker, Plett, et~al.]{pracht2024crest}
Philipp Pracht, Stefan Grimme, Christoph Bannwarth, Fabian Bohle, Sebastian
  Ehlert, Gereon Feldmann, Johannes Gorges, Marcel M{\"u}ller, Tim Neudecker,
  Christoph Plett, et~al.
\newblock {CREST}---a program for the exploration of low-energy molecular
  chemical space.
\newblock \emph{The Journal of Chemical Physics}, 160\penalty0 (11), 2024.

\bibitem[{RDKit}(2025)]{RDKitMurckoScaffoldDocs}
{RDKit}.
\newblock \emph{{RDKit} \texttt{MurckoScaffold} module documentation}.
\newblock RDKit, 2025.
\newblock URL
  \url{https://www.rdkit.org/docs/source/rdkit.Chem.Scaffolds.MurckoScaffold.html}.
\newblock Accessed 2025-09-14.

\bibitem[Sargent et~al.(2025)Sargent, Hsu, Johnson, Fei-Fei, and
  Wu]{sargent2025flow}
Kyle Sargent, Kyle Hsu, Justin Johnson, Li~Fei-Fei, and Jiajun Wu.
\newblock Flow to the mode: Mode-seeking diffusion autoencoders for
  state-of-the-art image tokenization.
\newblock \emph{arXiv preprint arXiv:2503.11056}, 2025.

\bibitem[Sobornova et~al.(2024)Sobornova, Belov, Krestyaninov, and
  Khodov]{sobornova2024influence}
Valentina~V Sobornova, Konstantin~V Belov, Michael~A Krestyaninov, and Ilya~A
  Khodov.
\newblock Influence of solvent polarity on the conformer ratio of bicalutamide
  in saturated solutions: Insights from {NOESY} {NMR} analysis and
  quantum-chemical calculations.
\newblock \emph{International Journal of Molecular Sciences}, 25\penalty0
  (15):\penalty0 8254, 2024.

\bibitem[Stuke et~al.(2020)Stuke, Kunkel, Golze, Todorovi{\'c}, Margraf,
  Reuter, Rinke, and Oberhofer]{stuke2020atomic}
Annika Stuke, Christian Kunkel, Dorothea Golze, Milica Todorovi{\'c},
  Johannes~T Margraf, Karsten Reuter, Patrick Rinke, and Harald Oberhofer.
\newblock Atomic structures and orbital energies of 61,489 crystal-forming
  organic molecules.
\newblock \emph{Scientific data}, 7\penalty0 (1):\penalty0 58, 2020.

\bibitem[Thompson \& Day(2014)Thompson and Day]{C4SC01132E}
Hugh P.~G. Thompson and Graeme~M. Day.
\newblock Which conformations make stable crystal structures? mapping
  crystalline molecular geometries to the conformational energy landscape.
\newblock \emph{Chem. Sci.}, 5:\penalty0 3173--3182, 2014.
\newblock \doi{10.1039/C4SC01132E}.
\newblock URL \url{http://dx.doi.org/10.1039/C4SC01132E}.

\bibitem[Thölke \& Fabritiis(2022)Thölke and
  Fabritiis]{thölke2022torchmdnetequivarianttransformersneural}
Philipp Thölke and Gianni~De Fabritiis.
\newblock {TorchMD-NET}: Equivariant transformers for neural network based
  molecular potentials, 2022.
\newblock URL \url{https://arxiv.org/abs/2202.02541}.

\bibitem[Tong et~al.(2023)Tong, Fatras, Malkin, Huguet, Zhang, Rector-Brooks,
  Wolf, and Bengio]{tong2023improving}
Alexander Tong, Kilian Fatras, Nikolay Malkin, Guillaume Huguet, Yanlei Zhang,
  Jarrid Rector-Brooks, Guy Wolf, and Yoshua Bengio.
\newblock Improving and generalizing flow-based generative models with
  minibatch optimal transport.
\newblock \emph{arXiv preprint arXiv:2302.00482}, 2023.

\bibitem[Wang \& Pang(2007)Wang and Pang]{10.1371/journal.pone.0000820}
Qi~Wang and Yuan-Ping Pang.
\newblock Preference of small molecules for local minimum conformations when
  binding to proteins.
\newblock \emph{PLOS ONE}, 2:\penalty0 1--10, 09 2007.
\newblock \doi{10.1371/journal.pone.0000820}.
\newblock URL \url{https://doi.org/10.1371/journal.pone.0000820}.

\bibitem[Wang et~al.(2024)Wang, Elhag, Jaitly, Susskind, and
  Bautista]{wang2024swallowing}
Yuyang Wang, Ahmed~AA Elhag, Navdeep Jaitly, Joshua~M Susskind, and
  Miguel~{\'A}ngel Bautista.
\newblock Swallowing the bitter pill: Simplified scalable conformer generation.
\newblock In \emph{International Conference on Machine Learning}, pp.\
  50400--50418. PMLR, 2024.

\bibitem[Weikl \& Paul(2014)Weikl and Paul]{weikl2014conformational}
Thomas~R Weikl and Fabian Paul.
\newblock Conformational selection in protein binding and function.
\newblock \emph{Protein Science}, 23\penalty0 (11):\penalty0 1508--1518, 2014.

\bibitem[Weininger(1988)]{weininger1988smiles}
David Weininger.
\newblock {SMILES}, a chemical language and information system. 1. introduction
  to methodology and encoding rules.
\newblock \emph{Journal of chemical information and computer sciences},
  28\penalty0 (1):\penalty0 31--36, 1988.

\bibitem[Wood et~al.(2026)Wood, Dzamba, Fu, Gao, Shuaibi, Barroso-Luque,
  Abdelmaqsoud, Gharakhanyan, Kitchin, Levine, Michel, Sriram, Cohen, Das,
  Sahoo, Rizvi, Ulissi, and Zitnick]{wood2026uma}
Brandon~M Wood, Misko Dzamba, Xiang Fu, Meng Gao, Muhammed Shuaibi, Luis
  Barroso-Luque, Kareem Abdelmaqsoud, Vahe Gharakhanyan, John~R. Kitchin,
  Daniel~S. Levine, Kyle Michel, Anuroop Sriram, Taco Cohen, Abhishek Das,
  Sushree~Jagriti Sahoo, Ammar Rizvi, Zachary~Ward Ulissi, and C.~Lawrence
  Zitnick.
\newblock {UMA}: A family of universal models for atoms.
\newblock In \emph{The Thirty-ninth Annual Conference on Neural Information
  Processing Systems}, 2026.
\newblock URL \url{https://openreview.net/forum?id=SvopaNxYWt}.

\bibitem[Xu et~al.(2022)Xu, Yu, Song, Shi, Ermon, and Tang]{xu2022geodiff}
Minkai Xu, Lantao Yu, Yang Song, Chence Shi, Stefano Ermon, and Jian Tang.
\newblock {GeoDiff}: A geometric diffusion model for molecular conformation
  generation.
\newblock \emph{arXiv preprint arXiv:2203.02923}, 2022.

\bibitem[Zhang et~al.(2019)Zhang, Tong, Xu, and Maciejewski]{zhang2019graph}
Si~Zhang, Hanghang Tong, Jiejun Xu, and Ross Maciejewski.
\newblock Graph convolutional networks: a comprehensive review.
\newblock \emph{Computational Social Networks}, 6\penalty0 (1):\penalty0 1--23,
  2019.

\end{thebibliography}
\bibliographystyle{tmlr}

\appendix
\section{Data Preparation}
\label{appendix:data}

\paragraph{GEOM QM9 and DRUGS.}
We follow the data processing pipeline of ETFlow~\citep{hassan2024etflow} and adopt the same train/validation/test splits. For GEOM-DRUGS, we applied the data cleaning strategies reported in \citet{D5DD00206K} that fix valency, bond-order, and
force-field inconsistencies in the original datasets. For each molecule, we retain the top 30 conformers ranked by Boltzmann weight. Each molecule is stored as a PyTorch Geometric object containing atomic numbers, 3D positions, and a canonical SMILES string. The QM9 training set contains 106,587 molecules (794,960 conformations) and the validation set 13,323 molecules (99,945 conformations); the test set has 1,000 molecules.
The DRUGS training set contains 272,641 molecules (5,770,377 conformations) and the validation set 30,405 molecules (715,592 conformations); the test set has 1,000 molecules.

\paragraph{SPICE and CREST-relaxed SPICE.}
We use the PubChem subset of SPICE~\citep{eastman2023spice}, which contains small drug-like molecules (18--50 atoms; elements Br, C, Cl, F, H, I, N, O, P, S).

Because the SPICE conformers are not fully relaxed, we construct a complementary relaxed set via CREST~\citep{pracht2024crest}. For each molecule we (i) generate 50 initial conformers with RDKit \texttt{EmbedMultipleConfs} (pruning threshold $0.01$\,\AA, MMFF optimization), (ii) deduplicate with \texttt{GetBestRMS} (threshold $0.1$\,\AA), (iii) refine the ten lowest-energy conformers with extended tight binding (xTB)~\citep{grimme2017robust, friede2024dxtb}, and (iv) run CREST v3.0.2 from the lowest-xTB-energy conformer with a 6\,kcal/mol cutoff. Molecules with cis/trans stereo-SMILES were excluded because the stereochemistry is not preserved through the pipeline.

For both SPICE and CREST-relaxed SPICE, we align atom ordering between the dataset coordinates and the RDKit molecule built from the canonical SMILES using \texttt{GetSubstructMatch}. We split 0.8/0.1/0.1 into train/validation/test. The training set contains 19,198 molecules (959,802 SPICE conformers; 2,203,392 CREST conformers), the validation set 2,508 molecules (125,370; 278,673), and the test set 2,941 molecules.

\paragraph{OE62.}
OE62~\citep{stuke2020atomic} provides PBE-relaxed geometries of crystal-forming organic molecules. We read the SMILES and coordinates, align atom ordering via \texttt{GetSubstructMatch} as above, and split 0.8/0.1/0.1. The training set contains 28,038 molecules, the validation set 4,297, and the test set 3,933 (one conformation each).

\paragraph{PDBBind.}
We use PDBBind-v2020~\citep{Liu2015PDBbind}. For each protein--ligand pair, we extract the ligand from the SDF file, add hydrogens, and generate a canonical SMILES. Atom ordering between the SDF-derived and SMILES-derived RDKit molecules is aligned via \texttt{GetSubstructMatch}.

The protein binding pocket serves as the environment graph. We parse the pocket PDB file and retain only heavy atoms, adding edges between atoms within the same residue. Node features follow ETFlow conventions.

We use the original PDBBind splits. After filtering invalid ligands, the training set has 9,940 examples, the validation set 614, and the test set 218 (one ligand conformation each).

\paragraph{Aquamarine (AQM).}
Aquamarine~\citep{medrano2024dataset} contains conformations relaxed in implicit water. We build RDKit molecules from SMILES, add hydrogens, and align atom ordering as above. The environment graph uses a fixed RDKit-embedded water proxy with four normalized solvation descriptors appended to node features: cavity surface area (sCAV), cavity volume (vCAV), free energy in electrolyte (eSOLV), and non-electrostatic free energy (eNELEC).

After filtering, we split molecules 0.8/0.1/0.1. The processed files retain all Aquamarine conformers, but the default one-reference protocol selects the dataset's \texttt{c1} (index-$0$) conformer for training and evaluation, yielding 1,276 training, 175 validation, and 157 test molecules. This is a fixed canonical-index choice rather than an energy ranking; we report a conformer-selection sensitivity analysis in Appendix~\ref{appendix:aqm-selection}.

\paragraph{Crystallography Open Database (COD).}
The COD~\citep{gravzulis2009crystallography, gravzulis2012crystallography} is an open-access collection of crystal structures. We select organic molecular crystals by applying the following filters: the structure must (i) contain organic molecules, (ii) not be marked as disordered~\citep{groom2016cambridge}, (iii) not be missing hydrogen, (iv) have a valid SMILES~\citep{weininger1988smiles} with a single fragment, (v) pass structural validation~\citep{ONG2013314}, and (vi) contain exactly one formula unit in the asymmetric unit ($Z'=1$).

We recover the molecular topology with RDKit maximum-common-substructure matching (\texttt{FindMCS})~\citep{RDKit} rather than the often-unreliable COD SMILES, and extract the space group, Wyckoff positions, and Cartesian coordinates for each formula unit using \texttt{pymatgen}~\citep{ONG2013314}. We process structures into PyTorch Geometric objects, aligning atom ordering via RDKit \texttt{GetSubstructMatch}. Lattice parameters ($a, b, c, \alpha, \beta, \gamma$) and the space group number are included as features.

After filtering, we split 0.8/0.1/0.1. The training set has 20,765 molecules, the validation set 3,239, and the test set 3,235 (one conformation each).

\section{Distributions of Atom Counts and Rotatable-Bond Counts}
\label{appendix:distribution}

\begin{figure}[t]
    \centering
    \begin{subfigure}[t]{0.45\textwidth}
        \centering
        \caption{GEOM DRUGS Atoms}
        \includegraphics[height=3.5cm]{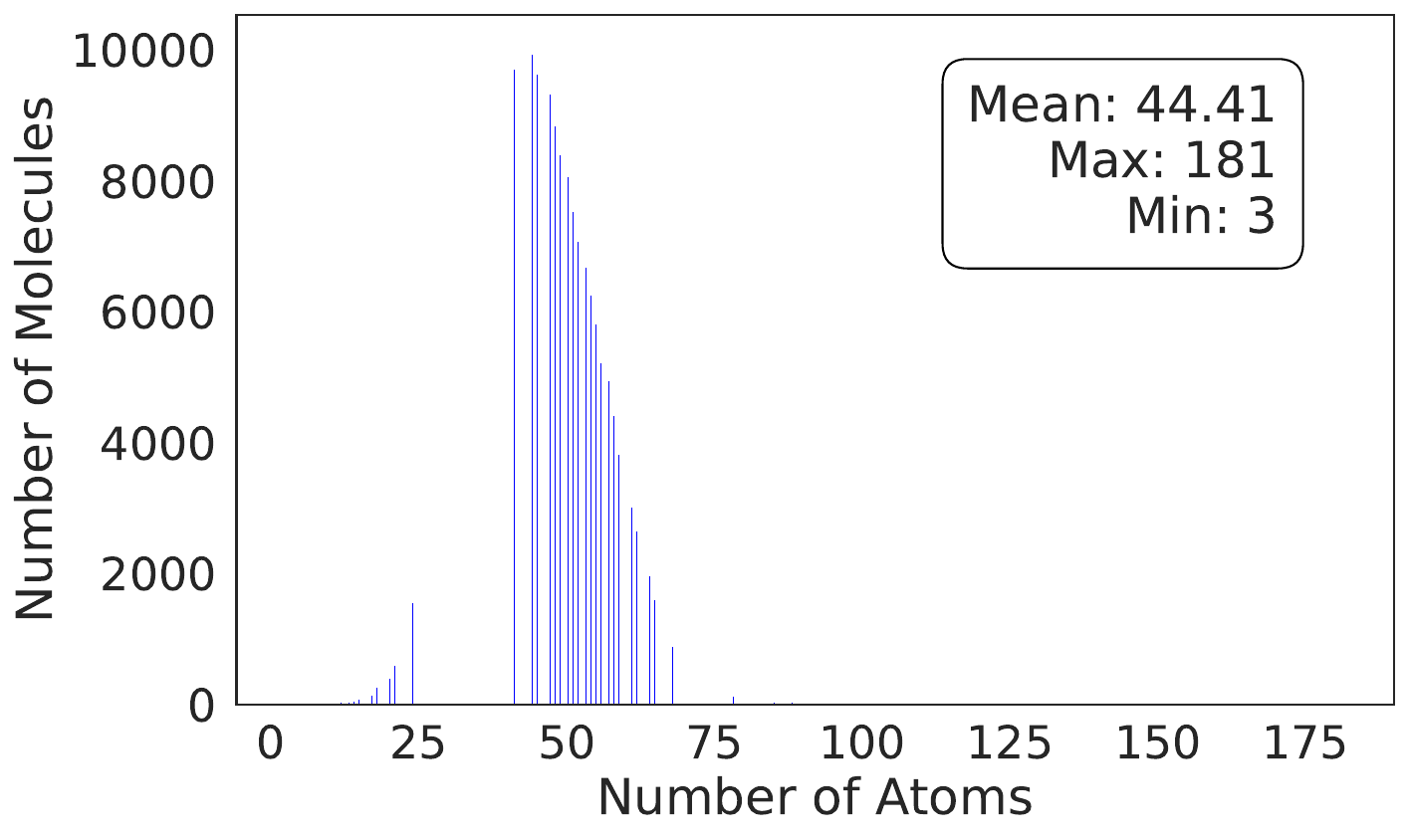}
        \label{fig:drugs-atoms}
    \end{subfigure}
    \hfill
    \begin{subfigure}[t]{0.45\textwidth}
        \centering
        \caption{GEOM DRUGS Rotatable Bonds}
        \includegraphics[height=3.5cm]{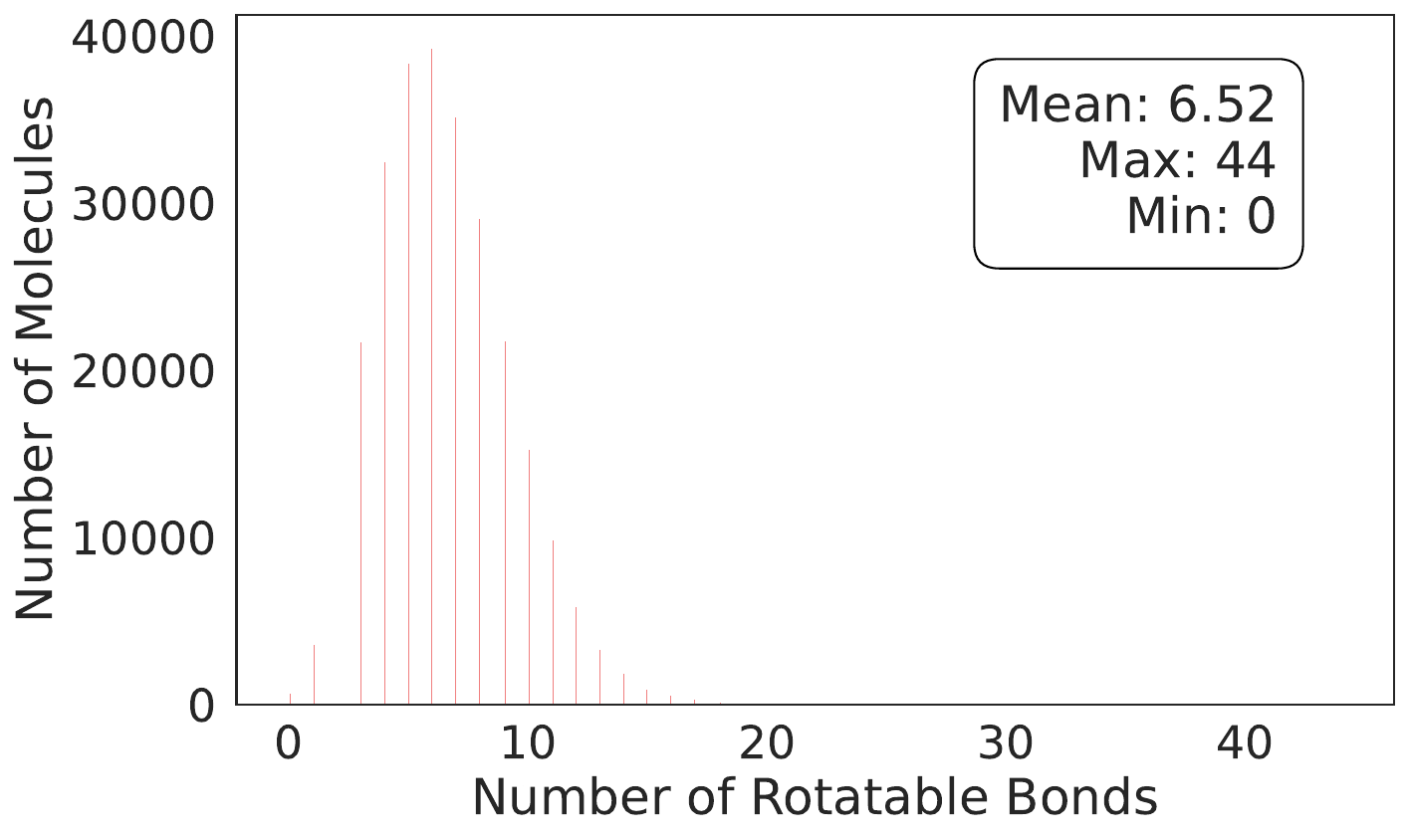}
        \label{fig:drugs-rot}
    \end{subfigure}
    \vfill
    \begin{subfigure}[t]{0.45\textwidth}
        \centering
        \caption{GEOM QM9 Atoms}
        \includegraphics[height=3.5cm]{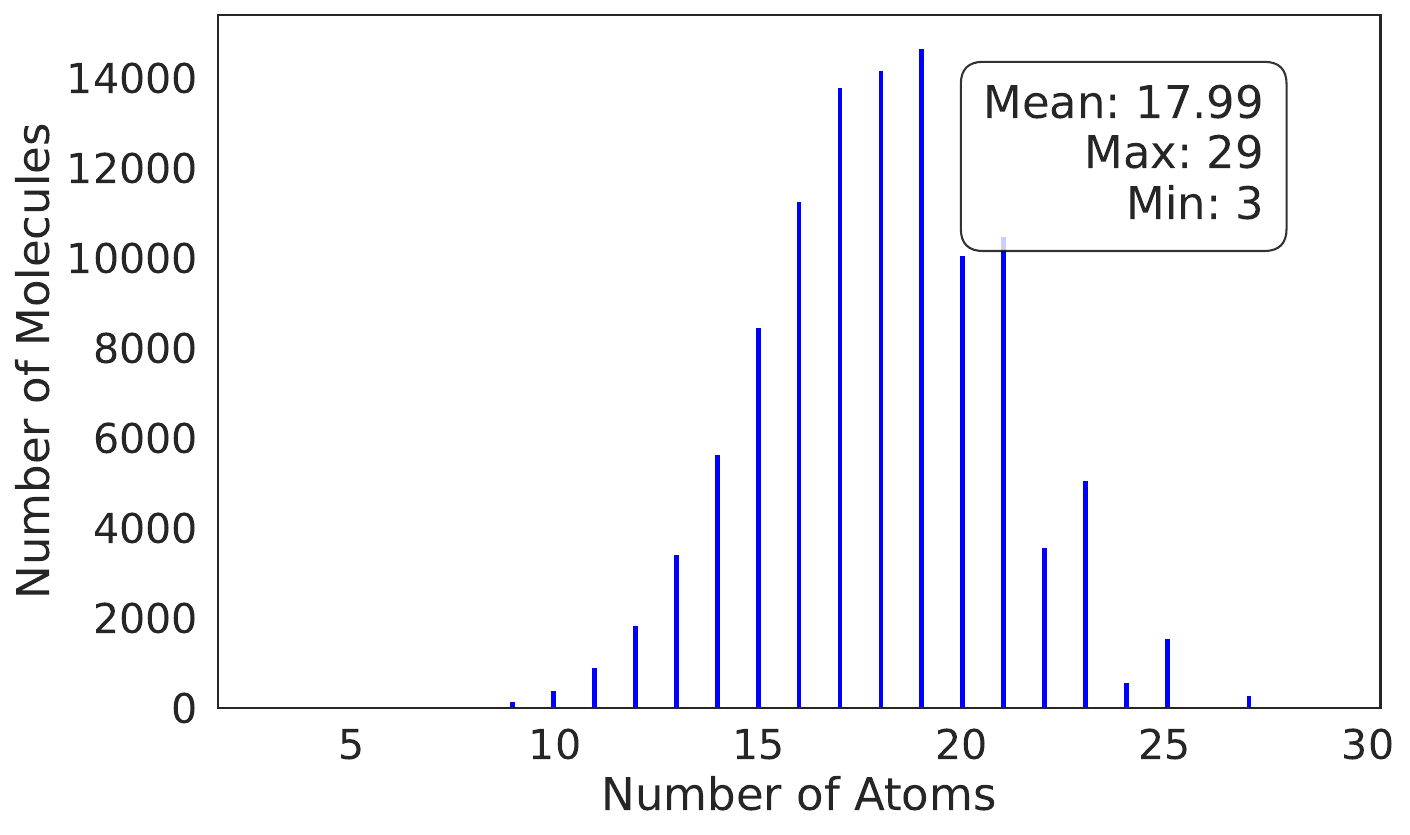}
        \label{fig:qm9-atoms}
    \end{subfigure}
    \hfill
    \begin{subfigure}[t]{0.45\textwidth}
        \centering
        \caption{GEOM QM9 Rotatable Bonds}
        \includegraphics[height=3.5cm]{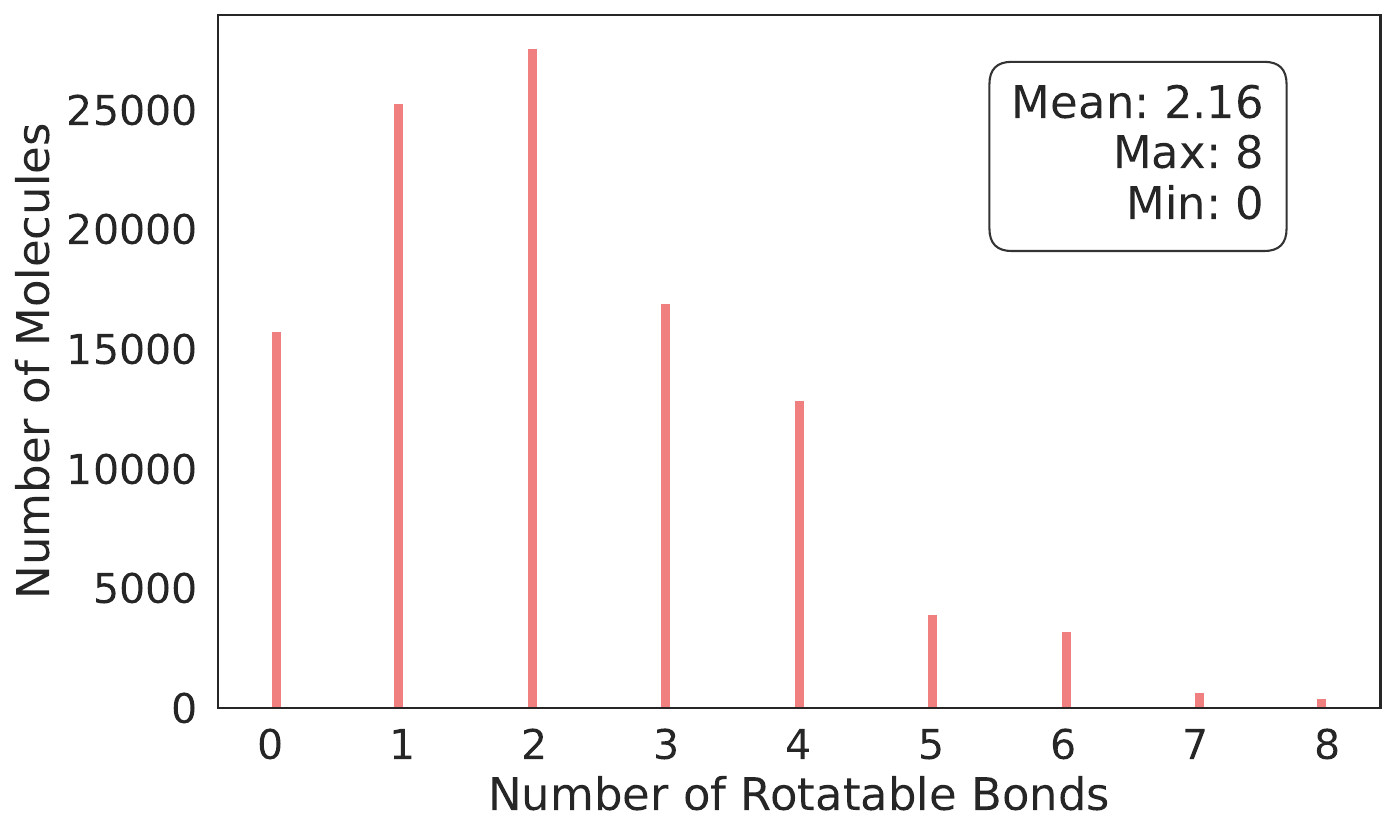}
        \label{fig:qm9-rot}
    \end{subfigure}
    \vfill
    \begin{subfigure}[t]{0.45\textwidth}
        \centering
        \caption{SPICE/CREST Atoms}
        \includegraphics[height=3.5cm]{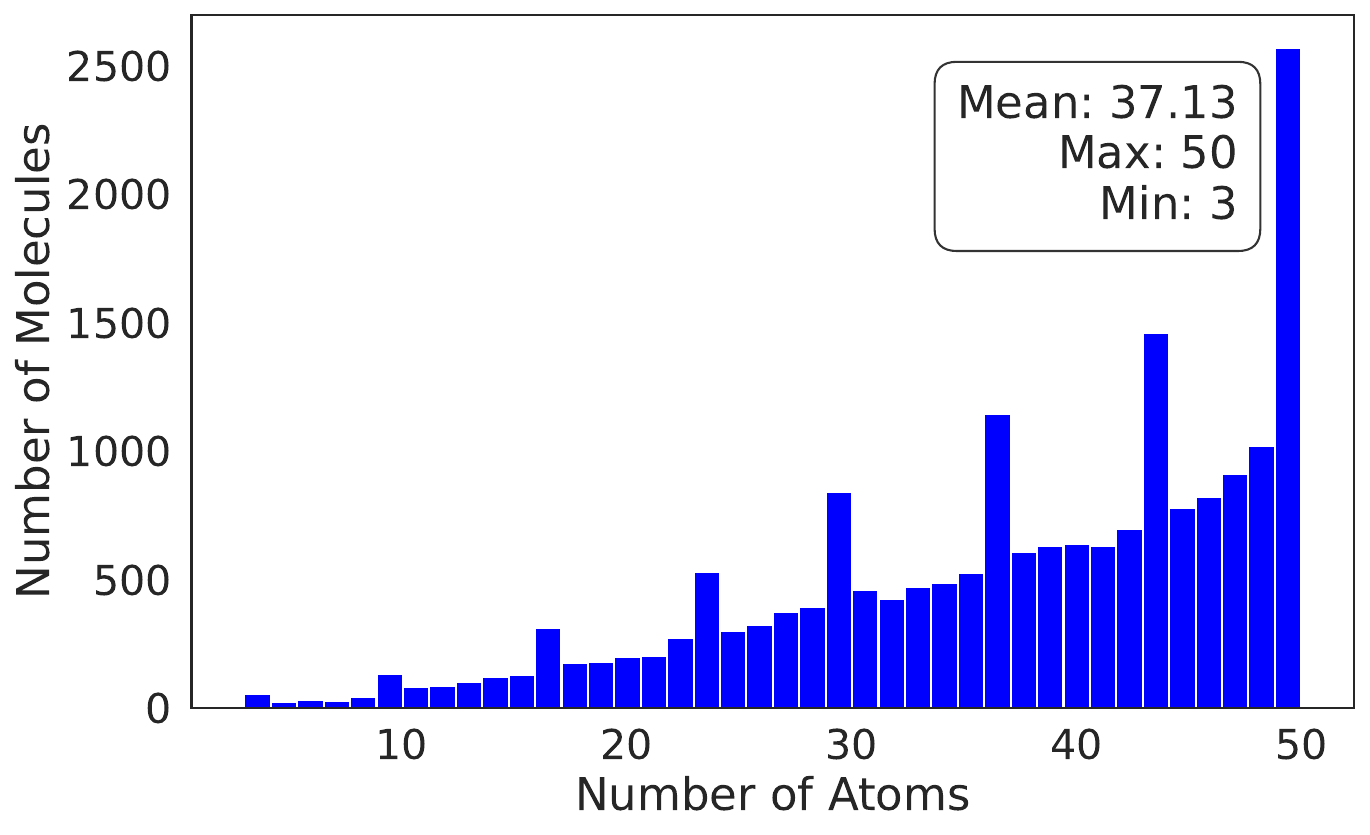}
        \label{fig:spice-atoms}
    \end{subfigure}
    \hfill
    \begin{subfigure}[t]{0.45\textwidth}
        \centering
        \caption{SPICE/CREST Rotatable Bonds}
        \includegraphics[height=3.5cm]{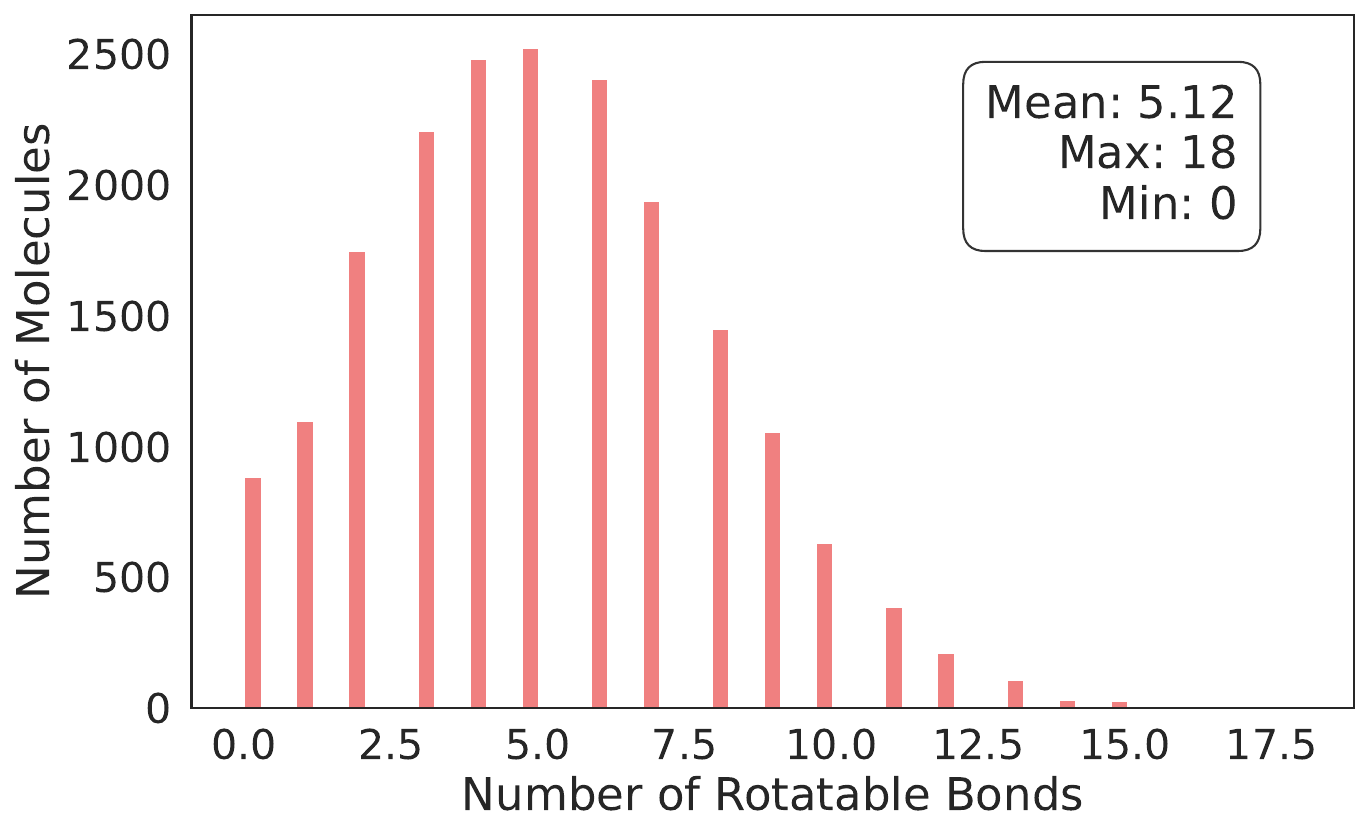}
        \label{fig:spice-rot}
    \end{subfigure}
    \vfill
    \begin{subfigure}[t]{0.45\textwidth}
        \centering
        \caption{OE62 Atoms}
        \includegraphics[height=3.5cm]{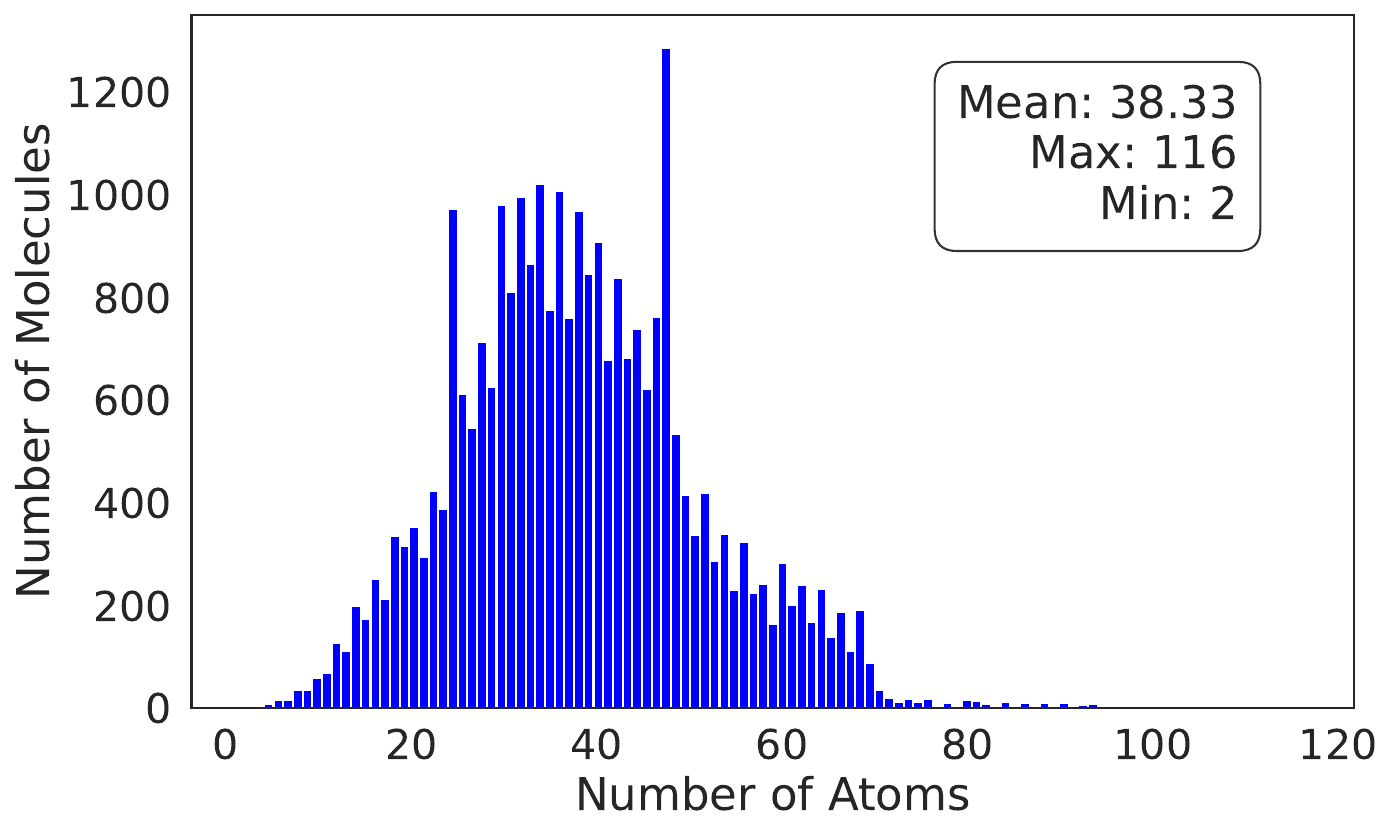}
        \label{fig:oe62-atoms}
    \end{subfigure}
    \hfill
    \begin{subfigure}[t]{0.45\textwidth}
        \centering
        \caption{OE62 Rotatable Bonds}
        \includegraphics[height=3.5cm]{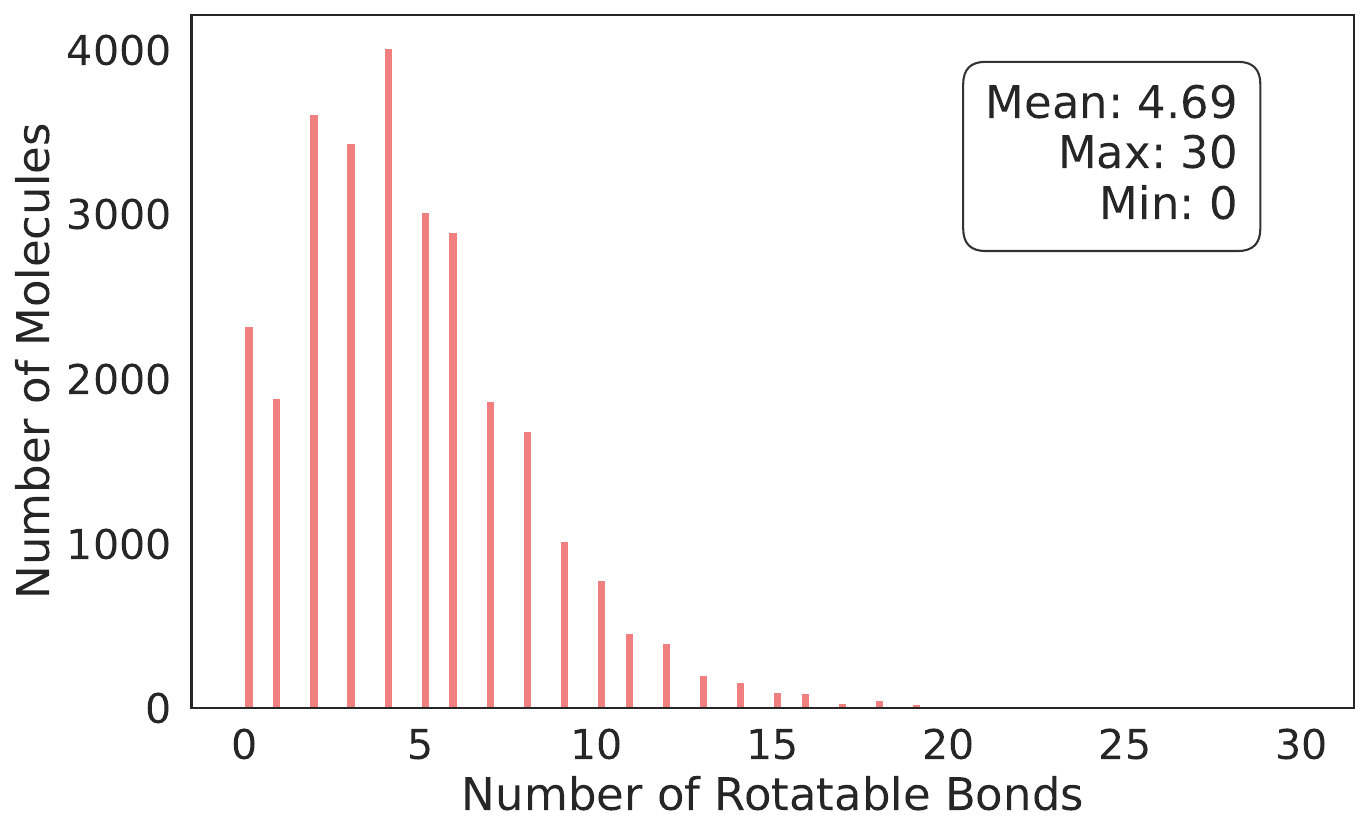}
        \label{fig:oe62-rot}
    \end{subfigure}
    
\caption{Histograms of atom and rotatable-bond counts for the pretraining datasets: GEOM DRUGS, GEOM QM9, SPICE, CREST-relaxed SPICE, and OE62.}
\label{fig:hist_pretrain}
\end{figure}

\begin{figure}[t]
    \centering
    \begin{subfigure}[t]{0.45\textwidth}
        \centering
        \caption{PDBBind Atoms}
        \includegraphics[height=3.5cm]{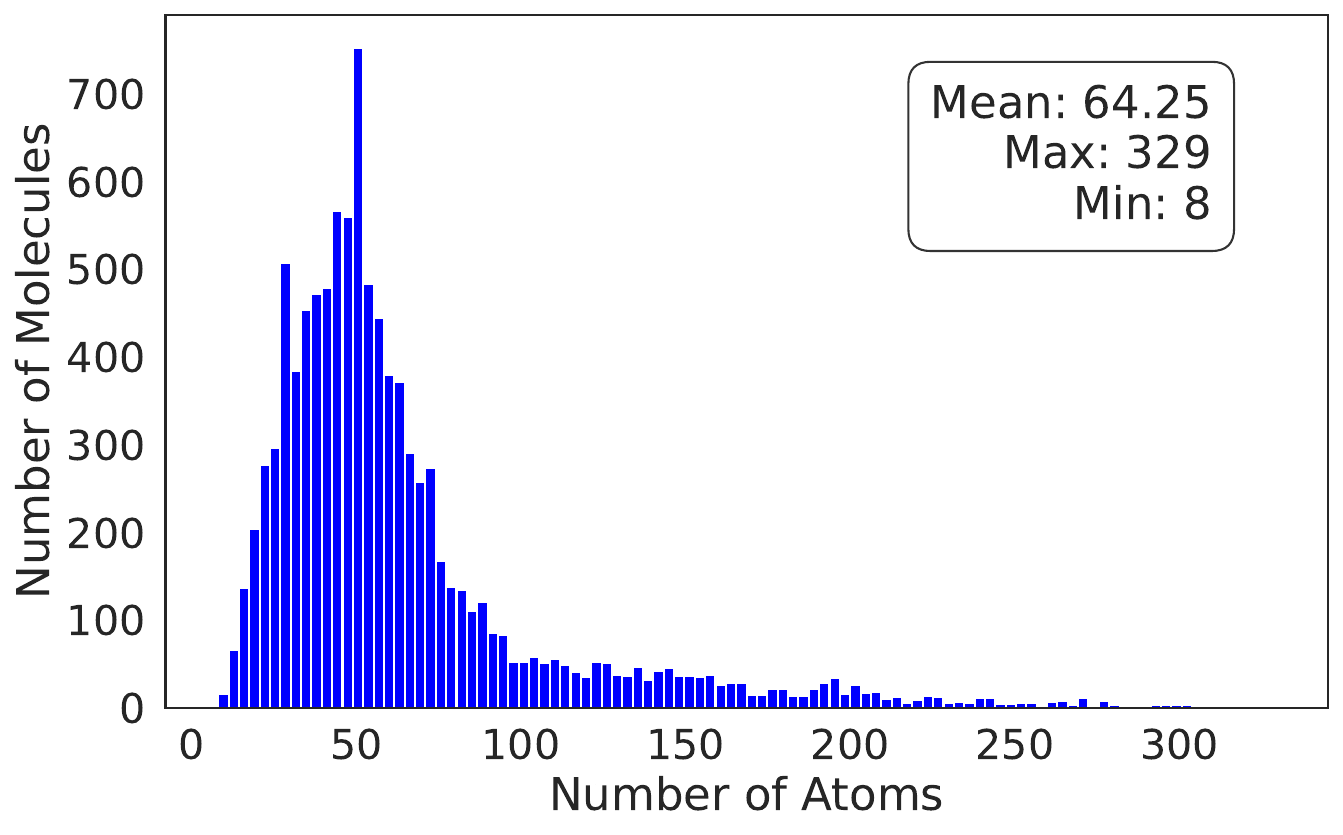}
        \label{fig:pdb-atoms}
    \end{subfigure}
    \hfill
    \begin{subfigure}[t]{0.45\textwidth}
        \centering
        \caption{PDBBind Rotatable Bonds}
        \includegraphics[height=3.5cm]{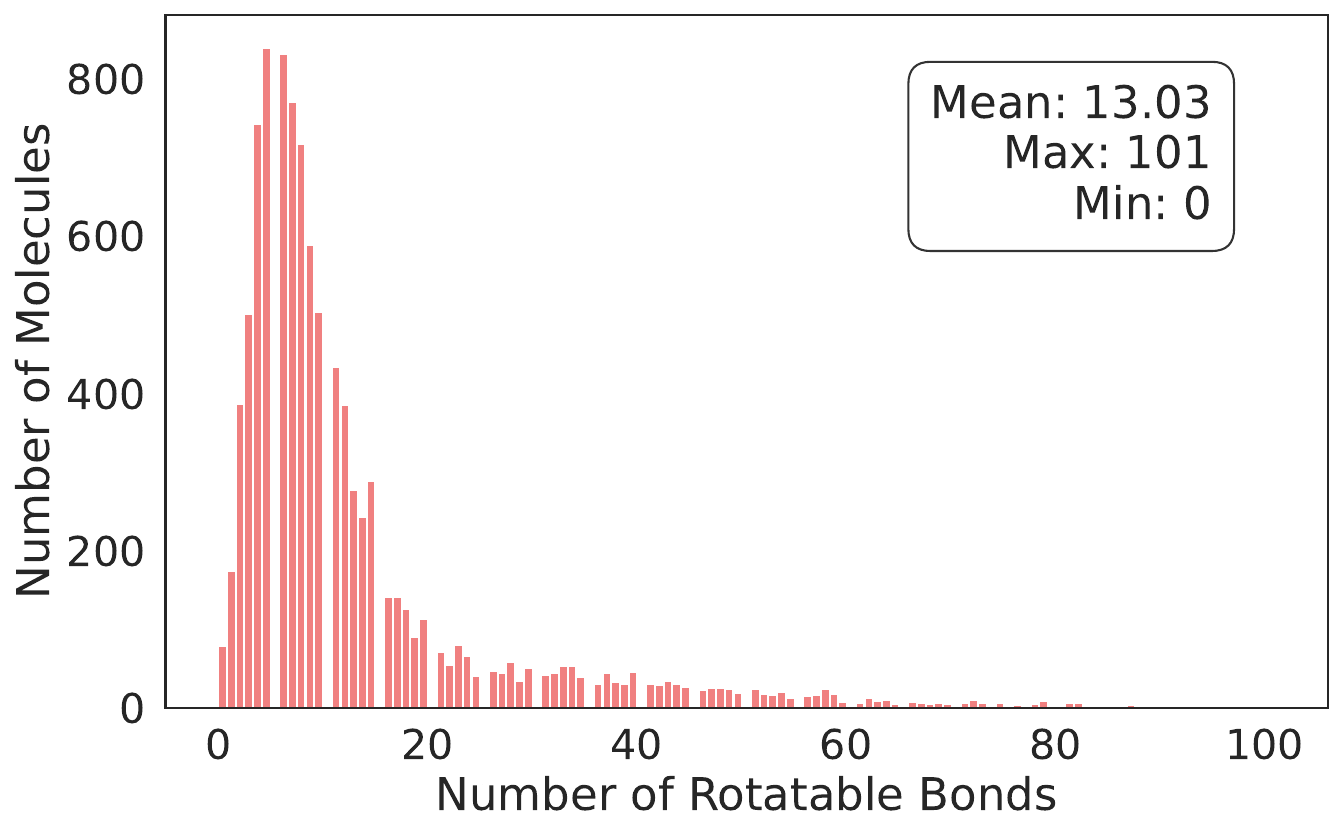}
        \label{fig:pdb-rot}
    \end{subfigure}
    \vfill
    \begin{subfigure}[t]{0.45\textwidth}
        \centering
        \caption{AQM Atoms}
        \includegraphics[height=3.5cm]{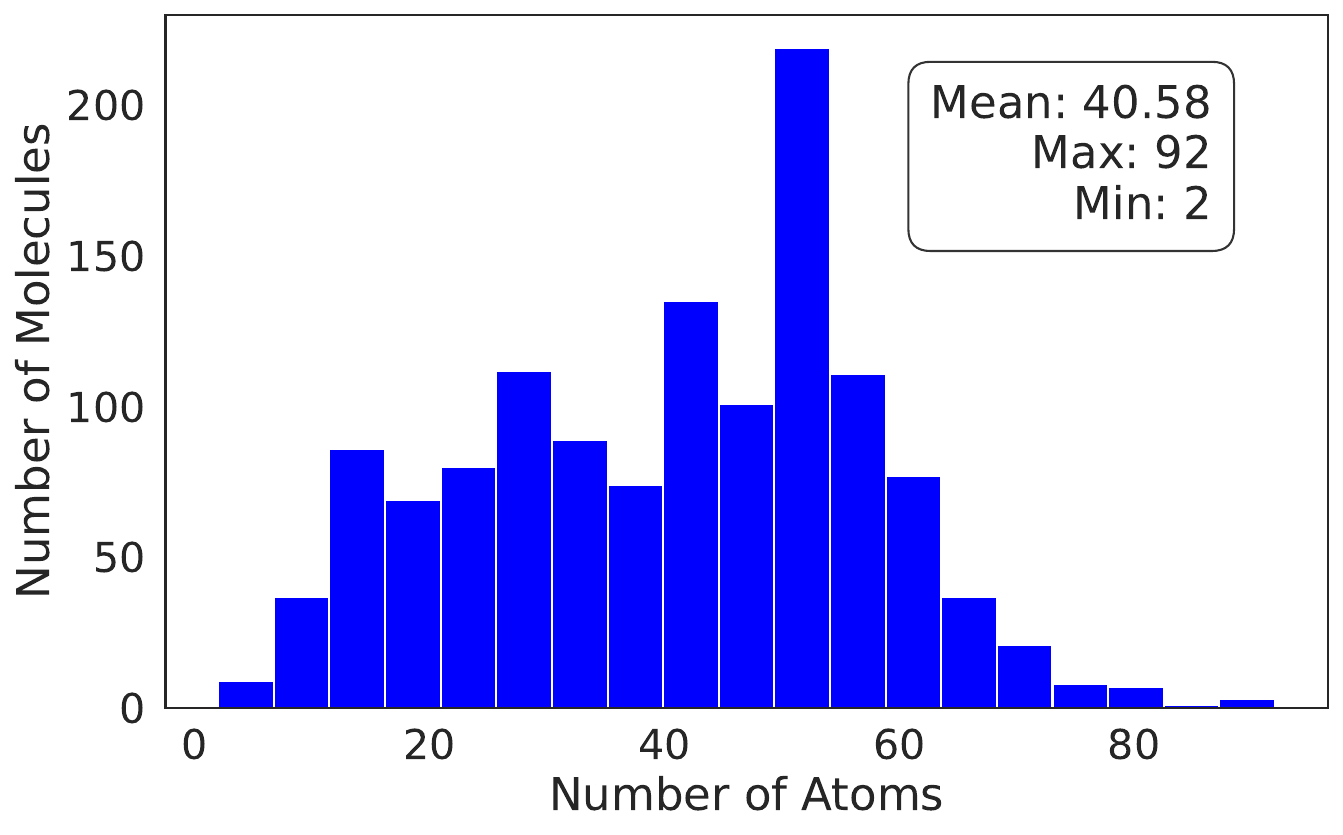}
        \label{fig:solv-atoms}
    \end{subfigure}
    \hfill
    \begin{subfigure}[t]{0.45\textwidth}
        \centering
        \caption{AQM Rotatable Bonds}
        \includegraphics[height=3.5cm]{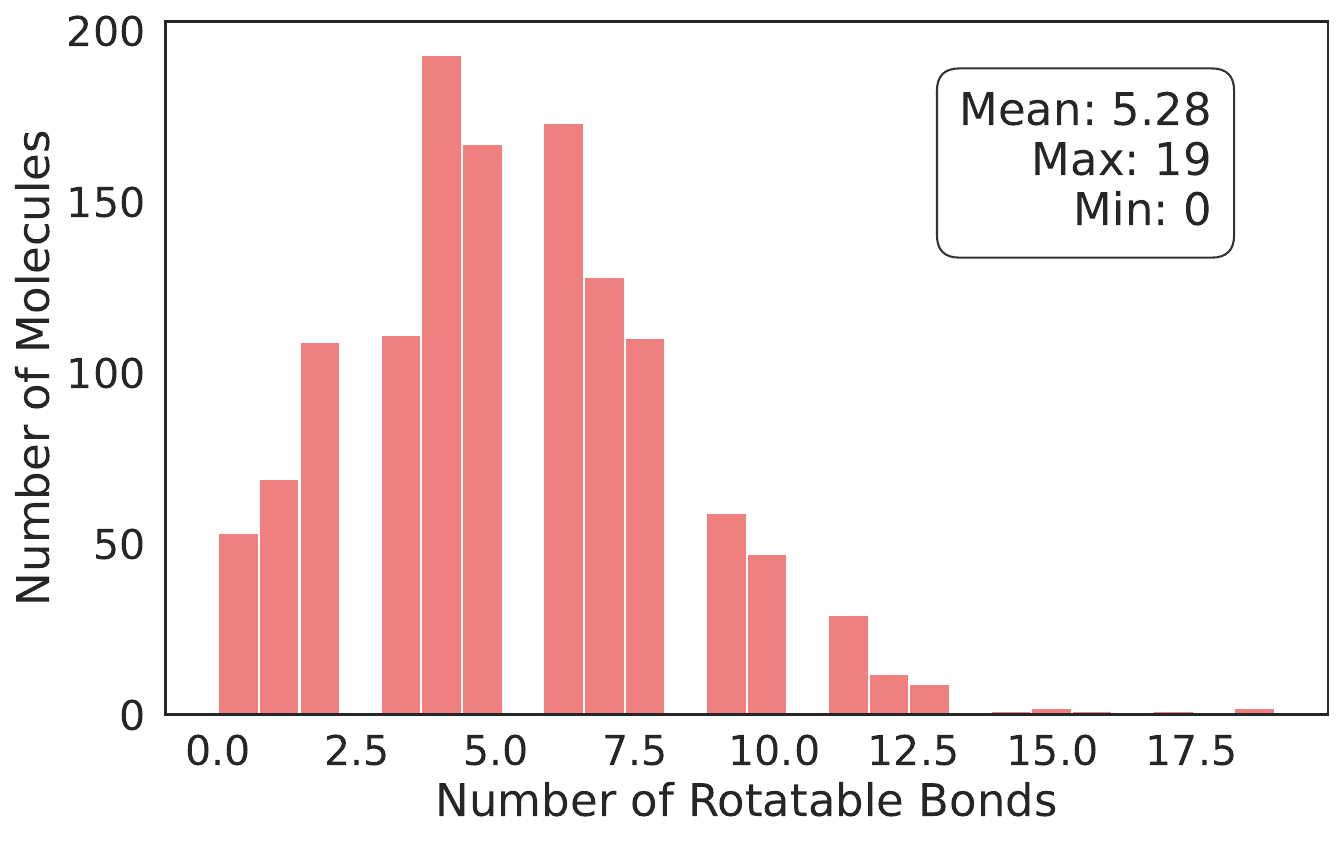}
        \label{fig:solv-rot}
    \end{subfigure}
    \vfill
    \begin{subfigure}[t]{0.45\textwidth}
        \centering
        \caption{COD Atoms}
        \includegraphics[height=3.5cm]{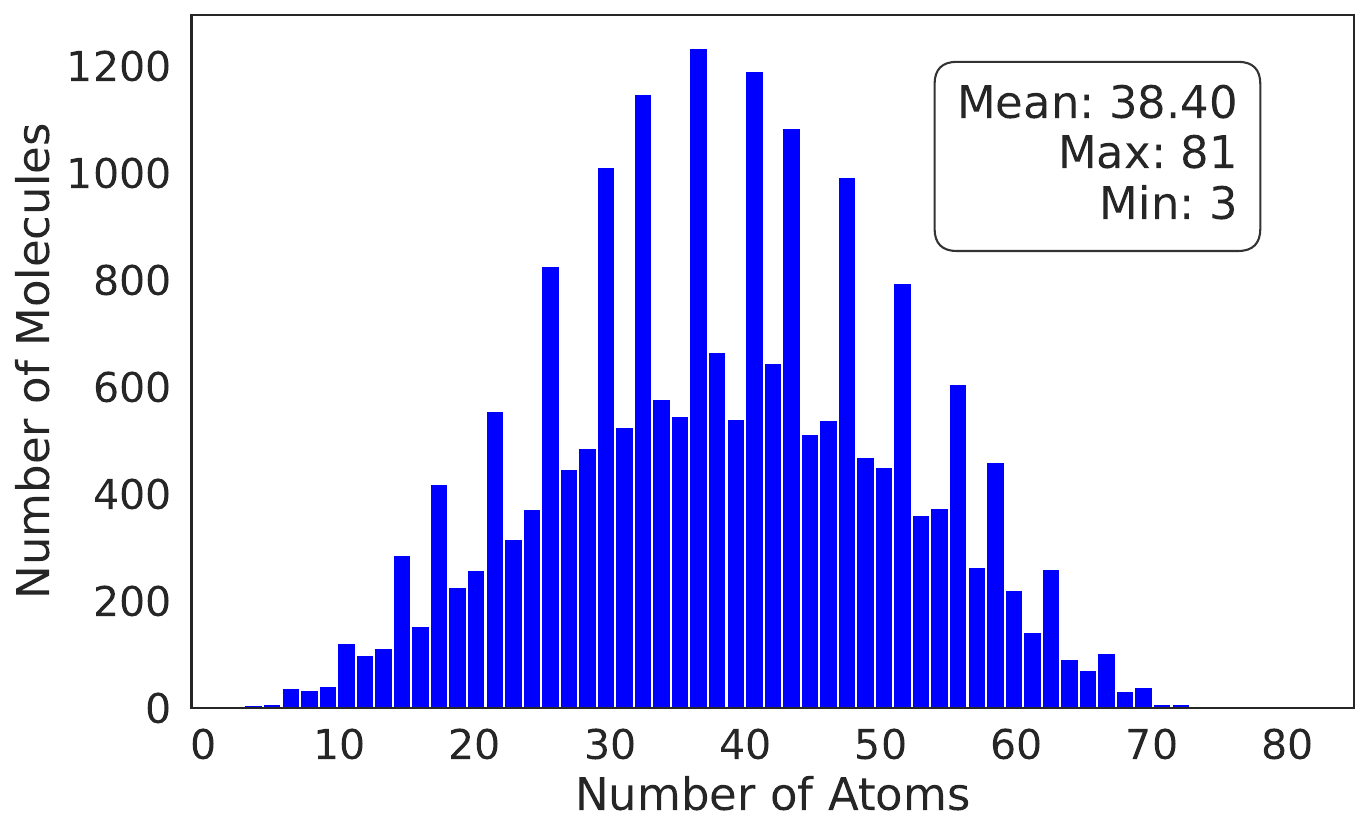}
        \label{fig:cod-atoms}
    \end{subfigure}
    \hfill
    \begin{subfigure}[t]{0.45\textwidth}
        \centering
        \caption{COD Rotatable Bonds}
        \includegraphics[height=3.5cm]{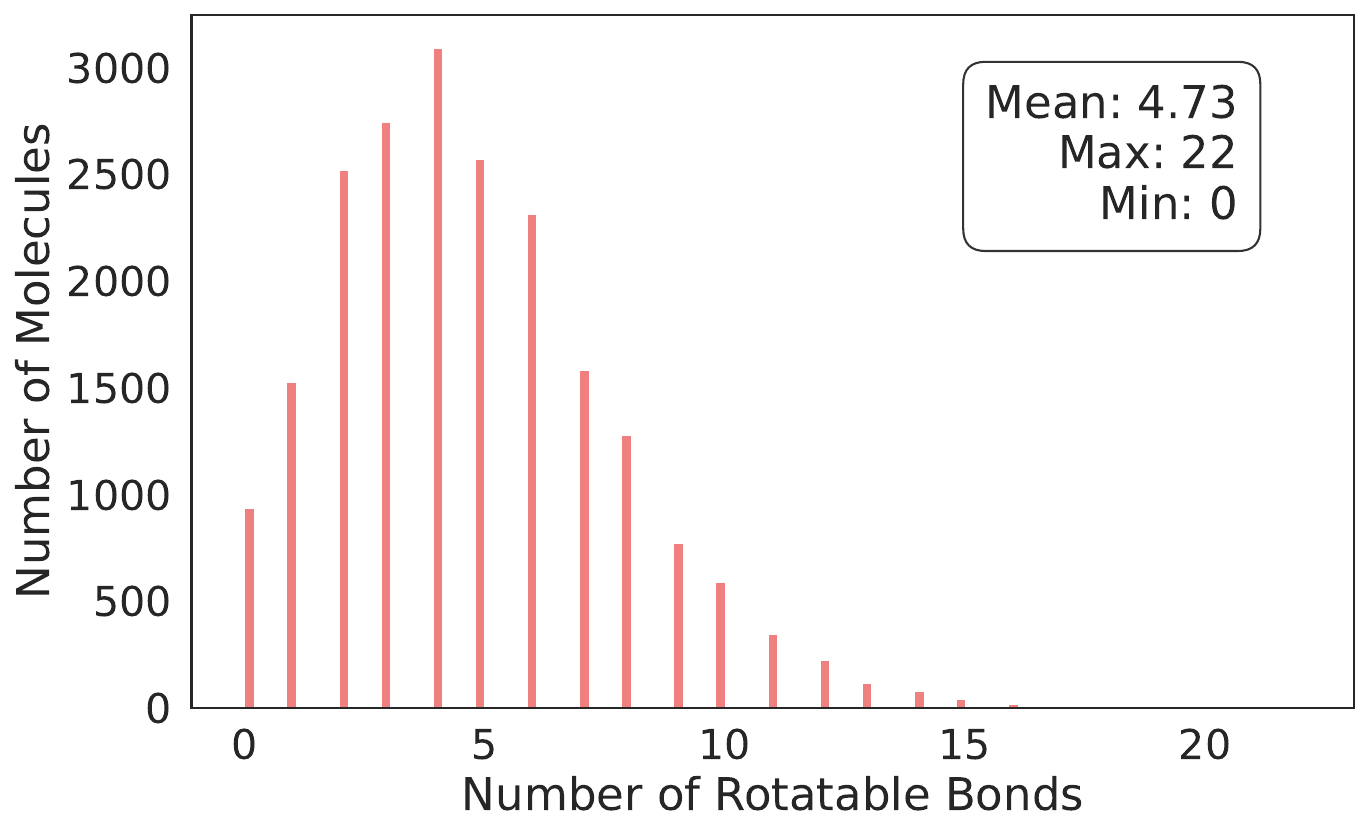}
        \label{fig:cod-rot}
    \end{subfigure}
\caption{Histograms of atom and rotatable-bond counts for PDBBind, AQM, and COD. GEOM DRUGS is shared with the pretraining datasets (Figure \ref{fig:hist_pretrain}).}
\label{fig:hist_finetune}
\end{figure}

To characterize the intrinsic difficulty of conformer generation across datasets, we report histograms of molecular size (number of atoms, $N_\mathrm{atoms}$) and flexibility (number of rotatable bonds, $N_\mathrm{rot}$) for each dataset (Figures \ref{fig:hist_pretrain} and \ref{fig:hist_finetune}). We compute $N_\mathrm{rot}$ using \texttt{rdMolDescriptors.CalcNumRotatableBonds} from RDKit with its default definition (non-ring single bonds between non-terminal heavy atoms).

Across the pretraining sets (GEOM Drugs, GEOM QM9, SPICE, CREST-relaxed SPICE, and OE62), $N_\mathrm{atoms}$ concentrates in the mid-size regime with means in the $\sim 17$--$44$ range, while $N_\mathrm{rot}$ is typically modest (most molecules have fewer than 15 rotatable bonds).

For the finetuning sets (GEOM Drugs, PDBBind, AQM, and COD), the atom-count distribution shifts larger with means in the $\sim 38-64$ range. In particular, the PDBBind dataset exhibits a long tail (very large molecules), indicating increased geometric and torsional complexity. 

\section{Training Details and Hyperparameters}
\label{appendix:training}

All models are trained on 8 NVIDIA A100 GPUs with AdamW and a cosine-annealing learning rate schedule with linear warmup. Table \ref{tab:hyper} lists the hyperparameters.

\textbf{Pretraining.}
We train for 800 epochs, resampling 100K examples per epoch from the pretraining pool with mixing probabilities 0.6 (GEOM-DRUGS), 0.2 (OE62), 0.1 (GEOM-QM9), 0.05 (SPICE), and 0.05 (CREST).

\textbf{Unified finetuning.}
Starting from the pretrained checkpoint, we finetune for 800 epochs using nominal allocations from a 100K-example epoch: 0.6 (Vacuum / GEOM-DRUGS), 0.2 (Packing / COD), 0.15 (Docking / PDBBind), and 0.05 (Solvation / AQM). Each allocation is capped at the available dataset size, giving $91{,}216$ effective samples per epoch.

\textbf{Individual finetuning.}
We finetune on each environment's full training set for 800 epochs using the same optimizer and schedule.

\begin{table}[!t]
    \centering
    \caption{Hyperparameters for EVA-Flow.}
    \label{tab:hyper}
    \begin{tabular}{lc}
    \toprule
    Hyperparameter & Value \\
    \midrule
    Batch size & 32 \\
    FM decoder layers & 20 \\
    Hidden channels & 160 \\
    Attention heads & 8 \\
    Latent dim ($d_z$) & 128 \\
    Environment dim ($d_e$) & 128 \\
    Encoder layers & 3 \\
    Base layers & 3 \\
    Environment layers & 3 \\
    Cutoff range & $[0.0,\, 10.0]$ \AA \\
    Activation & SiLU \\
    RBF type / count & ExpNorm / 64 \\
    Optimizer & AdamW \\
    AdamW $(\beta_1, \beta_2)$ & $(0.95,\, 0.999)$ \\
    Max LR & $5 \times 10^{-4}$ \\
    Min LR & $10^{-8}$ \\
    Weight decay & $10^{-8}$ \\
    Warmup steps & 31{,}250 \\
    Cosine cycle length & 312{,}500 steps \\
    Gradient clip (norm) & 1.0 \\
    Random seed & 42 \\
    Inference sampler / steps & SDE / 50 \\
    SDE noise ($s_{\mathrm{churn}}$, std) & $(1.0,\, 1.0)$ \\
    GPUs & $8 \times$ A100 \\
    \midrule
    Total parameters & 12.1M \\
    \bottomrule
    \end{tabular}
\end{table}

\subsection{Training, Inference, and Reproducibility Details}
\label{appendix:impl}

\textbf{Optimization and schedule.}
We use AdamW with $(\beta_1,\beta_2)=(0.95,\,0.999)$, weight decay $10^{-8}$, and gradient-norm clipping at $1.0$. The learning rate increases linearly over $31{,}250$ steps to a peak of $5\times10^{-4}$, then decays along a cosine schedule over a $312{,}500$-step cycle to a minimum of $10^{-8}$.

\textbf{Data splits, validation, and checkpointing.}
Each environment uses fixed train/validation/test splits. We validate every epoch and monitor the total validation loss (\texttt{val/total\_loss}); the model checkpoint with the lowest validation loss is used for all reported results.

\textbf{Random seeds.}
We use training seed $42$. Inference uses a stochastic SDE sampler. Table~\ref{tab:results} reports mean $\pm$ standard deviation over five independent sampling seeds from each fixed checkpoint; these values quantify inference-sampler variability, not training uncertainty. The separate all-conformer ETFlow diagnostic uses six runs.

\textbf{Inference sampler and noise.}
Unless otherwise noted, all reported numbers use the SDE sampler with $50$ integration steps, additive-noise scale $\mathrm{std}=1.0$, churn $s_{\mathrm{churn}}=1.0$, and a stochastic window $t\in[10^{-4},\,0.9999]$; the initial state $x_0$ is drawn from the harmonic prior with no additional perturbation. The deterministic ODE sampler is also supported and trades a small amount of Recall COV for speed (Vacuum: $0.737$ ODE vs.\ $0.747$ SDE).

The inference-time perturbation schedule is a sampling heuristic, distinct from the stochastic interpolation $\sigma\sqrt{t(1-t)}$ used to construct training paths. The term ``SDE sampler'' follows the implementation and does not imply that this procedure is the probability-flow SDE associated with the training interpolation.

\textbf{Inference time.}
Table~\ref{tab:timing} reports per-molecule inference time on a single A100 GPU (GEOM-DRUGS test set). EVA-Flow is modestly faster than retrained ETFlow at the same step count in our implementation; both use their respective graph-conditioned vector fields and the fixed harmonic initialization.

\begin{table}[!t]
    \centering
    \caption{Per-molecule inference time on a single A100 GPU (GEOM-DRUGS, $1{,}000$ molecules). EVA-Flow uses the Pretrain+Unified setup.}
    \label{tab:timing}
    \begin{tabular}{lcccc}
    \toprule
    Method & Sampler & Steps & R-COV $\uparrow$ & Time (s/mol) $\downarrow$ \\
    \midrule
    ETFlow (retrained) & SDE & 50 & 0.725 & 0.309 \\
    EVA-Flow (P+U) & SDE & 50 & 0.747 & 0.266 \\
    EVA-Flow (P+U) & SDE & 100 & 0.752 & 0.530 \\
    \bottomrule
    \end{tabular}
\end{table}

\textbf{Compute.}
Models are trained on $8\times$ NVIDIA A100 GPUs. Each $800$-epoch stage (pretraining and unified finetuning) takes roughly $5$ minutes per epoch, or approximately $67$ wall-clock hours ($\sim$$530$ A100-GPU-hours); individual finetuning is run per environment on its full training set.

\begin{samepage}
\textbf{Dataset preprocessing.}
Each dataset is processed by a dedicated script that will be released in the future.
Molecular graphs and features are built with \texttt{RDKit} and \texttt{datamol}; crystal lattices and space groups for Packing use \texttt{pymatgen}, with \texttt{OpenBabel} for bond perception from CIF files; numerical processing uses \texttt{NumPy} and \texttt{SciPy}.

\end{samepage}

\section{Evaluation Metric Definition}
\label{appendix:eval}

Following prior work~\citep{ganea2021geomol,xu2022geodiff,jing2022torsional}, we evaluate generated conformers using average minimum RMSD (AMR) and Coverage (COV), each measured for Recall and Precision. Recall measures the extent to which the generated conformers capture the ground-truth conformers, while Precision indicates the proportion of generated
conformers that match the ground-truth conformers. 
Let $C_g$ and $C_r$ denote the generated and reference conformation sets, respectively. The metrics are:
\begin{equation}
    \text{AMR-R}(C_g, C_r) = \frac{1}{|C_r|}\sum_{\mathbf{R} \in C_r} \min_{\hat{\mathbf{R}}\in C_g} \mathrm{RMSD} (\mathbf{R},\hat{\mathbf{R}})
\end{equation}
\begin{equation}
    \text{COV-R}(C_g, C_r) = \frac{1}{|C_r|}|\{\mathbf{R} \in C_r | \mathrm{RMSD}(\mathbf{R},\mathbf{\hat{R}})<\delta, \hat{\mathbf{R}} \in C_g\}|
\end{equation}
\begin{equation}
    \text{AMR-P}(C_r, C_g) = \frac{1}{|C_g|}\sum_{\mathbf{\hat{\mathbf{R}}} \in C_g} \min_{\mathbf{R}\in C_r} \mathrm{RMSD} (\hat{\mathbf{R}}, \mathbf{R})
\end{equation}
\begin{equation}
    \text{COV-P}(C_r, C_g) = \frac{1}{|C_g|}|\{\hat{\mathbf{R}} \in C_g | \mathrm{RMSD}(\mathbf{\hat{R}}, \mathbf{R})<\delta, \mathbf{R} \in C_r\}|
\end{equation}

A lower AMR signifies closer matches. Recall COV measures reference coverage, whereas Precision COV measures the fraction of generated samples close to a finite reference set; neither alone establishes sample diversity or calibrated probabilities. The threshold $\delta$ is set to $0.5$\,\AA\ for Solvation, $0.75$\,\AA\ for Vacuum~\citep{hassan2024etflow} and Packing, and $2.0$\,\AA\ for Docking~\citep{corso2022diffdock}.

Recall and Precision must be read jointly because Recall grows with sample budget while Precision penalizes samples that do not match the finite reference set.

\section{Additional Results}
\label{appendix:results}

\subsection{Additional Precision Results}
\begin{figure}[t]
    \centering
    \begin{subfigure}[t]{0.48\linewidth}
        \centering
        \caption{Vacuum}
        \includegraphics[width=\linewidth]{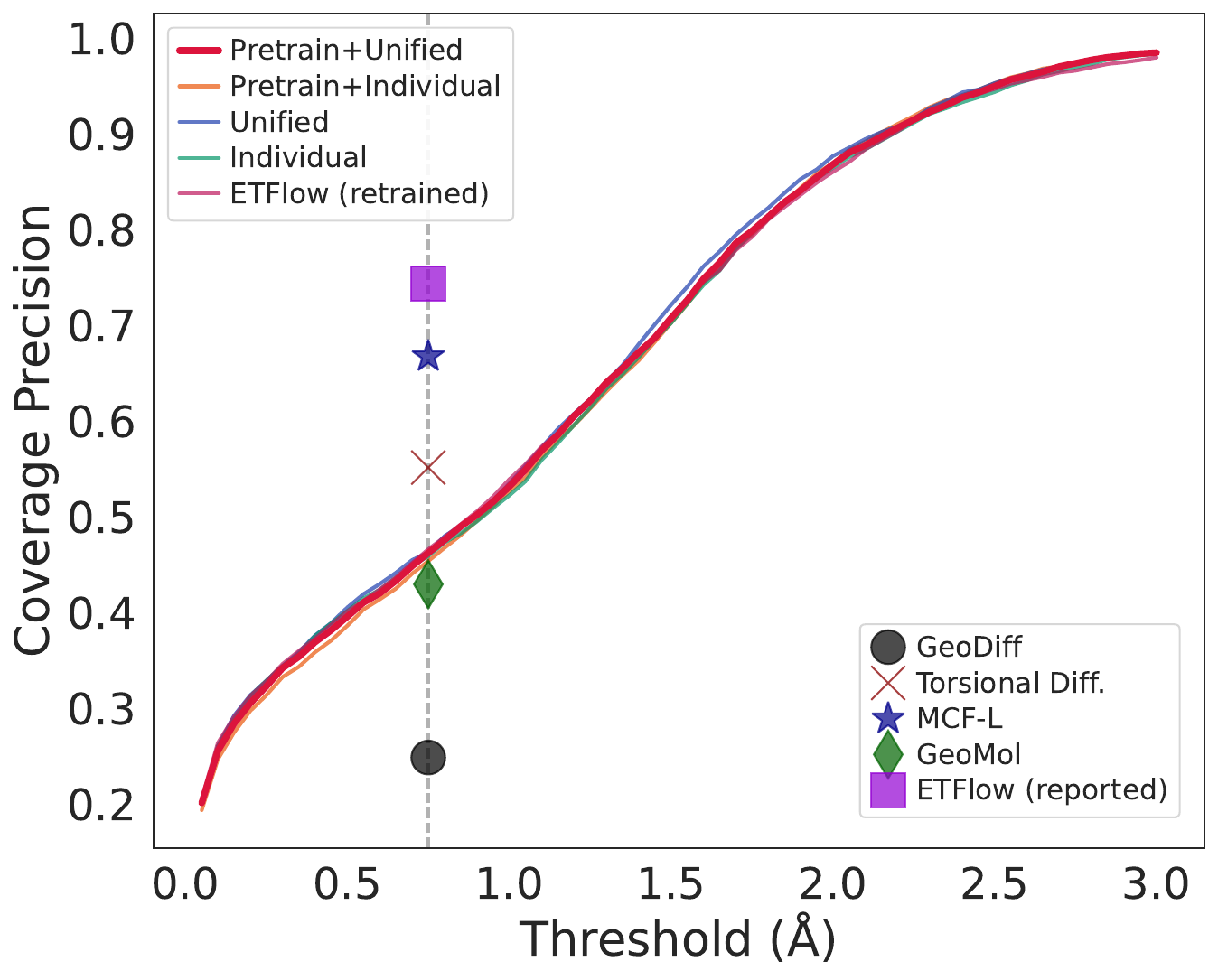}
        \label{fig:drugs_precision_scaling}
    \end{subfigure}\hfill
    \begin{subfigure}[t]{0.48\linewidth}
        \centering
        \caption{Docking}
        \includegraphics[width=\linewidth]{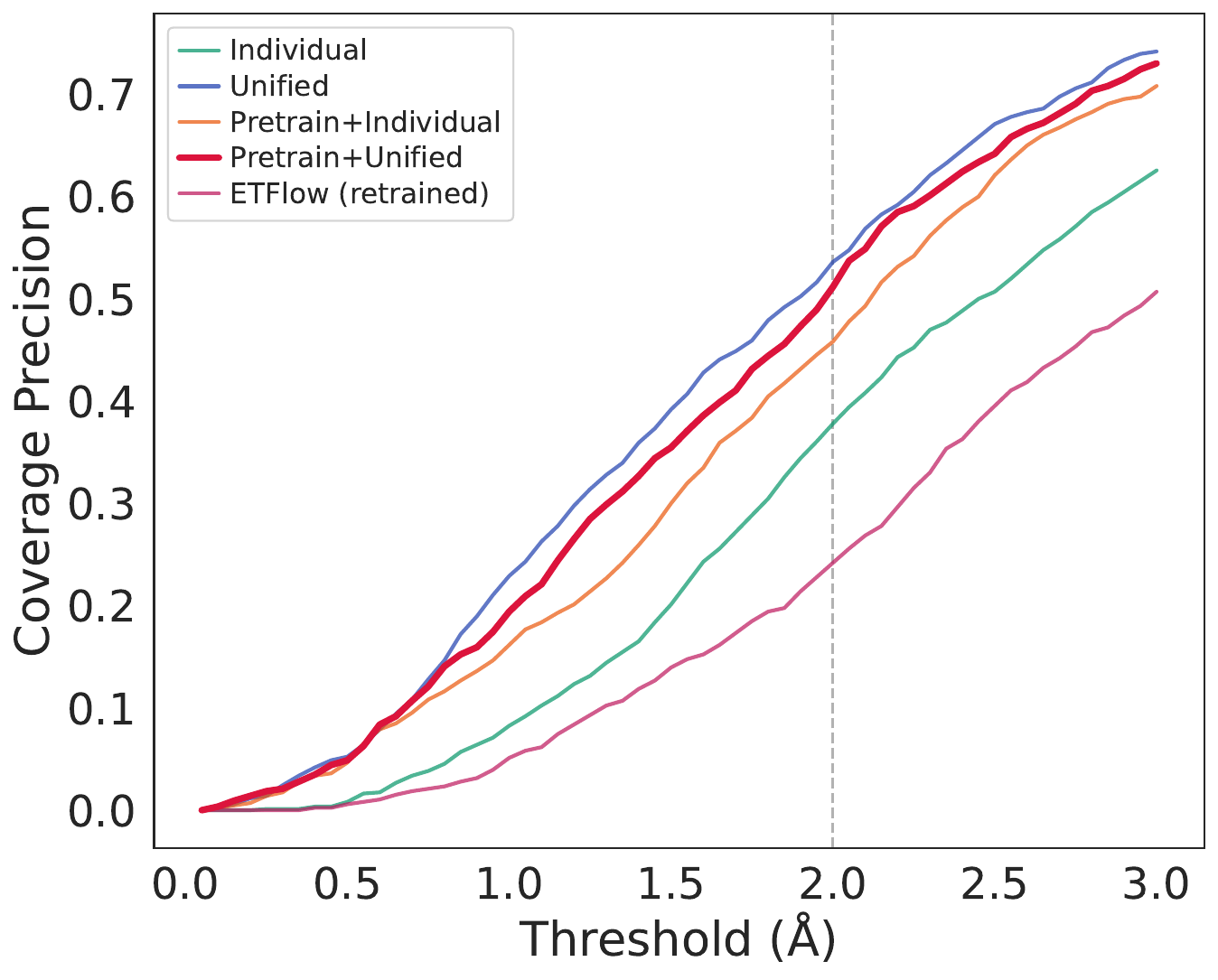}
        \label{fig:dock_precision_scaling}
    \end{subfigure}

    \vspace{0.6em}

    \begin{subfigure}[t]{0.48\linewidth}
        \centering
        \caption{Solvation}
        \includegraphics[width=\linewidth]{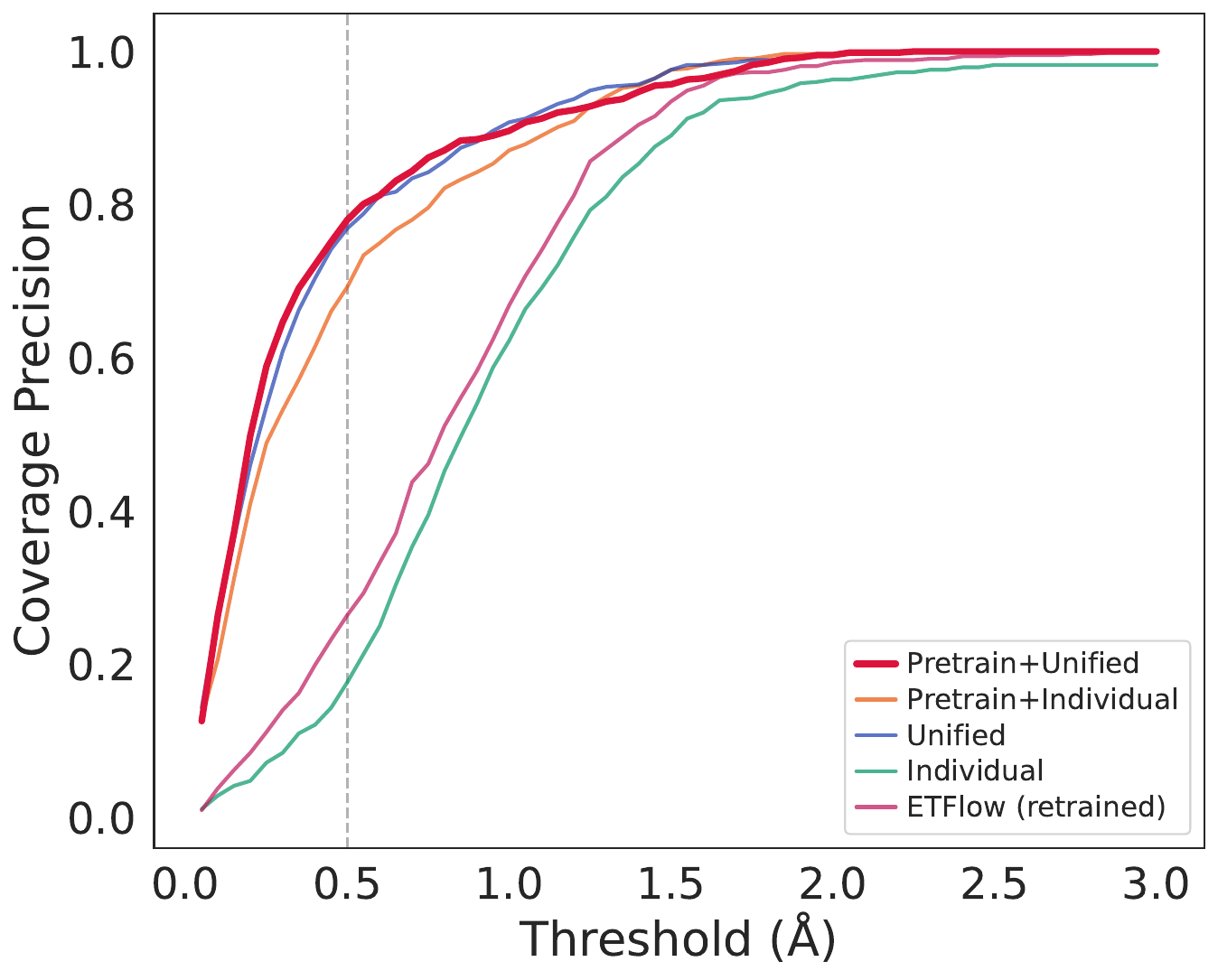}
        \label{fig:solv_precision_scaling}
    \end{subfigure}\hfill
    \begin{subfigure}[t]{0.48\linewidth}
        \centering
        \caption{Packing}
        \includegraphics[width=\linewidth]{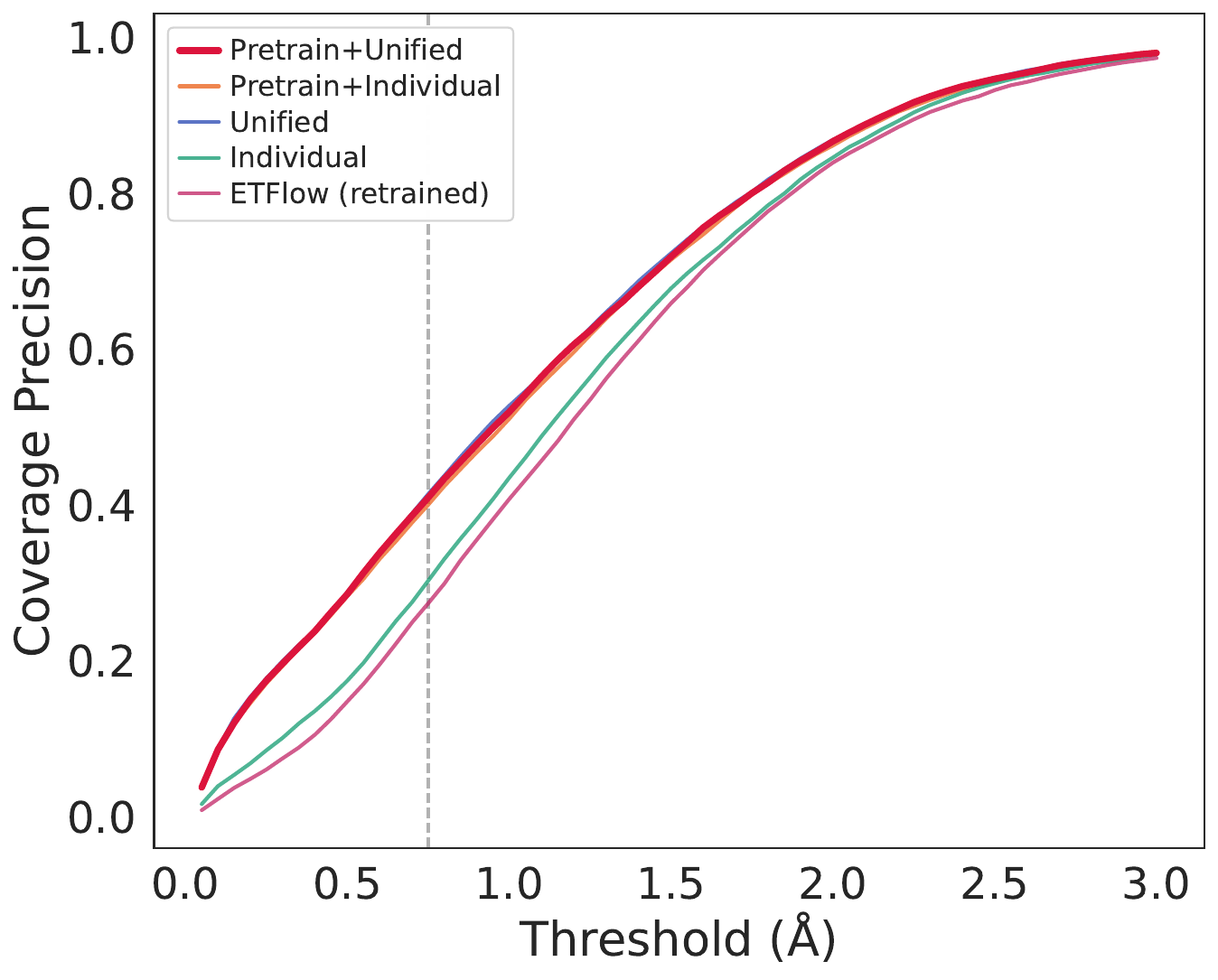}
        \label{fig:pack_precision_scaling}
    \end{subfigure}
\caption{Precision Coverage for all four environments: Vacuum, Docking, Solvation, and Packing.
}
\label{fig:results_precision}
\end{figure}

Precision Coverage plots are shown in Figure~\ref{fig:results_precision}.

\subsection{Median Coverage under the 4-Conformer Protocol}
\label{appendix:median-cov}

Table~\ref{tab:median-cov} reports median Coverage. With one reference and four generated conformers per molecule, per-molecule Recall COV is binary and Precision COV is restricted to $\{0,0.25,0.5,0.75,1\}$. Consequently, the median values are strongly quantized and often identical across models; mean COV and median AMR in Table~\ref{tab:results} provide substantially finer discrimination.

\begin{table}[t]
\centering
\caption{Median Recall and Precision COV under the 4-conformer protocol.}
\label{tab:median-cov}
\resizebox{\linewidth}{!}{%
\begin{tabular}{llcc}
\toprule
Environment & Setup & R-COV median $\uparrow$ & P-COV median $\uparrow$ \\
\midrule
\multirow{5}{*}{Vacuum}
 & Pretrain+Unified & 1.00 & 0.50 \\
 & Pretrain+Individual & 1.00 & 0.50 \\
 & Unified & 1.00 & 0.50 \\
 & Individual & 1.00 & 0.50 \\
 & ETFlow (retrained) & 1.00 & 0.50 \\
\midrule
\multirow{5}{*}{Docking}
 & Pretrain+Unified & 1.00 & 0.50 \\
 & Pretrain+Individual & 1.00 & 0.25 \\
 & Unified & 1.00 & 0.50 \\
 & Individual & 1.00 & 0.25 \\
 & ETFlow (retrained) & 0.00 & 0.00 \\
\midrule
\multirow{5}{*}{Solvation}
 & Pretrain+Unified & 1.00 & 1.00 \\
 & Pretrain+Individual & 1.00 & 0.75 \\
 & Unified & 1.00 & 1.00 \\
 & Individual & 0.00 & 0.00 \\
 & ETFlow (retrained) & 1.00 & 0.25 \\
\midrule
\multirow{5}{*}{Packing}
 & Pretrain+Unified & 1.00 & 0.25 \\
 & Pretrain+Individual & 1.00 & 0.25 \\
 & Unified & 1.00 & 0.25 \\
 & Individual & 1.00 & 0.25 \\
 & ETFlow (retrained) & 0.00 & 0.00 \\
\bottomrule
\end{tabular}}
\end{table}

\subsection{Generation Diversity}
\label{appendix:diversity}

To test for sample collapse, we measure the mean symmetry-aware heavy-atom best-RMS over all unique pairs of the four generated conformers per molecule, averaged over the test set (Pretrain+Unified model). Table~\ref{tab:diversity} reports non-zero spread in every environment, establishing non-trivial sample diversity across all four settings.

\begin{table}[t]
\centering
\caption{Pairwise-RMSD diversity (\AA) among the four generated conformers per molecule (Pretrain+Unified).}
\label{tab:diversity}
\begin{tabular}{lccc}
\toprule
Environment & Mean $\uparrow$ & Median & Std \\
\midrule
Vacuum    & 0.915 & 0.903 & 0.620 \\
Docking   & 2.227 & 1.828 & 1.508 \\
Solvation & 0.456 & 0.359 & 0.315 \\
Packing   & 0.967 & 0.912 & 0.634 \\
\bottomrule
\end{tabular}
\end{table}

\subsection{Multi-Reference Solvation Evaluation}
\label{appendix:solv-multiref}

Docking and Packing remain single-reference structure-prediction evaluations. The multi-reference evaluation below concerns Solvation.

Aquamarine test molecules contain a mean of $37$ conformers (up to $431$), although the default training and 4-conformer evaluation use \texttt{c1}. We therefore evaluate against all available references with the $2K$ protocol (Table~\ref{tab:solv-multiref}). Recall and Precision are reported jointly: high Recall demonstrates broad coverage of the available ensemble, while Precision penalizes generations that do not match that finite reference set. The high Recall ($0.981$) shows the single-conformer-trained model covers the available solvated ensemble.

\begin{table}[t]
\centering
\caption{Solvation evaluation with single- and multi-reference protocols.}
\label{tab:solv-multiref}
\resizebox{\linewidth}{!}{%
\begin{tabular}{llrrcccc}
\toprule
Training & Protocol & Refs/mol & Samples/mol & R-COV $\uparrow$ & R-AMR $\downarrow$ & P-COV $\uparrow$ & P-AMR $\downarrow$ \\
\midrule
Single conformer & 4-conf. & 1 & 4 & 0.936 & 0.157 & 0.693 & 0.431 \\
Single conformer & 2K & mean 37 & $2K$ & 0.981 & 0.102 & 0.541 & 0.591 \\
Multiple conformers & 2K & mean 37 & $2K$ & 0.962 & 0.111 & 0.518 & 0.612 \\
\bottomrule
\end{tabular}}
\end{table}

\subsection{Physical Reference Methods}
\label{appendix:physical-baselines}

We add two non-learned reference methods under the same four-sample evaluation budget: RDKit ETKDG followed by MMFF relaxation for all environments, and GFN2-xTB with ALPB-water implicit solvation for Solvation. These are physical reference points rather than controlled ablations of environment conditioning because they differ from EVA-Flow in model class, data, and optimization.

\begin{table}[t]
\centering
\caption{Non-learned reference methods under the 4-conformer protocol.}
\label{tab:physical-baselines}
\begin{tabular}{llcc}
\toprule
Environment & Method & R-COV $\uparrow$ & R-AMR $\downarrow$ \\
\midrule
Vacuum & ETKDG+MMFF & 0.409 & 1.18 \\
       & EVA-Flow (P+U) & 0.747 & 0.468 \\
Docking & ETKDG+MMFF & 0.581 & 2.09 \\
        & EVA-Flow (P+U) & 0.684 & 1.899 \\
Solvation & ETKDG+MMFF & 0.006 & 3.42 \\
          & GFN2-xTB+ALPB-water & 0.006 & 3.38 \\
          & EVA-Flow (P+U) & 0.949 & 0.153 \\
Packing & ETKDG+MMFF & 0.606 & 0.72 \\
        & EVA-Flow (P+U) & 0.644 & 0.664 \\
\bottomrule
\end{tabular}
\end{table}

\subsection{Reconciling Published and Retrained ETFlow Results}
\label{appendix:etflow-reconciliation}

\paragraph{Baseline scope and training.}
\begin{samepage}
For Solvation and Packing, no established learned baselines share our task definitions and splits; we report retrained ETFlow under identical evaluation. For Docking, retrained ETFlow is selected by validation loss after $100$ epochs on the same PDBBind split, molecular inputs, sample budget, and evaluator. ETFlow has no published internal-conformation result under this protocol. This is a matched ligand-only reference, but a shared schedule does not guarantee architecture-specific optimal tuning; the inference-time NoPocket ablation (Appendix~\ref{appendix:envsens}) more directly tests whether EVA-Flow uses pocket information.

\end{samepage}

\paragraph{Vacuum comparisons.}
On Vacuum under the 2K protocol (Table~\ref{tab:results-2k}), EVA-Flow is comparable to retrained ETFlow (Recall COV 0.721 vs.\ 0.712; Precision COV 0.707 vs.\ 0.704) and remains behind the 20$\times$-larger MCF-L on Recall (0.721 vs.\ 0.847; 12.1M vs.\ 242M parameters), while exceeding MCF-L on Precision COV (0.707 vs.\ 0.668). EVA-Flow is not a new state of the art for Vacuum.

The ETFlow number reported in its paper uses the original GEOM-DRUGS benchmark, all available reference conformers, and a $2K$ sampling budget. This differs from our default Table~\ref{tab:results}, which uses the chemically cleaned benchmark, one reference, and four samples. Table~\ref{tab:etflow-reconciliation} evaluates the published checkpoint and retrained ETFlow/EVA-Flow models through our evaluation code. The original and cleaned test sets share $998/1{,}000$ molecules, but differ in preprocessing and reference counts (up to $2{,}497$ versus a cap of $30$), which also changes the $2K$ sample budget. Comparisons within a matched block support model comparisons; values across datasets or implementation pipelines are diagnostic only and are not used for causal attribution.

\begin{table*}[t]
\centering
\caption{Matched $2K$ evaluation ($0.75$\,\AA) on original and cleaned GEOM-DRUGS test sets. All rows within a block use identical evaluation code and test references.}
\label{tab:etflow-reconciliation}
\resizebox{\linewidth}{!}{%
\begin{tabular}{lllcccc}
\toprule
Model & Training set & Test set & R-COV $\uparrow$ & R-AMR $\downarrow$ & P-COV $\uparrow$ & P-AMR $\downarrow$ \\
\midrule
Published ETFlow checkpoint & Original & Original & 0.549 & 0.766 & 0.339 & 1.057 \\
ETFlow (retrained) & Original & Original & 0.712 & 0.562 & 0.704 & 0.594 \\
EVA-Flow & Original & Original & \textbf{0.721} & \textbf{0.550} & \textbf{0.707} & \textbf{0.589} \\
\midrule
Published ETFlow checkpoint & Original & Cleaned & 0.517 & 0.799 & 0.294 & 1.193 \\
ETFlow (retrained) & Original & Cleaned & 0.759 & 0.505 & 0.629 & 0.688 \\
EVA-Flow & Original & Cleaned & \textbf{0.763} & \textbf{0.503} & \textbf{0.632} & \textbf{0.686} \\
\midrule
ETFlow (retrained) & Cleaned & Cleaned & 0.798 & 0.443 & 0.728 & 0.532 \\
EVA-Flow & Cleaned & Cleaned & \textbf{0.810} & \textbf{0.415} & \textbf{0.740} & \textbf{0.505} \\
\bottomrule
\end{tabular}}
\end{table*}

\section{Physical Validity of Generated Conformers}
\label{appendix:uma}

Coverage and AMR measure geometric similarity to reference conformers but not energetic plausibility. To assess physical validity directly, we relax all $4{,}000$ generated Vacuum conformers ($1{,}000$ molecules $\times\,4$ conformers) with UMA~\citep{wood2026uma}, a large-scale pretrained molecular potential, and measure the energy change and geometric shift upon relaxation (Table~\ref{tab:uma}). The very small energy changes and atomic displacements indicate that EVA-Flow generates conformers near local energy minima under UMA.

\begin{table}[!t]
    \centering
    \caption{UMA relaxation of EVA-Flow Vacuum conformers ($4{,}000$ conformers from $1{,}000$ molecules).}
    \label{tab:uma}
    \begin{tabular}{lc}
    \toprule
    Metric & Value \\
    \midrule
    UMA relaxation converged & 99.9\% (3{,}995/4{,}000) \\
    Mean $\Delta E$ after relaxation & 0.302 kcal/mol \\
    Median $\Delta E$ after relaxation & 0.115 kcal/mol \\
    Fraction with $\Delta E < 1$ kcal/mol & 97.1\% \\
    Mean RMSD shift during relaxation & 0.056 \AA \\
    Fraction with RMSD shift $< 0.1$ \AA & 84.2\% \\
    \bottomrule
    \end{tabular}
\end{table}

\section{Shared-Molecule Controls}
\label{appendix:overlap-scaled}

Shared molecules are matched by canonical SMILES and aligned with consistent heavy-atom mappings before RMSD calculation. Table~\ref{tab:overlap-heldout} reports the control in which all $1{,}201$ molecules occurring in at least two finetuning environments are removed from every finetuning training set. For pair $A/B$, own $A/B$ compares generations conditioned on each context with that context's reference; cross $A/B$ swaps the reference; and generation shift compares the two context-conditioned generations. All entries are mean heavy-atom RMSD in \AA. Because the broad pretraining corpus includes Vacuum molecules, this control isolates finetuning exposure but does not guarantee pretraining decontamination.

\begin{table*}[t]
\centering
\caption{Shared-molecule analysis after excluding all shared molecules from finetuning.}
\label{tab:overlap-heldout}
\begin{tabular}{lrrrrrr}
\toprule
Pair ($A/B$) & $n$ & own $A$ & own $B$ & cross $A$ & cross $B$ & generation shift \\
\midrule
Solvation--Vacuum & 142 & 0.29 & 0.90 & 3.35 & 3.25 & 3.32 \\
Docking--Vacuum & 558 & 0.95 & 0.81 & 1.17 & 1.19 & 1.16 \\
Vacuum--Packing & 487 & 0.38 & 0.43 & 0.64 & 0.59 & 0.62 \\
Docking--Solvation & 13 & 0.63 & 0.35 & 2.34 & 2.27 & 2.23 \\
Packing--Solvation & 23 & 0.14 & 0.14 & 1.48 & 1.47 & 1.50 \\
Packing--Docking & 64 & 0.36 & 0.46 & 0.51 & 0.37 & 0.40 \\
\bottomrule
\end{tabular}
\end{table*}

The own-versus-cross pattern remains for pairs with well-separated references, while Packing--Docking stays weakly separated because those reference geometries are physically similar. These held-out data show the conditioning effect persists on molecules never seen in training, ruling out finetuning memorization as the explanation.

\section{Environment-Graph Sensitivity}
\label{appendix:envsens}

We measure the performance change after zeroing domain-specific environment features at inference time (Table~\ref{tab:envsens}). \textbf{NoSolvDesc} zeros the four solvation descriptors; \textbf{NoLattice} zeros the seven lattice/space-group parameters; \textbf{NoPocket} zeros the entire environment embedding for docking. All three ablations degrade performance, confirming that the model relies on each environment component, and the docking pocket is the most critical.

\begin{table}[!t]
    \centering
    \caption{Environment-feature sensitivity (inference-time zeroing, no retraining). Thresholds: Solvation $0.5$\,\AA, Packing $0.75$\,\AA, Docking $2.0$\,\AA.}
    \label{tab:envsens}
    \begin{tabular}{llcccc}
    \toprule
    Environment & Ablation & R-COV $\uparrow$ & R-AMR $\downarrow$ & P-COV $\uparrow$ & P-AMR $\downarrow$ \\
    \midrule
    \multirow{2}{*}{Solvation} & Full & \textbf{0.949} & \textbf{0.153} & \textbf{0.780} & \textbf{0.367} \\
                               & NoSolvDesc & 0.905 & 0.191 & 0.734 & 0.399 \\
    \midrule
    \multirow{2}{*}{Packing}   & Full & \textbf{0.644} & \textbf{0.664} & \textbf{0.410} & \textbf{1.073} \\
                               & NoLattice & 0.575 & 0.762 & 0.360 & 1.154 \\
    \midrule
    \multirow{2}{*}{Docking}   & Full & \textbf{0.684} & \textbf{1.899} & \textbf{0.512} & \textbf{2.471} \\
                               & NoPocket & 0.400 & 2.590 & 0.257 & 3.197 \\
    \bottomrule
    \end{tabular}
\end{table}

\section{Full Ablation Metrics}
\label{appendix:ablation-full}

Replacing the FM decoder with regression targets the conditional mean of a multi-modal distribution, explaining the collapse in Recall COV for NoFM.

\begin{samepage}
Figure~\ref{fig:ablation} reports Recall COV for the original ablations. Table~\ref{tab:ablation-full} provides their complete metrics across all four environments. NoBase reduces performance within the original base--latent parameterization; the full latent-free training-regime grid (Appendix~\ref{appendix:latent-free-grid}) shows that removing both the learned base and latent (the latent-free variant) has mixed effects under Individual training but trails the full model in all twelve cross-environment transfer comparisons.

\end{samepage}

\begin{table}[!t]
    \centering
    \caption{Full ablation metrics (Unified setup, small model). Thresholds: Vacuum/Packing $0.75$\,\AA, Solvation $0.5$\,\AA, Docking $2.0$\,\AA. Best per metric/environment in \textbf{bold}.}
    \label{tab:ablation-full}
    \begin{tabular}{llcccc}
    \toprule
    Metric & Setup & Vacuum & Docking & Solvation & Packing \\
    \midrule
    \multirow{6}{*}{R-COV $\uparrow$}
      & EVA (Full) & \textbf{0.740} & \textbf{0.674} & \textbf{0.930} & \textbf{0.640} \\
      & GCN        & 0.670 & 0.665 & 0.803 & 0.605 \\
      & Gaussian   & 0.405 & 0.237 & 0.822 & 0.312 \\
      & NoConsist  & 0.338 & 0.577 & 0.134 & 0.373 \\
      & NoBase     & 0.321 & 0.577 & 0.255 & 0.398 \\
      & NoFM       & 0.000 & 0.126 & 0.006 & 0.005 \\
    \midrule
    \multirow{6}{*}{P-COV $\uparrow$}
      & EVA (Full) & \textbf{0.463} & \textbf{0.536} & \textbf{0.769} & \textbf{0.414} \\
      & GCN        & 0.396 & 0.490 & 0.497 & 0.384 \\
      & Gaussian   & 0.241 & 0.179 & 0.618 & 0.201 \\
      & NoConsist  & 0.184 & 0.379 & 0.102 & 0.206 \\
      & NoBase     & 0.177 & 0.376 & 0.140 & 0.220 \\
      & NoFM       & 0.000 & 0.057 & 0.003 & 0.002 \\
    \midrule
    \multirow{6}{*}{R-AMR $\downarrow$}
      & EVA (Full) & \textbf{0.456} & \textbf{1.889} & \textbf{0.166} & \textbf{0.682} \\
      & GCN        & 0.623 & 2.153 & 0.285 & 0.714 \\
      & Gaussian   & 1.078 & 4.528 & 0.291 & 1.248 \\
      & NoConsist  & 1.108 & 2.221 & 0.896 & 0.990 \\
      & NoBase     & 1.113 & 2.160 & 0.790 & 0.963 \\
      & NoFM       & 2.136 & 3.536 & 1.735 & 2.214 \\
    \midrule
    \multirow{6}{*}{P-AMR $\downarrow$}
      & EVA (Full) & \textbf{0.954} & \textbf{2.391} & \textbf{0.370} & \textbf{1.067} \\
      & GCN        & 1.117 & 2.937 & 0.593 & 1.117 \\
      & Gaussian   & 1.551 & 5.414 & 0.492 & 1.601 \\
      & NoConsist  & 1.569 & 2.941 & 0.921 & 1.400 \\
      & NoBase     & 1.560 & 2.711 & 0.902 & 1.363 \\
      & NoFM       & 2.502 & 4.020 & 2.145 & 2.581 \\
    \bottomrule
    \end{tabular}
\end{table}

\section{Latent-Free Model: Full Training-Regime Grid}
\label{appendix:latent-free-grid}

To directly test whether the base and latent are necessary, we train the latent-free variant (environment embedding + fixed harmonic base + FM decoder; both the learned base and the VAE latent removed) in all four training regimes and evaluate every environment. Table~\ref{tab:e4} reports Recall COV for the latent-free model against the full model (Table~\ref{tab:results}) in each cell. Under Individual training, full-model to latent-free Recall COV changes from $0.728$ to $0.727$ in Vacuum, $0.553$ to $0.470$ in Docking, $0.414$ to $0.503$ in Solvation, and $0.512$ to $0.493$ in Packing. Thus, performance is nearly unchanged in Vacuum, improves in Solvation, and declines in Docking and Packing. In contrast, the latent-free model trails in all three cross-environment transfer regimes (Unified and the two pretraining-based settings): every one of the twelve transfer cells favors the full model, with the gap concentrated in the low-data Solvation environment ($-0.16$ to $-0.26$ Recall COV, several times the sampler standard deviation). These results support a consistent benefit from the coupled base and latent in the tested cross-environment transfer regimes; their contribution under Individual training depends on the environment.

\begin{table}[!t]
\centering
\caption{Full architecture (reported model): Recall COV at the environment threshold across all four training regimes.}
\label{tab:full-training-grid}
\begin{tabular}{lrrrr}
\toprule
Environment & Individual & Unified & Pretrain+Ind. & Pretrain+Unified \\
\midrule
Vacuum & 0.728 & 0.740 & 0.733 & 0.747 \\
Docking & 0.553 & 0.674 & 0.656 & 0.684 \\
Solvation & 0.414 & 0.930 & 0.936 & 0.949 \\
Packing & 0.512 & 0.640 & 0.633 & 0.644 \\
\bottomrule
\end{tabular}
\end{table}

\begin{table}[!t]
\centering
\caption{Latent-free model (learned base and latent removed): Recall COV at the environment threshold across all four training regimes. Compared with the full model (Table~\ref{tab:full-training-grid}), Individual performance is nearly unchanged in Vacuum, higher in Solvation, and lower in Docking and Packing. All twelve transfer comparisons favor the full model, with the largest gaps in Solvation.}
\label{tab:e4}
\begin{tabular}{lrrrr}
\toprule
Environment & Individual & Unified & Pretrain+Ind. & Pretrain+Unified \\
\midrule
Vacuum & 0.727 & 0.708 & 0.688 & 0.672 \\
Docking & 0.470 & 0.660 & 0.614 & 0.628 \\
Solvation & 0.503 & 0.675 & 0.771 & 0.771 \\
Packing & 0.493 & 0.622 & 0.615 & 0.613 \\
\bottomrule
\end{tabular}
\end{table}

\begin{samepage}
\section{Data Exposure and Balanced Sampling}
\label{appendix:exposure}

To test whether the Unified advantage stems from cross-environment transfer rather than data volume, we analyze total molecule exposure and run a balanced-sampling control. Table~\ref{tab:exposure} (top) counts molecule samples over the stated $800$ epochs. Unified starts from nominal $60/20/5/15\%$ allocations of $100$K samples, then caps each environment at its dataset size, yielding $91{,}216$ effective samples per epoch; Individual models pass over their full dataset each epoch. Unified sees far fewer DRUGS samples than Individual ($48.0$M vs.\ $218.1$M) yet outperforms it; for the small-data environments the Individual models already overfit (validation loss plateaus by epoch $\sim$30--50), so additional repetitions alone cannot explain the gains.

\end{samepage}

Table~\ref{tab:exposure} (bottom) compares balanced ($25/25/25/25\%$) against the default weighted sampling, both from the same pretrained checkpoint. Solvation and Docking see the \emph{same} number of molecules per epoch under both schemes (capped at dataset size), yet Solvation drops $10.7\%$ when the DRUGS allocation falls from $60\%$ to $25\%$, indicating that DRUGS data benefits Solvation through cross-environment transfer in the shared backbone, not through Solvation data volume.

\begin{table}[!t]
    \centering
    \caption{Top: total molecule exposure across training. Bottom: balanced vs.\ default sampling (Recall COV at the environment threshold).}
    \label{tab:exposure}
    \begin{tabular}{lrrrrr}
    \toprule
    Environment & Dataset & Unified/ep & Unified tot. & Indiv./ep & Indiv. tot. \\
    \midrule
    Vacuum (DRUGS) & 272{,}641 & 60{,}000 & 48.0M & 272{,}641 & 218.1M \\
    Packing (COD)  & 20{,}765  & 20{,}000 & 16.0M & 20{,}765  & 16.6M \\
    Solvation (AQM)& 1{,}276   & 1{,}276  & 1.02M & 1{,}276   & 1.02M \\
    Docking (PDB)  & 9{,}940   & 9{,}940  & 7.95M & 9{,}940   & 7.95M \\
    \bottomrule
    \end{tabular}

    \vspace{0.6em}
    \begin{tabular}{lccc}
    \toprule
    Environment & Default (60/20/5/15) & Balanced (25/25/25/25) & Change \\
    \midrule
    Vacuum (0.75\,\AA)    & 0.740 & 0.692 & $-6.5\%$ \\
    Docking (2.0\,\AA)    & 0.684 & 0.670 & $-2.0\%$ \\
    Solvation (0.5\,\AA)  & 0.949 & 0.847 & $-10.7\%$ \\
    Packing (0.75\,\AA)   & 0.644 & 0.643 & $-0.2\%$ \\
    \bottomrule
    \end{tabular}
\end{table}

\section{Train--Test Consistency of the Base Distribution}
\label{appendix:base-consistency}

At inference the initial coordinates are drawn from the simple harmonic prior rather than the learned training-time base. The measured divergence at convergence is $0.036$ nats (validation; $0.026$ train). On $200$ test molecules, the learned-to-prior eigenvalue ratio has median $1.0000$ and all measured modes are within $1\%$ of unity, with posterior mean approximately zero. A noise-injection sweep also degrades gradually (Table~\ref{tab:noise}). These diagnostics indicate that the inference-time base mismatch is small. The large NoBase drop in the original parameterization is therefore an optimization interaction, not an inference-time mismatch: its latent activates while the full-model latent collapses (Table~\ref{tab:base-latent}), whereas the effect of the latent-free model under Individual training varies by environment (Table~\ref{tab:e4}). The base and latent are thus essential for low-data cross-environment transfer while remaining inactive at inference.

\begin{table}[!t]
    \centering
    \caption{Inference noise-injection sweep on Vacuum (Pretrain+Unified checkpoint, SDE $50$ steps): perturbing the harmonic prior by $\epsilon\cdot\mathcal{N}(\mathbf{0}, I)$.}
    \label{tab:noise}
    \begin{tabular}{lcccc}
    \toprule
    $\epsilon$ & R-COV $\uparrow$ & R-AMR $\downarrow$ & P-COV $\uparrow$ & P-AMR $\downarrow$ \\
    \midrule
    0.00 (default) & \textbf{0.747} & \textbf{0.468} & \textbf{0.464} & \textbf{0.979} \\
    0.01 & 0.738 & 0.470 & 0.455 & 0.983 \\
    0.05 & 0.735 & 0.493 & 0.459 & 0.981 \\
    0.10 & 0.745 & 0.471 & 0.464 & 0.973 \\
    0.20 & 0.730 & 0.486 & 0.452 & 0.985 \\
    0.50 & 0.734 & 0.480 & 0.453 & 0.982 \\
    1.00 & 0.704 & 0.529 & 0.413 & 1.079 \\
    \bottomrule
    \end{tabular}
\end{table}

\begin{table}[!t]
\centering
\caption{Latent activation in the full and NoBase models.}
\label{tab:base-latent}
\begin{tabular}{lcccc}
\toprule
Model & Active dims & KL/node (nats) & Pooled $\|\mu\|$ & Environment-dependent \\
\midrule
Full & 0/128 & $\approx10^{-6}$ & $\approx10^{-3}$ & No (cosine 1.000) \\
NoBase & 3/128 & up to $\sim3$ & $\approx0.21$ & Yes \\
\bottomrule
\end{tabular}
\end{table}

\section{Latent Posterior-Collapse Diagnostics}
\label{appendix:latent-diagnostics}

\begin{samepage}
We diagnose whether the variational latent $z$ carries information in the Pretrain+Unified checkpoint. We compute the per-node latent KL, posterior statistics, and active dimensions (a dimension is active when the variance across data of its posterior mean exceeds $10^{-2}$), then rerun generation after replacing sampled $z$ by zero while preserving all subsequent random draws. Table~\ref{tab:latent-collapse} reports the paired random-versus-zero intervention using one matched diagnostic sampling seed, rather than the five-seed means in Table~\ref{tab:results}. The posterior is effectively collapsed: Vacuum, Packing, and Solvation have mean KL approximately $1.5\times10^{-6}$ nats per node; Docking has a larger mean due to a small tail but the same near-zero median; no dimensions are active. For the same molecule across all six environment pairs, the pooled posterior mean has norm approximately $10^{-3}$ and is unchanged across contexts (cosine $1.000$, relative $L_2\leq0.001$). Setting $z=0$ leaves all metrics effectively unchanged. Thus the decoder primarily conditions on $G$ and $e$, and we do not attribute its performance to information carried by $z$. The NoConsist degradation is instead interpreted as an optimization effect of that objective; the latent-free variant shows that $z$ can be removed along with the learned base.

\end{samepage}

\begin{table*}[!t]
\centering
\caption{Latent-collapse diagnostics and intervention. ``Random'' samples $z\sim\mathcal{N}(\mathbf{0}, I)$; ``zero'' replaces the matched latent draw by zero. COV values use the environment-specific thresholds.}
\label{tab:latent-collapse}
\resizebox{\linewidth}{!}{%
\begin{tabular}{lcccccccc}
\toprule
Environment & KL/node mean & KL/node median & Active dims & Posterior $|\mu|$ & R-COV random & R-COV zero & P-COV random & P-COV zero \\
\midrule
Vacuum & $1.5{\times}10^{-6}$ & $1.4{\times}10^{-6}$ & 0/128 & $8.2{\times}10^{-5}$ & 0.707 & 0.706 & 0.433 & 0.433 \\
Docking & 0.0476 & $1.4{\times}10^{-6}$ & 0/128 & $1.9{\times}10^{-4}$ & 0.661 & 0.661 & 0.497 & 0.497 \\
Solvation & $1.6{\times}10^{-6}$ & $1.5{\times}10^{-6}$ & 0/128 & $8.2{\times}10^{-5}$ & 0.803 & 0.809 & 0.580 & 0.591 \\
Packing & $1.6{\times}10^{-6}$ & $1.5{\times}10^{-6}$ & 0/128 & $8.2{\times}10^{-5}$ & 0.649 & 0.649 & 0.419 & 0.418 \\
\bottomrule
\end{tabular}}
\end{table*}

\section{Loss-Weight Sensitivity}
\label{appendix:loss-sensitivity}

We use unit weights because the implemented losses are normalized per node/batch and have comparable scale.

We vary one regularizer weight at a time by $0.1\times$ and $10\times$, keeping the FM weight and training schedule fixed, and report the Recall-COV change from the unit-weight control (Table~\ref{tab:loss-sensitivity}). In $21/24$ cells, $|\Delta|<0.045$. The larger changes are Vacuum with latent-KL $10\times$ ($-0.045$), Solvation with consistency $0.1\times$ ($-0.064$), and Solvation with latent-KL $10\times$ ($-0.141$). Thus most cells are stable to large one-at-a-time changes, while Solvation is sensitive to particular latent-regularization settings. This experiment does not vary network capacity or interactions among weights.

\begin{table}[!t]
\centering
\caption{One-factor-at-a-time loss sensitivity: change in Recall COV from unit weights.}
\label{tab:loss-sensitivity}
\begin{tabular}{lrrrr}
\toprule
Weight change & Vacuum & Docking & Solvation & Packing \\
\midrule
latent-KL $0.1\times$ & $-0.018$ & $-0.010$ & $-0.020$ & $-0.007$ \\
latent-KL $10\times$ & $-0.045$ & $+0.018$ & $-0.141$ & $-0.030$ \\
base-KL $0.1\times$ & $-0.027$ & $+0.004$ & $+0.019$ & $-0.008$ \\
base-KL $10\times$ & $-0.014$ & $+0.032$ & $+0.019$ & $-0.015$ \\
consistency $0.1\times$ & $-0.018$ & $+0.009$ & $-0.064$ & $-0.010$ \\
consistency $10\times$ & $-0.004$ & $+0.028$ & $+0.025$ & $-0.001$ \\
\bottomrule
\end{tabular}
\end{table}

\section{Environment-Encoder Equivariance}
\label{appendix:env-equivariance}

We chose the GCN environment encoder as a common interface because some contexts have physical coordinates (pockets and molecular/crystal proxies) whereas others are descriptor-only; for Docking, we jointly rotate ligand and pocket during training to reduce frame dependence.

We replace the GCN environment encoder with a matched SE(3)-equivariant TorchMD encoder under the same controlled Unified training budget (Table~\ref{tab:env-equivariance}). Accuracy improves for Docking, whose context carries molecule-specific coordinates, while the encoder is neutral or worse for contexts dominated by topology, proxies, or scalar descriptors. This is an encoder comparison, not a claim that one architecture is uniformly superior.

\begin{table*}[!t]
\centering
\caption{Equivariant environment encoder minus matched GCN control. Positive $\Delta$R-COV and negative $\Delta$R-AMR favor the equivariant encoder.}
\label{tab:env-equivariance}
\begin{tabular}{llrr}
\toprule
Environment & Coordinate content of $G_E$ & $\Delta$R-COV & $\Delta$R-AMR \\
\midrule
Vacuum & zero-coordinate scaffold & $-0.003$ & $-0.011$ \\
Docking & physical pocket coordinates & $+0.028$ & $-0.082$ \\
Solvation & fixed proxy; scalar descriptors & $-0.153$ & $+0.139$ \\
Packing & RDKit proxy and lattice metadata & $-0.006$ & $+0.015$ \\
\bottomrule
\end{tabular}
\end{table*}

For a separate joint rigid-rotation stress test over all $218$ Docking test complexes, the GCN environment embedding changes by $1.33\%$ on average ($0.98\%$ median relative error), with mean cosine similarity $0.99998$ (minimum $0.99983$). The equivariant encoder changes by approximately $10^{-5}$ and has cosine similarity $1.0000$. This measures embedding stability, not end-to-end equivariance of generated coordinates.

\section{Aquamarine Conformer-Selection Sensitivity}
\label{appendix:aqm-selection}

The default protocol selects the stored \texttt{c1} conformer (array index 0), not an energy-ranked ground state. To quantify reference sensitivity, we hold the trained model, molecule split, sampling seed, and four-conformer budget fixed while changing the test reference to either the minimum-\texttt{eSOLV} conformer or one deterministic random conformer per molecule (Table~\ref{tab:aqm-selection}). We separately train matched 200-epoch models with each target rule (Table~\ref{tab:aqm-training-selection}). These controlled runs use a smaller budget than the 800-epoch model and are compared only within their block.

\begin{table}[!t]
\centering
\caption{Aquamarine test-reference sensitivity for a fixed model and sampling seed ($0.5$\,\AA{} threshold).}
\label{tab:aqm-selection}
\begin{tabular}{lcccc}
\toprule
Reference selection & R-COV $\uparrow$ & R-AMR $\downarrow$ & P-COV $\uparrow$ & P-AMR $\downarrow$ \\
\midrule
\texttt{c1} (default) & 0.803 & 0.291 & 0.580 & 0.538 \\
Minimum \texttt{eSOLV} & 0.777 & 0.319 & 0.588 & 0.538 \\
Deterministic random & 0.803 & 0.292 & 0.584 & 0.538 \\
\bottomrule
\end{tabular}
\end{table}

\begin{table}[!t]
\centering
\caption{Aquamarine training-target sensitivity in matched 200-epoch runs.}
\label{tab:aqm-training-selection}
\begin{tabular}{lc}
\toprule
Training-target selection & R-COV $\uparrow$ \\
\midrule
\texttt{c1} (default rule) & 0.580 \\
Minimum \texttt{eSOLV} & 0.561 \\
Deterministic random & 0.618 \\
\bottomrule
\end{tabular}
\end{table}

\section{Molecule Visualization}
\label{appendix:vis}

Table~\ref{tab:vis-more} visualizes reference and generated conformers for molecules shared between Packing and Docking.

\begin{figure}[t]
    \centering
    \begin{subfigure}[t]{0.48\linewidth}
        \centering
        \caption{Solvation-Docking}
        \includegraphics[width=\linewidth]{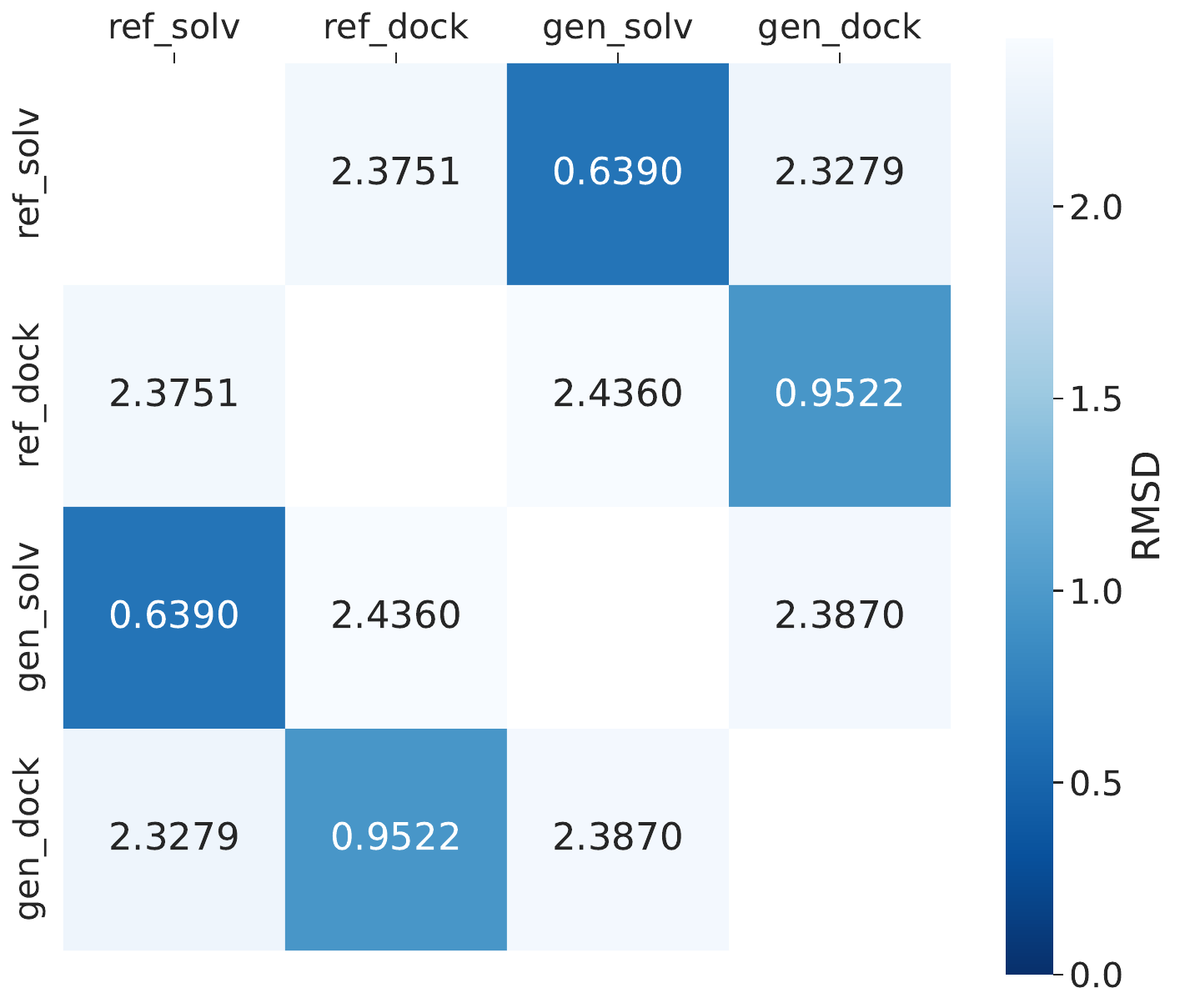}
        \label{fig:solv-dock}
    \end{subfigure}
    \begin{subfigure}[t]{0.48\linewidth}
        \centering
        \caption{Packing-Docking}
        \includegraphics[width=\linewidth]{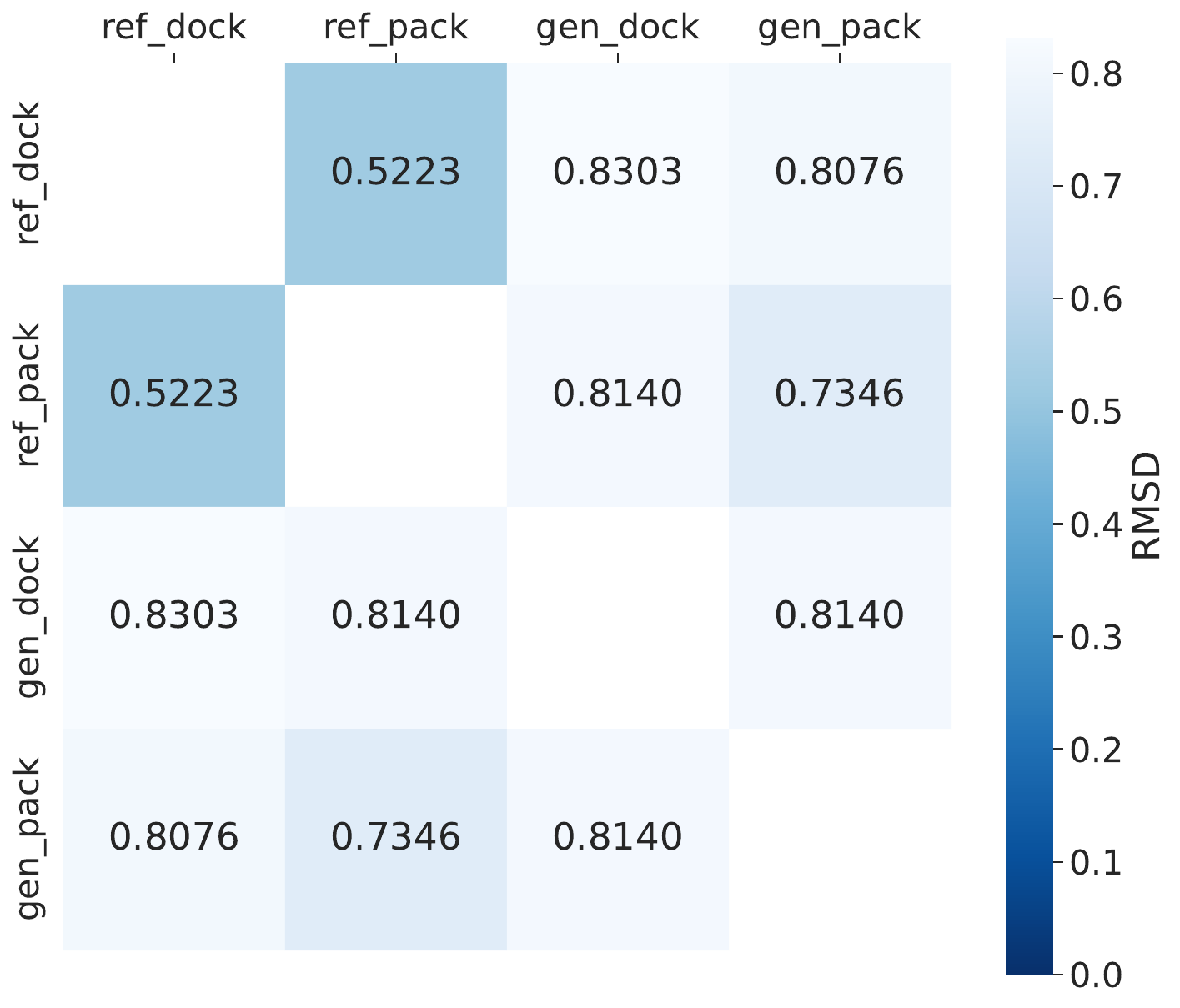}
        \label{fig:pack-dock}
    \end{subfigure}
    \vspace{-0.3cm}
\caption{Pairwise heavy-atom RMSD (\text{\AA}) between reference and generated conformers in the original overlap sets: (a) Solvation-Docking ($n=12$), and (b) Packing-Docking ($n=59$). In each heatmap we report the mean RMSD after alignment for: ref\_A $\leftrightarrow$ ref\_B, gen\_A $\leftrightarrow$ ref\_A, gen\_A $\leftrightarrow$ ref\_B, gen\_B $\leftrightarrow$ ref\_A, gen\_B $\leftrightarrow$ ref\_B, and gen\_A $\leftrightarrow$ gen\_B (A/B = the two tasks in the panel). Darker cells indicate lower RMSD (better agreement).}
\label{fig:overlap2}
\end{figure}

\begin{table}[t]
    \centering
    \caption{Visualization of reference and generated conformations for the shared molecules in Packing-Docking. The molecules are aligned to the Docking reference conformers.}
    \label{tab:vis-more}
    \begin{tabular}{cccc}
    \toprule
    \multicolumn{4}{c|}{Packing-Docking} \\
    \midrule
    ref\_dock & gen\_dock & ref\_pack & gen\_pack \\
    \midrule
    \includegraphics[trim={2cm 1cm 2cm 1cm},clip,height=2cm]{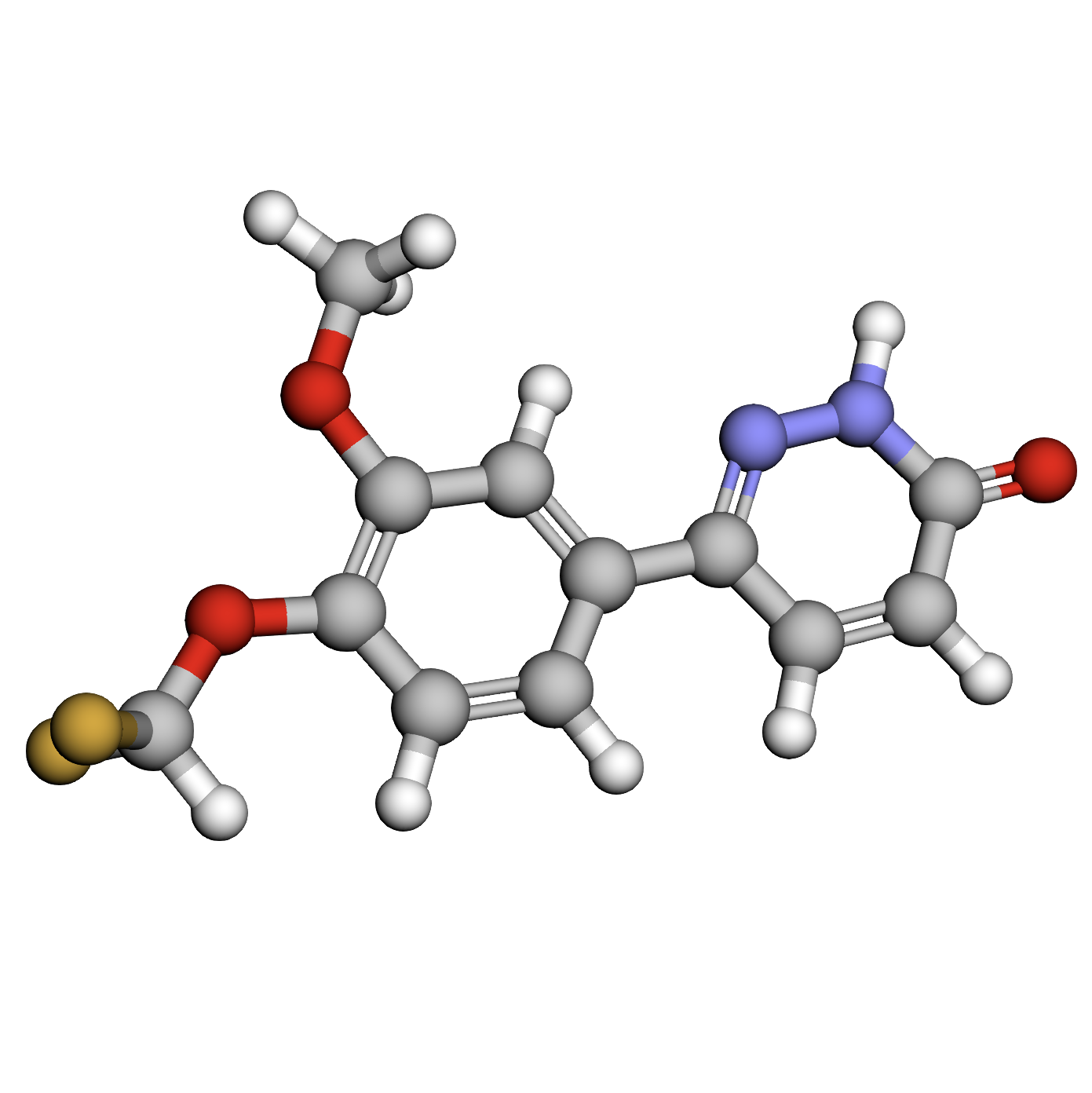} & \includegraphics[trim={2cm 1cm 2cm 1cm},clip,height=2cm]{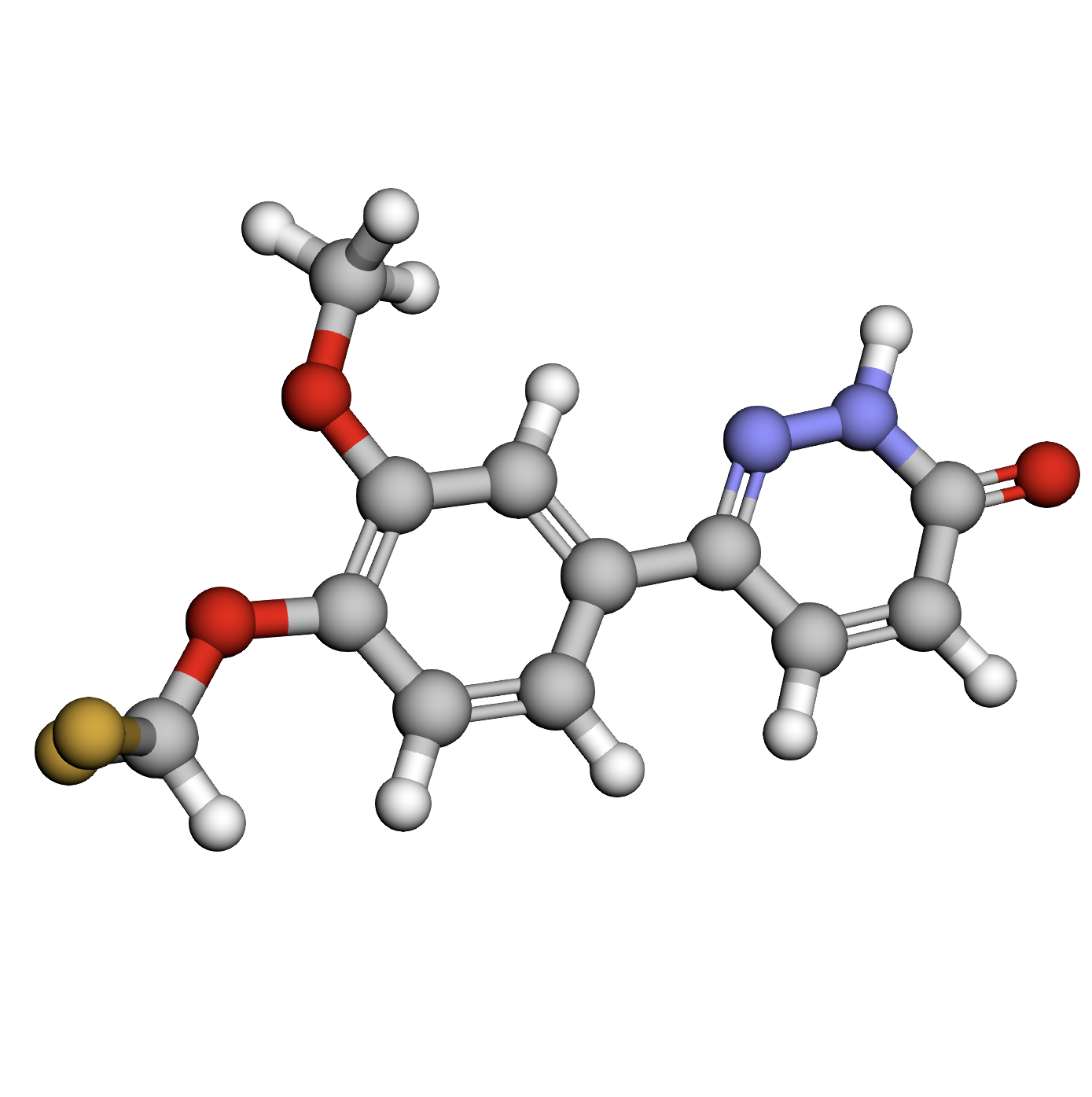} & \includegraphics[trim={2cm 1cm 2cm 1cm},clip,height=2cm]{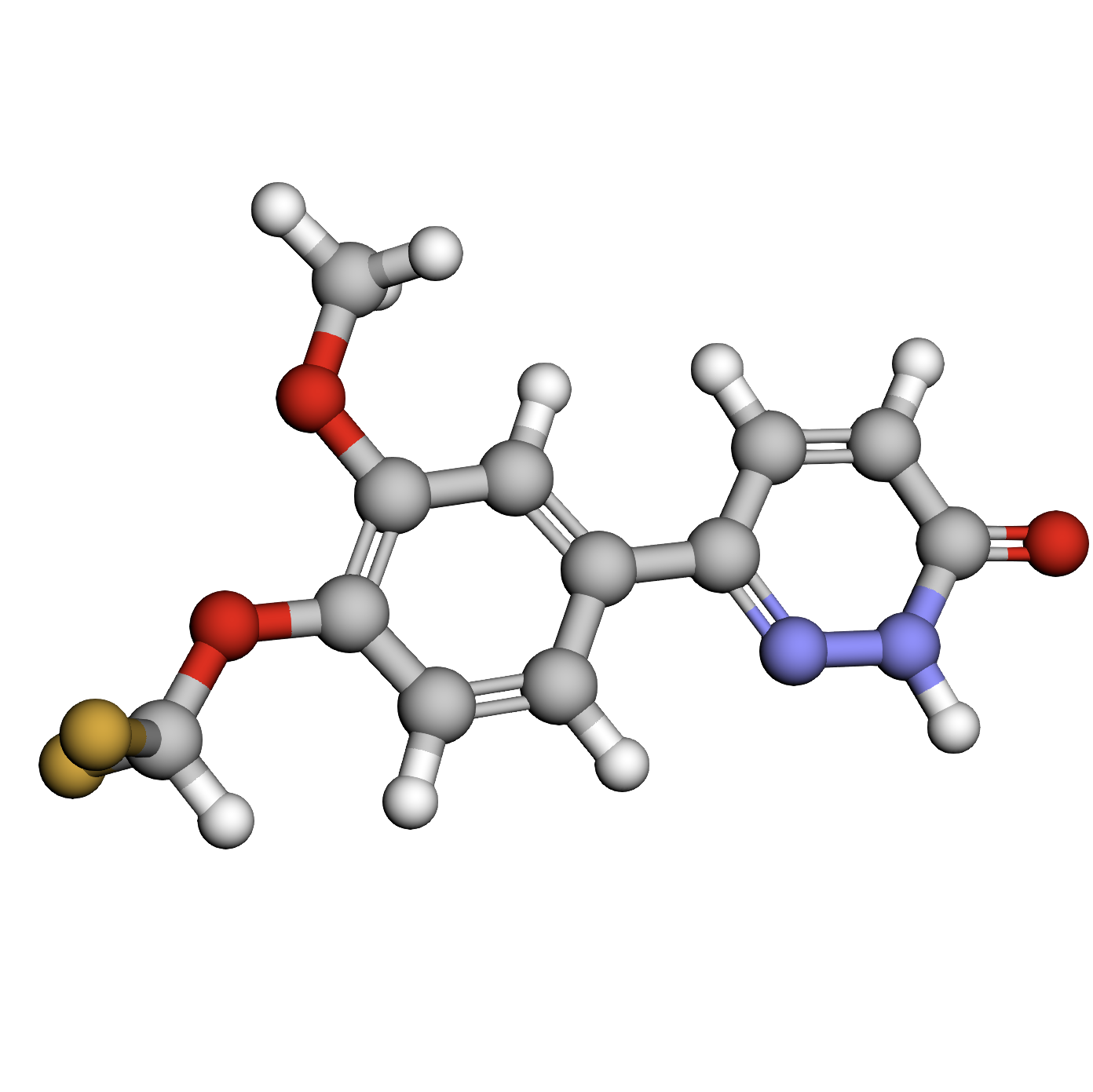} & \includegraphics[trim={2cm 1cm 2cm 1cm},clip,height=2cm]{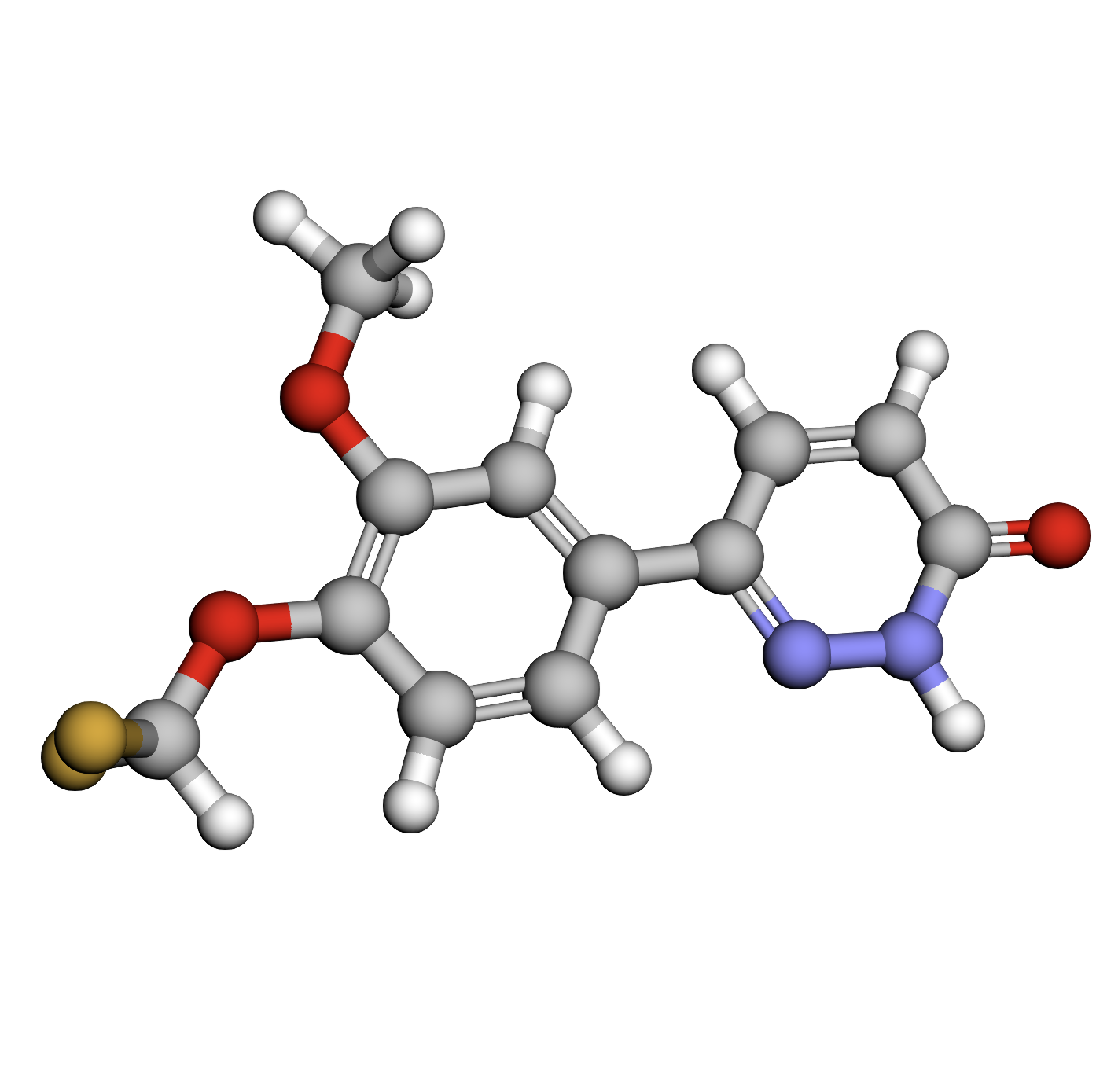} \\
    \includegraphics[trim={0cm 1cm 0cm 1cm},clip,height=2cm]{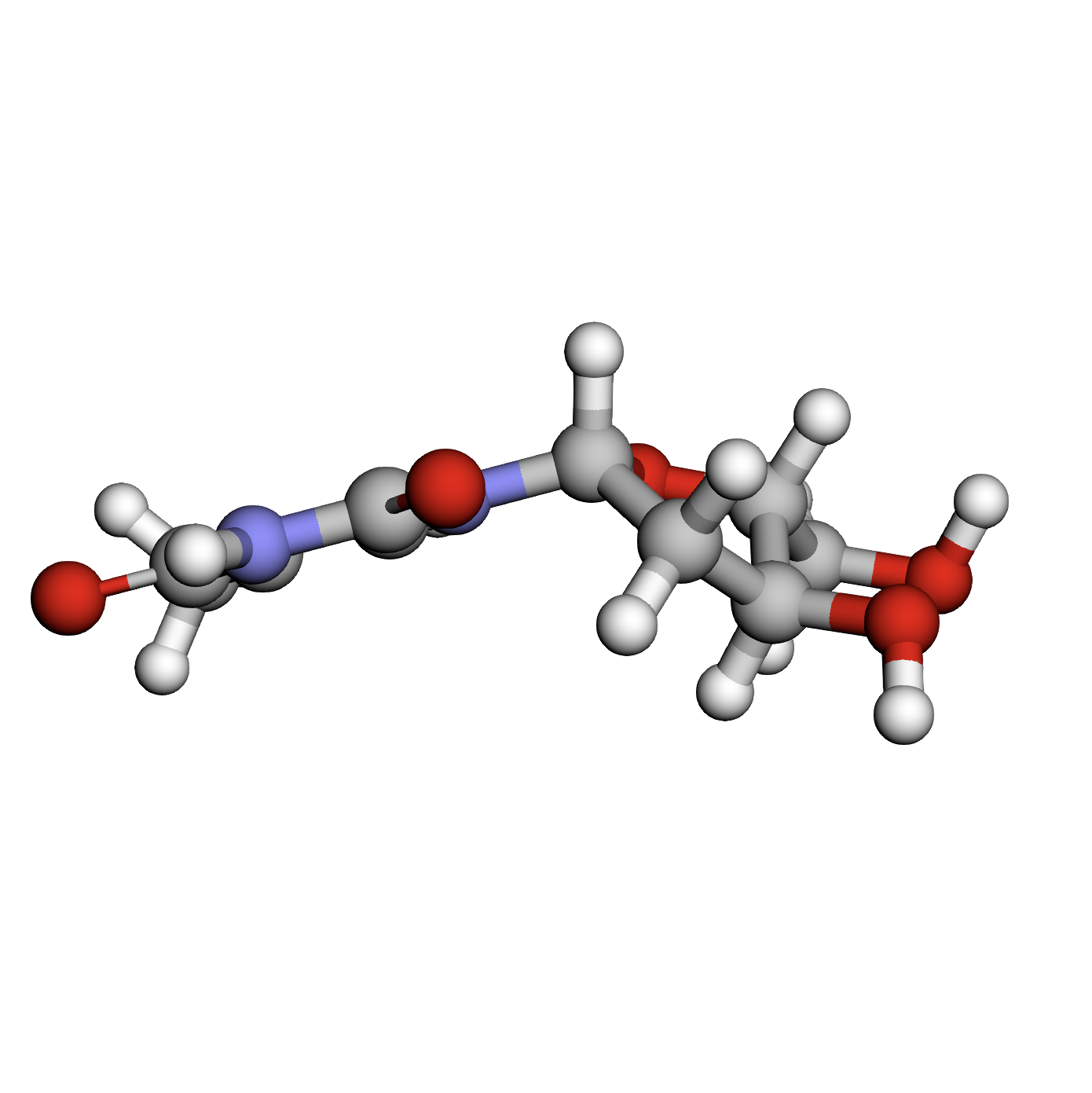} & \includegraphics[trim={0cm 1cm 0cm 1cm},clip,height=2cm]{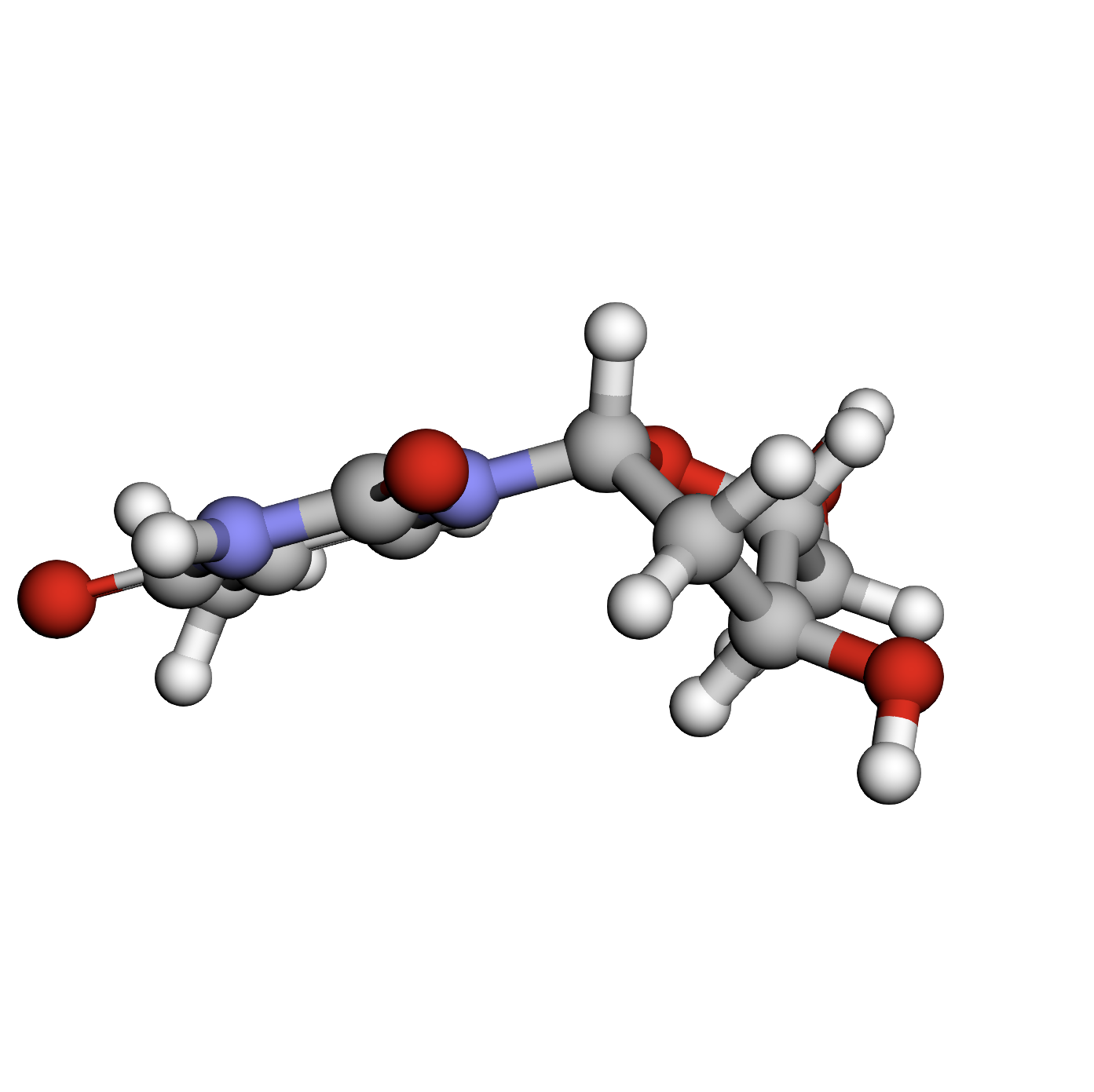} & \includegraphics[trim={0cm 1cm 0cm 1cm},clip,height=2cm]{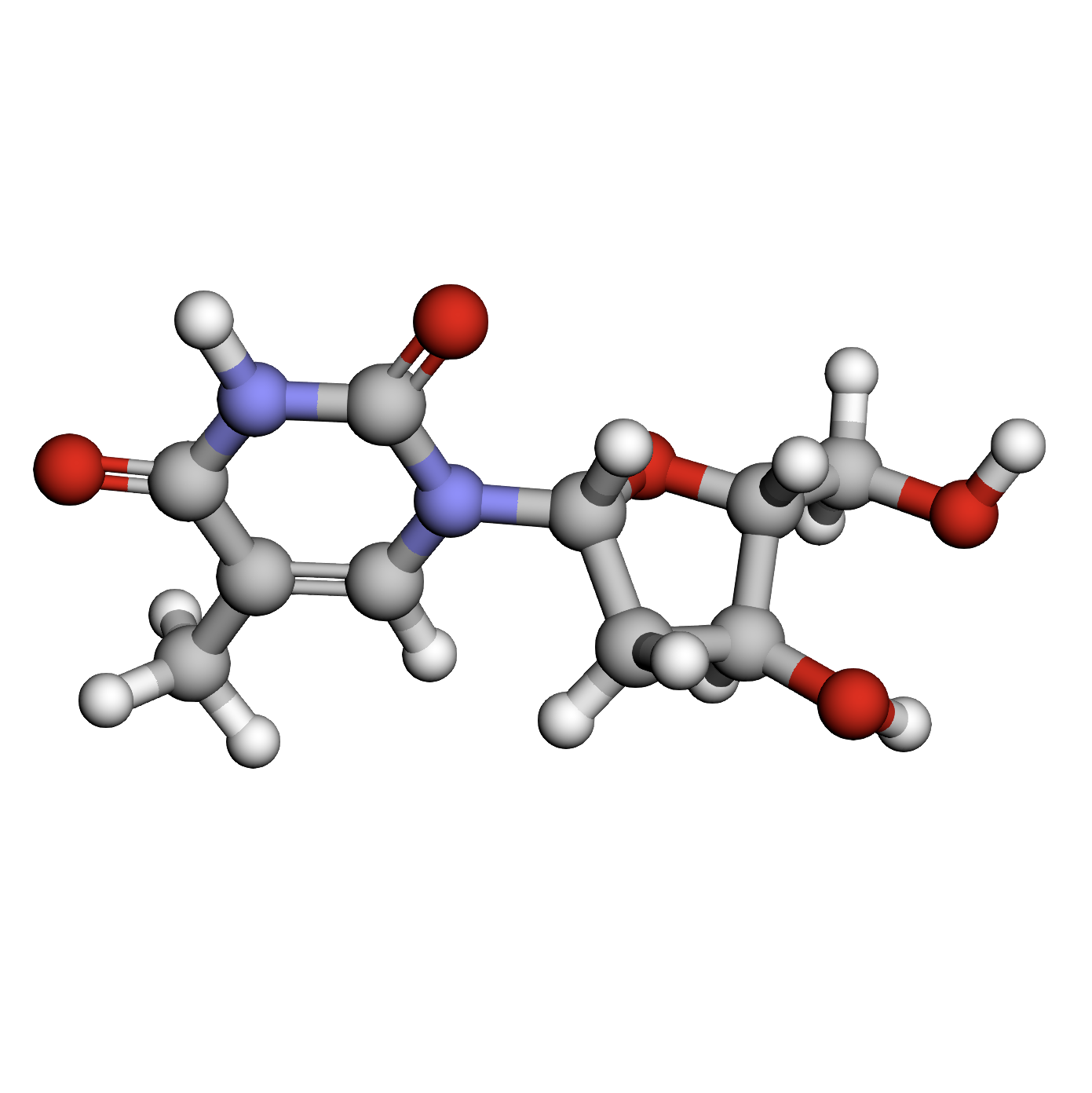} & \includegraphics[trim={0cm 1cm 0cm 1cm},clip,height=2cm]{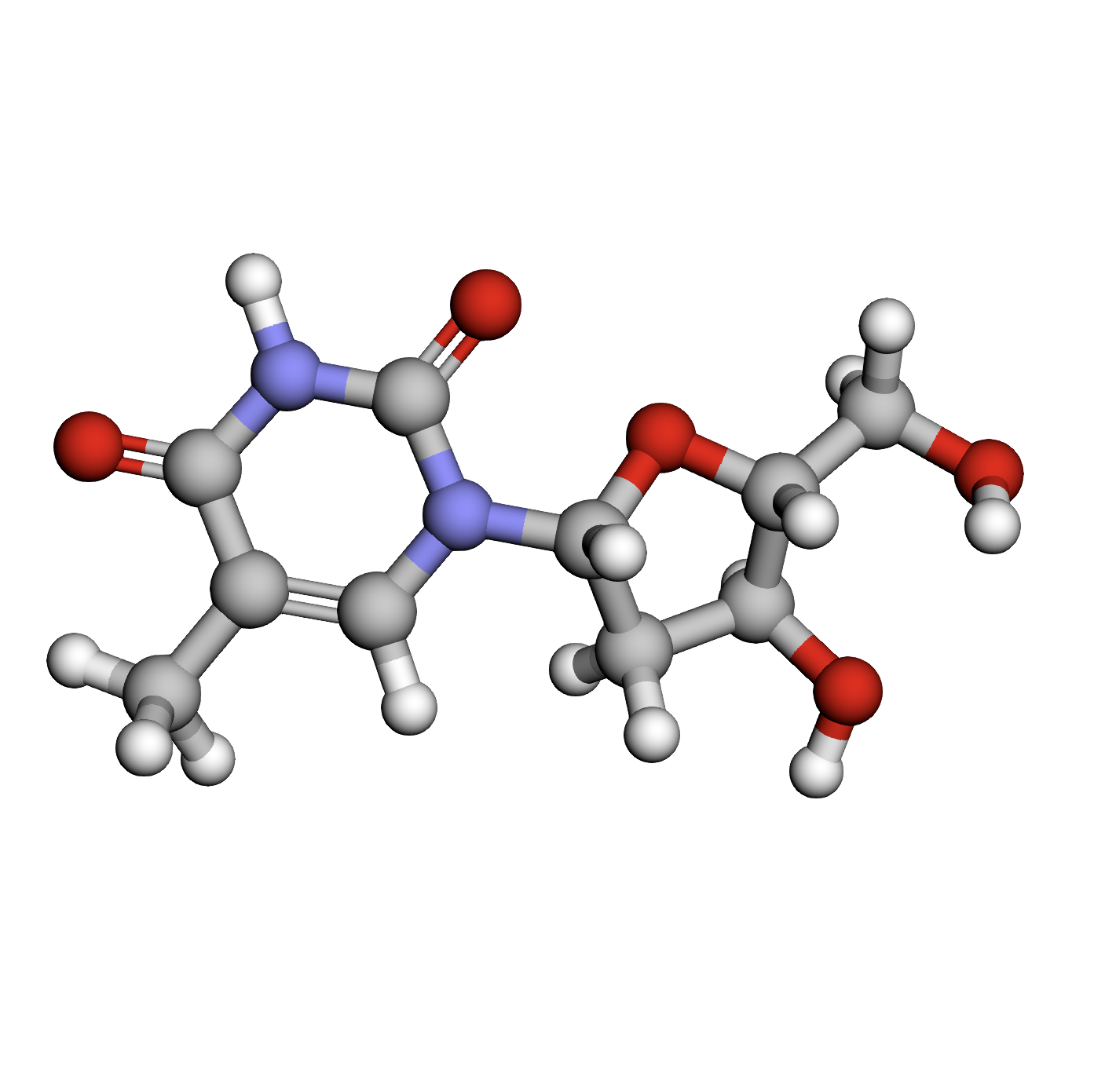} \\
    \includegraphics[trim={2cm 1cm 2cm 1cm},clip,height=2cm]{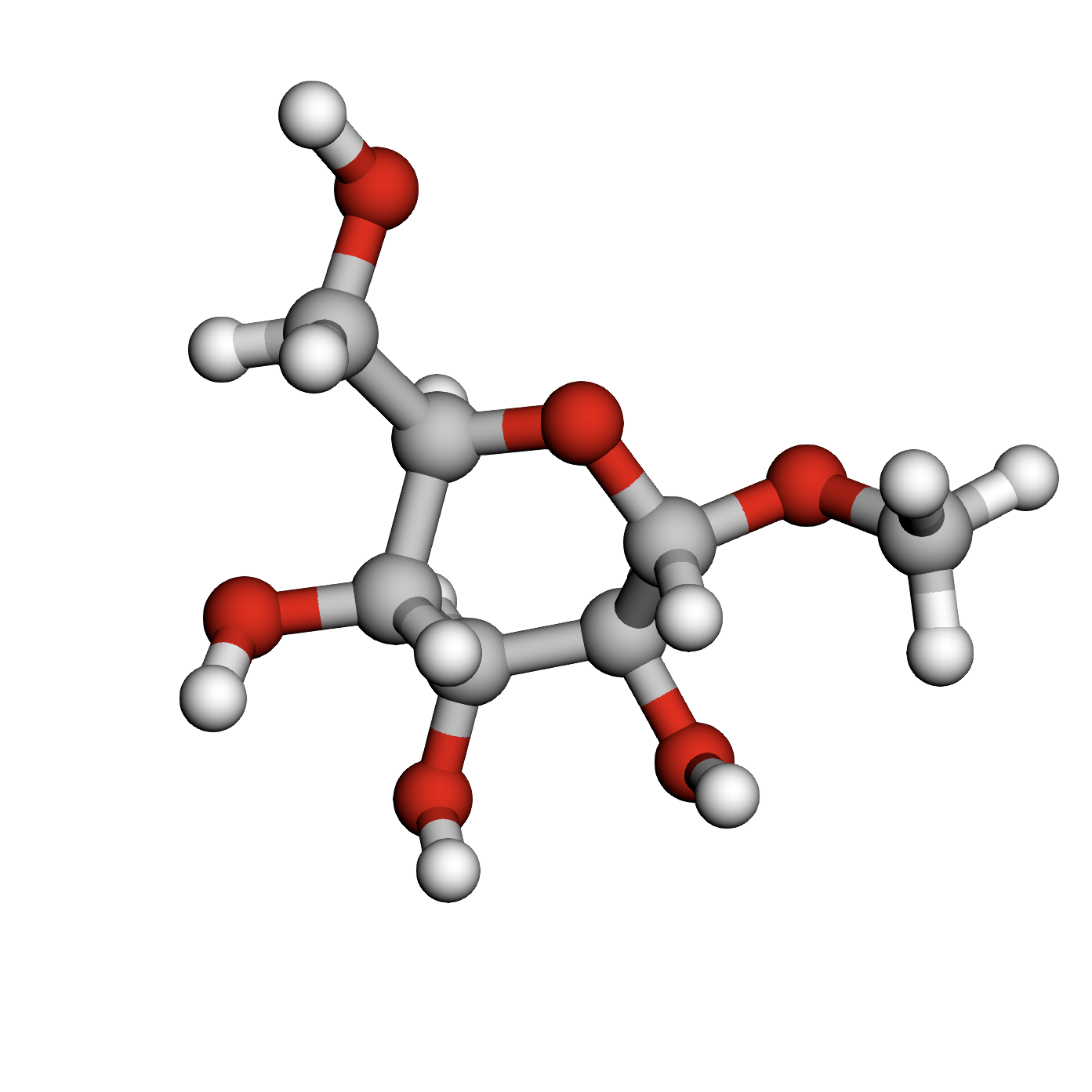} & \includegraphics[trim={2cm 1cm 2cm 1cm},clip,height=2cm]{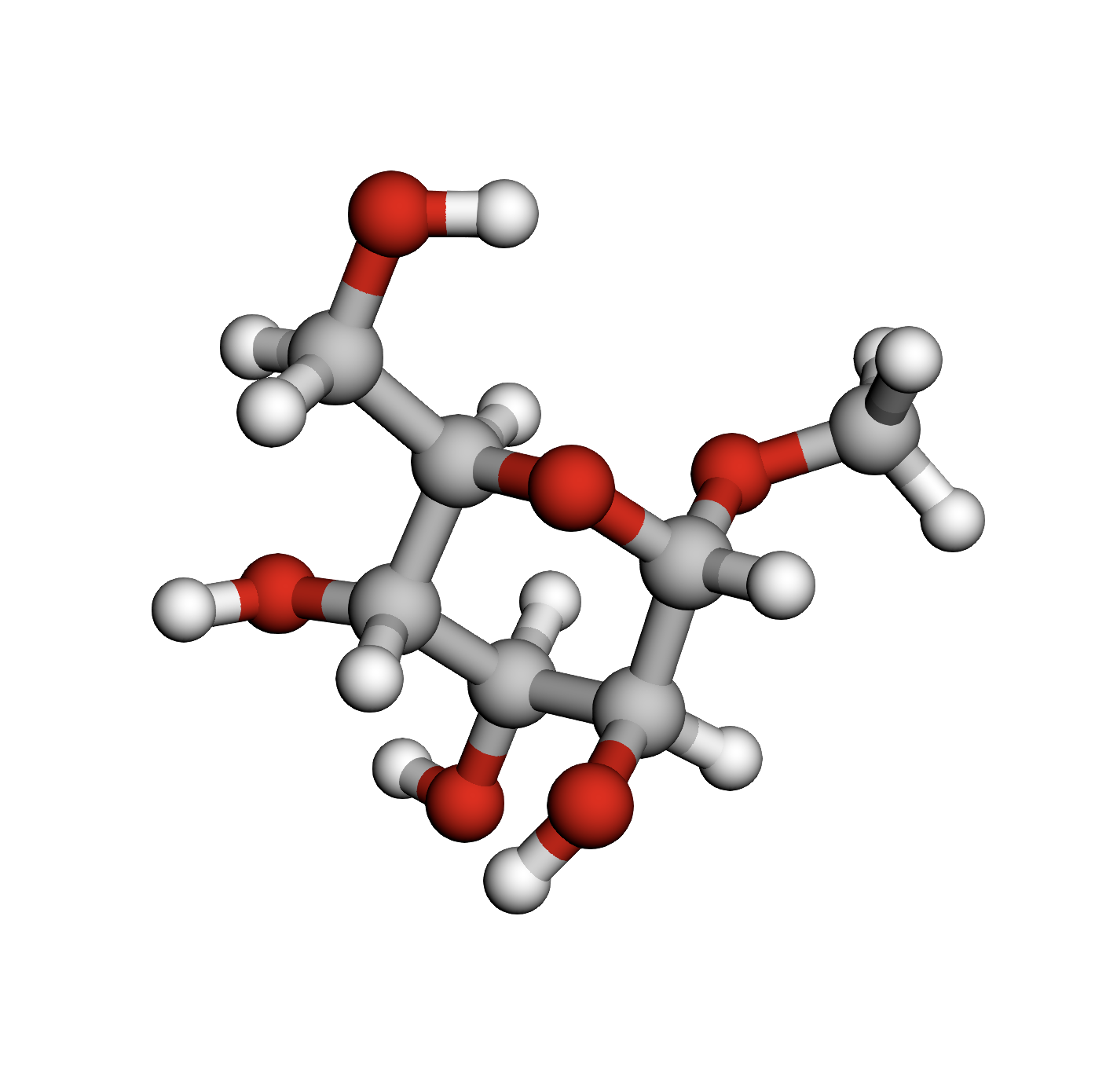} & \includegraphics[trim={2cm 1cm 2cm 1cm},clip,height=2cm]{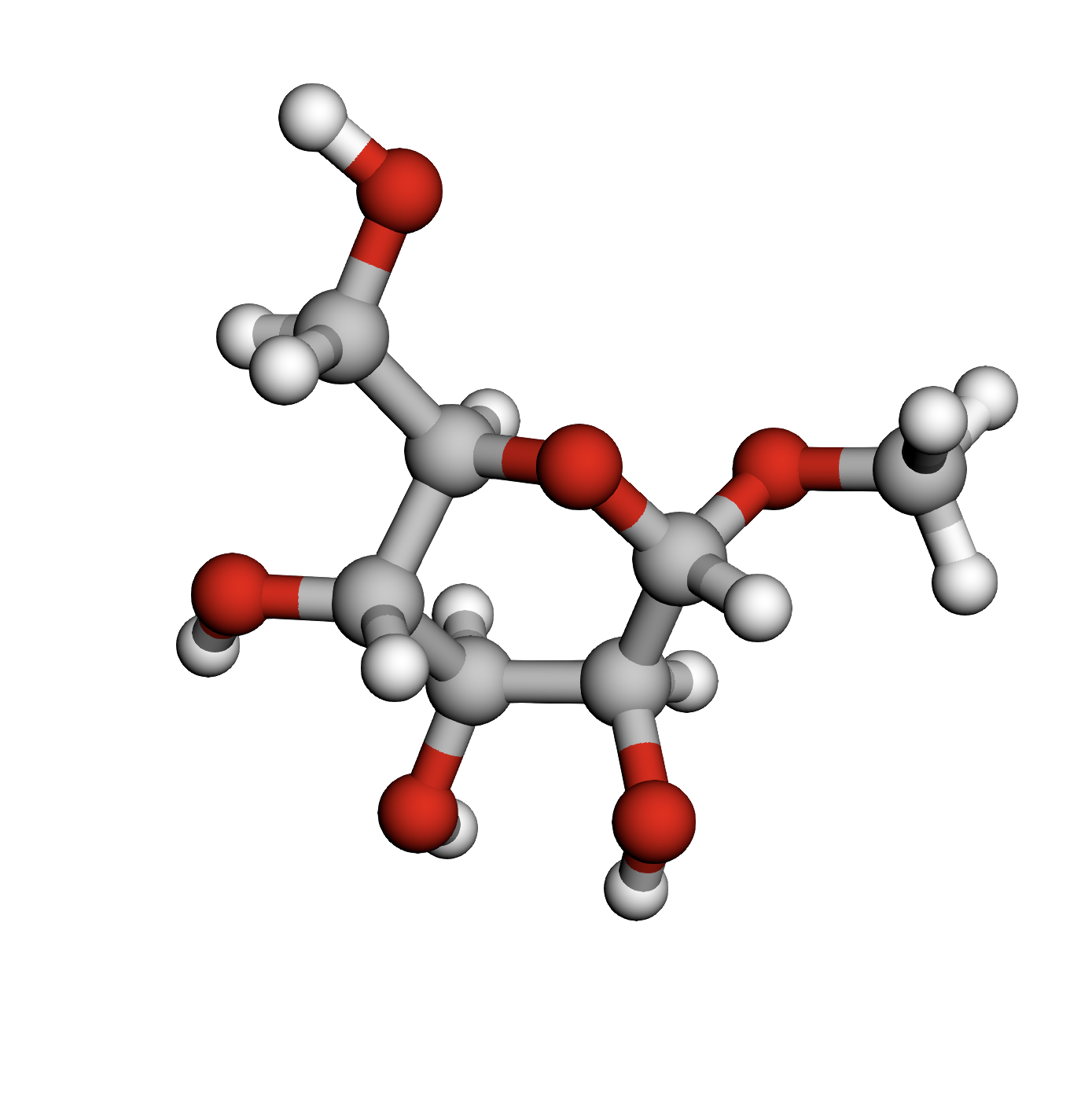} & \includegraphics[trim={2cm 1cm 2cm 1cm},clip,height=2cm]{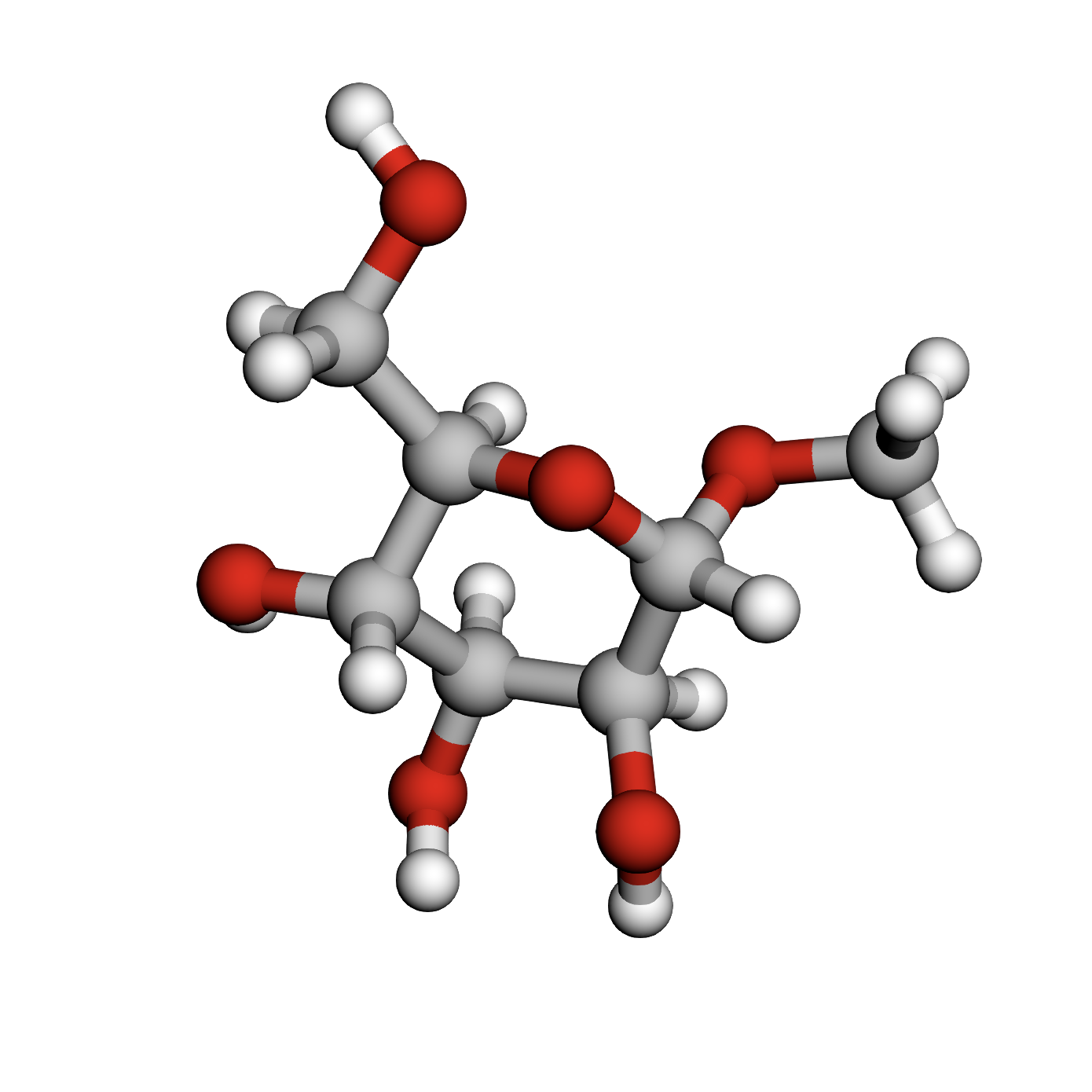} \\
    \includegraphics[trim={0cm 1cm 0cm 1cm},clip,height=2cm]{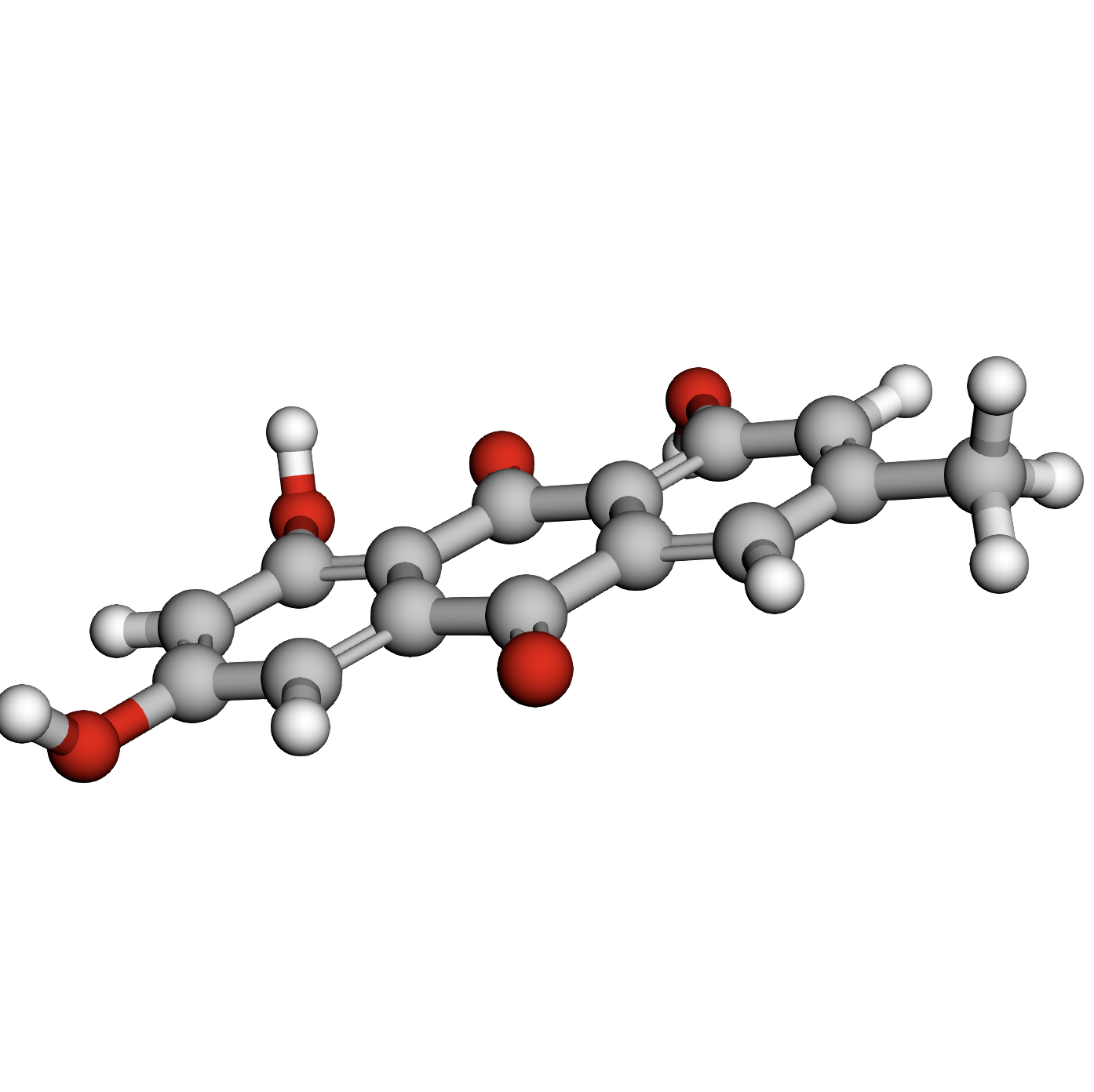} & \includegraphics[trim={0cm 1cm 0cm 1cm},clip,height=2cm]{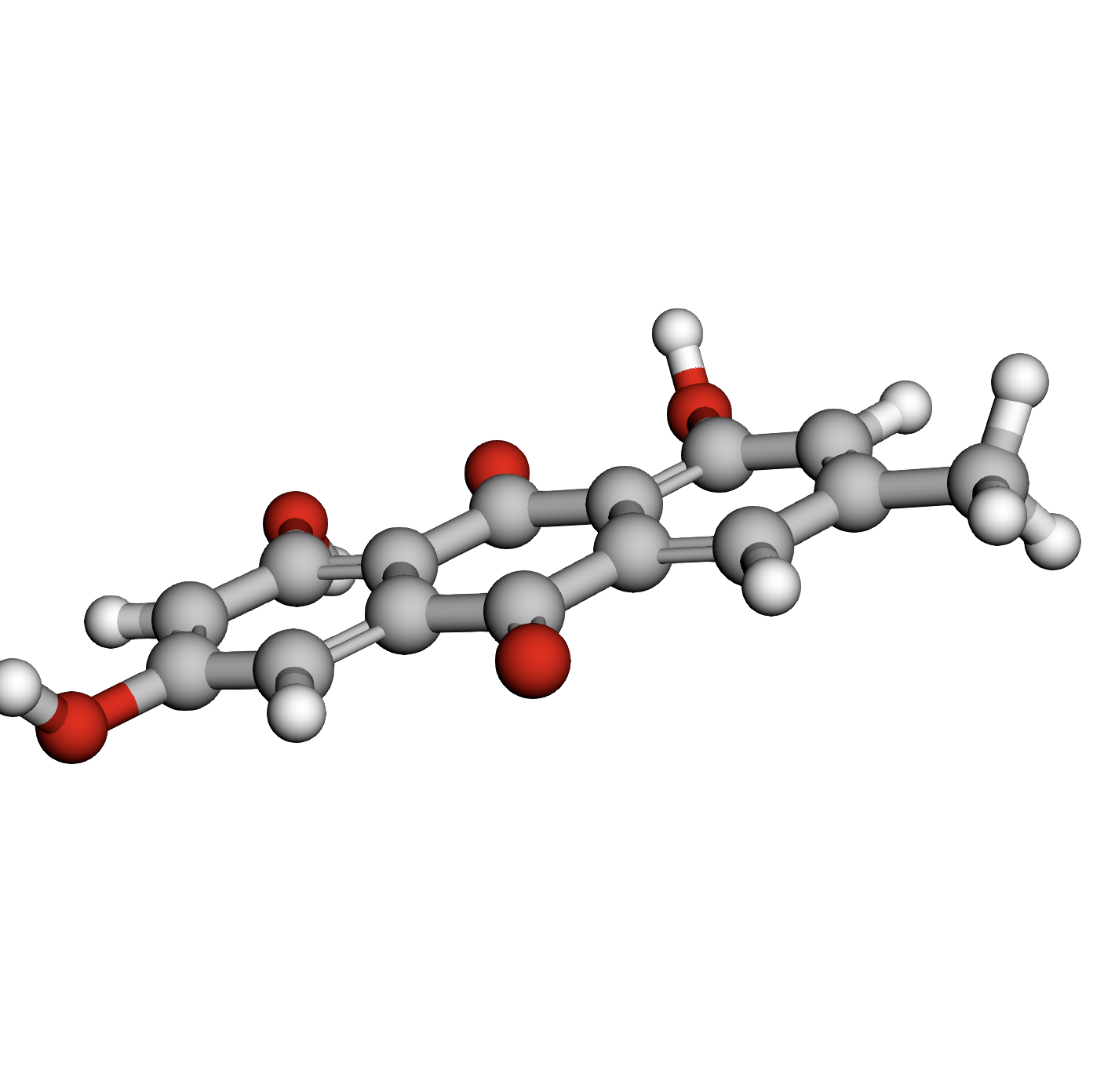} & \includegraphics[trim={0cm 1cm 0cm 1cm},clip,height=2cm]{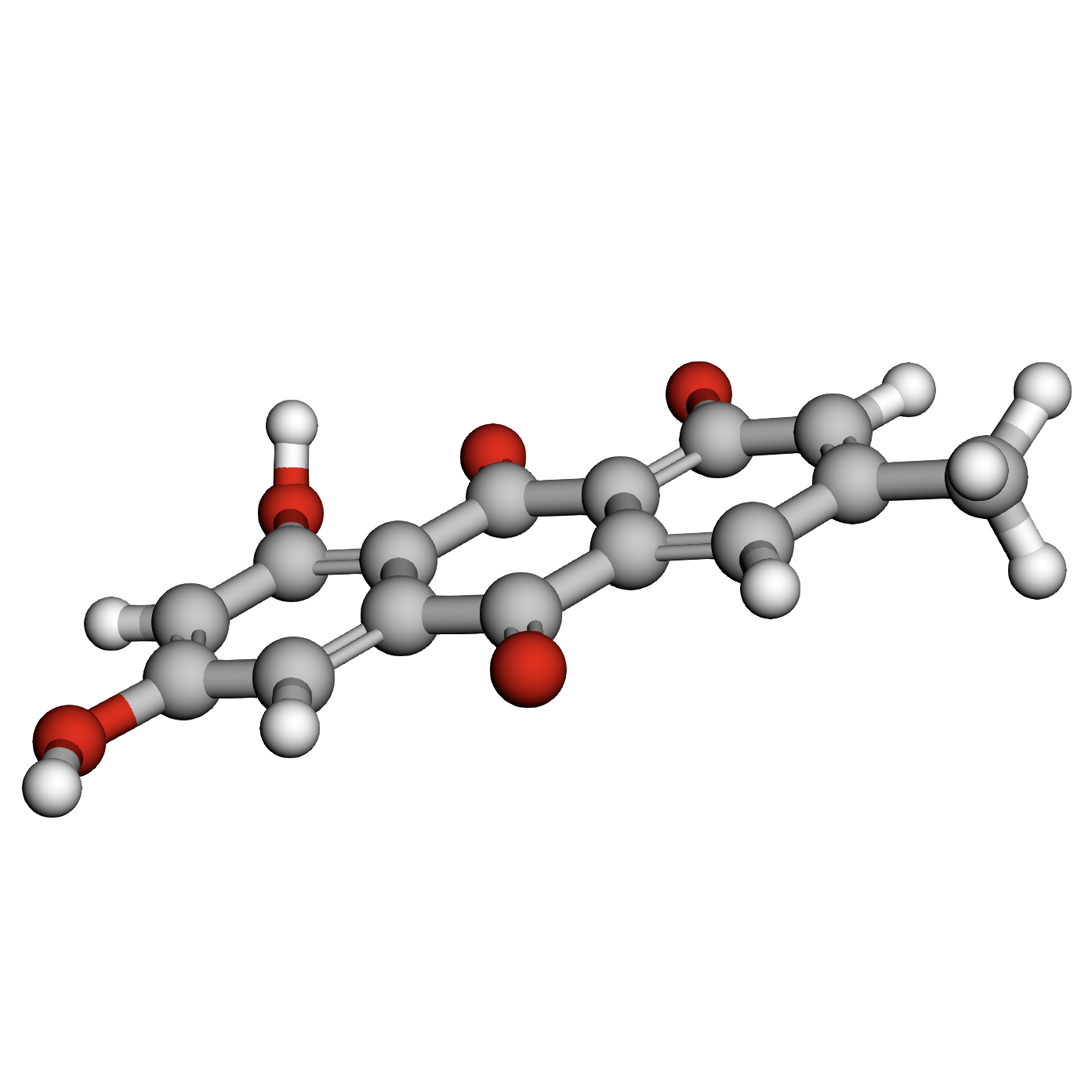} & \includegraphics[trim={0cm 1cm 0cm 1cm},clip,height=2cm]{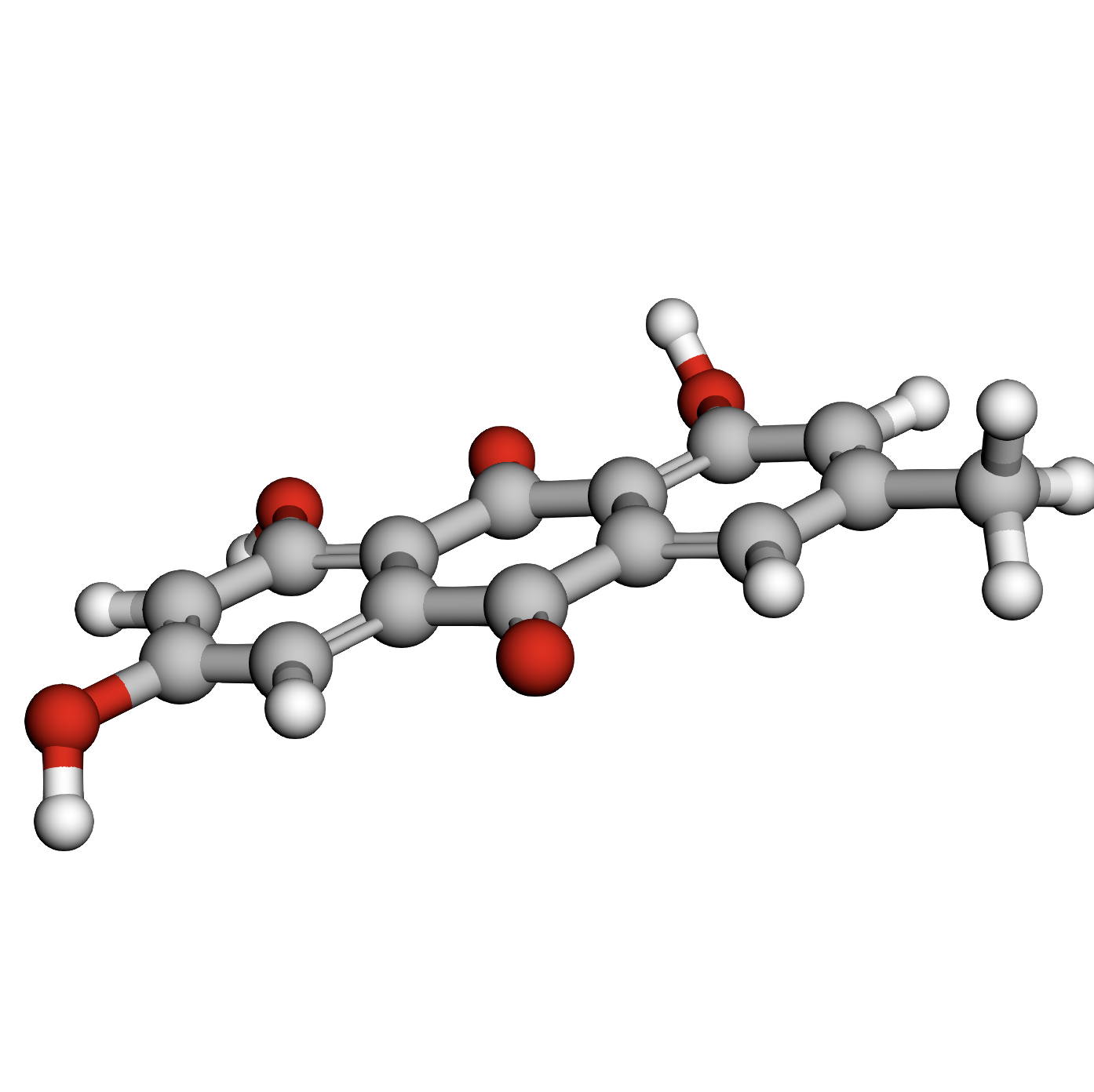} \\
    \bottomrule
    \end{tabular}
\end{table}

\end{document}